\documentclass[10pt]{article}

\usepackage[preprint]{tmlr}

\usepackage{microtype}
\usepackage{graphicx}
\usepackage{subcaption}
\usepackage{booktabs}
\usepackage{amsmath}
\usepackage{amssymb}
\usepackage{mathtools}
\usepackage{amsthm}
\usepackage{amsfonts}
\usepackage{bm}
\usepackage{multirow}
\usepackage{array}
\usepackage{pifont}
\usepackage{enumitem}
\usepackage{float}
\usepackage{wrapfig}
\usepackage{xcolor}
\usepackage{url}

\let\TMLRAND\AND       % Save TMLR's author-block separator 
\let\AND\undefined     % Clear it so algorithmic can define its own \AND
\usepackage{algorithm}
\usepackage{algorithmic}

\usepackage{etoc}
\usepackage{hyperref}
\usepackage[capitalize,noabbrev]{cleveref}
\definecolor{BestBlue}{RGB}{0,92,184}
\definecolor{SecondGreen}{RGB}{0,128,0}

\input{mysymbol.sty}

\newcommand{\cN}{\mathcal{N}}

\newcommand{\EE}{\mathbb{E}}

\newcommand{\new}[1]{#1}
\newcommand{\why}[1]{}

\title{Prior-Informed Flow Matching for Graph Reconstruction}

\author{%
\name Harvey Chen \email hc80@rice.edu \\
\addr Rice University, Houston, TX, USA
\TMLRAND
\name Nicolas Zilberstein \email nzilberstein@rice.edu \\
\addr Rice University, Houston, TX, USA
\TMLRAND
\name Santiago Segarra \email santiago.segarra@rice.edu \\
\addr Rice University, Houston, TX, USA
}

\begin{document}

\maketitle

\begin{abstract}
We introduce \textit{Prior-Informed Flow Matching (PIFM)}, a conditional flow model for graph reconstruction. 
Reconstructing graphs from partial observations remains a key challenge; classical embedding methods often lack global consistency, while modern generative models struggle to incorporate structural priors. 
PIFM bridges this gap by integrating embedding-based priors with continuous-time flow matching. Grounded in a permutation equivariant version of the distortion-perception theory, our method first uses a prior, such as GraphSAGE or node2vec, to form an informed initial estimate of the adjacency matrix based on local information. 
It then applies rectified flow matching to refine this estimate, transporting it toward the true distribution of clean graphs and learning a global coupling. 
Experiments on different datasets demonstrate that PIFM consistently enhances classical embeddings, outperforming them and state-of-the-art generative baselines in reconstruction accuracy. 
Our work is available at:
\begin{center}
\texttt{\url{https://github.com/HC-SGW/PIFM}}
\end{center}

\end{abstract}

\section{Introduction}

Graph generative models have seen remarkable progress in recent years, enabling the synthesis of realistic graph structures in domains such as drug design~\citep{yang2024molecule} and social networks~\citep{grover2019graphite}. 
In particular, diffusion-based~\citep{niu2020permutation, jo2022score, vignac2023digress} and flow-based~\citep{qin2024defog, eijkelboom2024variational} approaches have emerged as state-of-the-art. 
While these models excel at \textit{unconditional} generation and property-controlled generation, their application to inverse problems, and in particular, the reconstruction of a graph from partial observations, remains a fundamental open problem.

% \hc{We study \textit{conditional graph reconstruction}: given a partially observed adjacency matrix $\bbA^{\ccalO}$, the goal is to recover the full adjacency matrix $\bbA$ such that the predicted edges are jointly consistent with the distribution of ground-truth graphs. Graph reconstruction tasks also include link prediction (where the mask is fully known and only missing edges are inferred) and the broader settings where the mask is partial or the observed graph contains spurious edges to be removed. Existing methods fall on opposite sides of this requirement: classical embedding approaches such as Node2Vec~\citep{node2vec, DeepWalk10.1145/2623330.2623732} and inductive methods such as GraphSAGE~\citep{GraphSAGE10.5555/3294771.3294869} produce expressive local features but model each edge prediction independently, with no mechanism to couple predictions across the graph. Recent generative models adapted from image inpainting~\citep{vignac2023digress, trivedi2024editing} or guided by posterior sampling~\citep{sharma2024diffuse, tenorio2025graph} produce globally plausible completions but are designed for constrained generation rather than faithful recovery. This leaves a gap: no existing method combines informative local priors with a global coupling over the full edge set.}

Conditional graph reconstruction aims to recover the topology of a graph from a partially observed set of edges while ensuring that the reconstructed graph aligns with the global distribution of valid graphs. The most common formulation of this problem is link prediction, which typically assumes that the locations of the missing edges are known. However, existing link prediction methods generally model each missing connection independently, failing to enforce global consistency across the reconstructed graph. Classical embedding techniques, such as Node2Vec \citep{node2vec, DeepWalk10.1145/2623330.2623732}, and inductive architectures, like GraphSAGE \citep{GraphSAGE10.5555/3294771.3294869}, learn expressive local representations for this task but do not explicitly capture dependencies between edge predictions or preserve global structural properties.
More recently, diffusion and flow-based generative models have been adapted for graph inpainting and posterior-guided sampling \citep{vignac2023digress, trivedi2024editing, sharma2024diffuse, tenorio2025graph}. While these approaches use powerful graph priors to generate globally plausible completions, they are primarily designed for constrained generation rather than faithful, observation-driven reconstruction. Consequently, a gap remains: existing methods either rely on local, decoupled edge predictions or generate structurally coherent samples that lack reconstruction fidelity.

\begin{figure}[t]
    \centering
    \includegraphics[width=0.75 \linewidth]{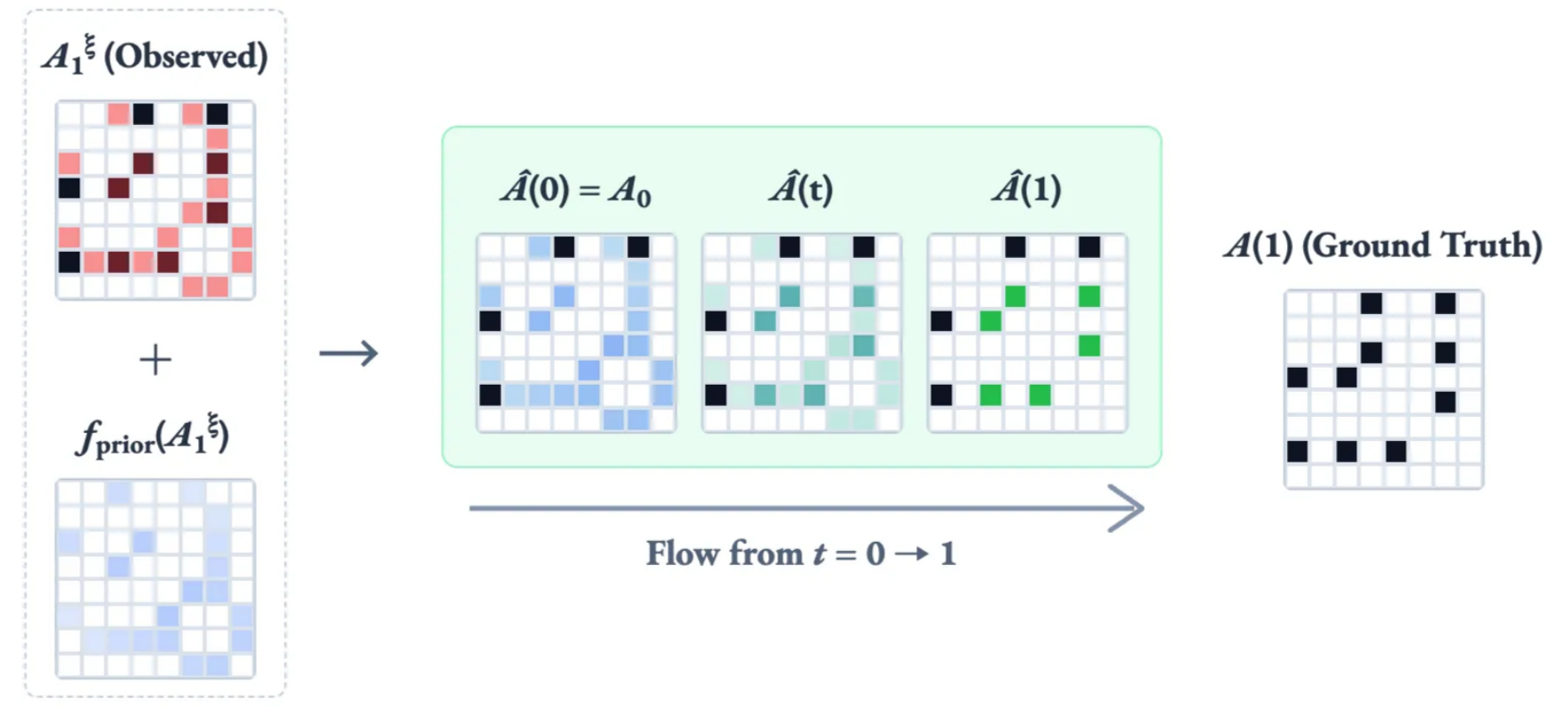}
    \caption{\small{Overview of the \textbf{Prior-Informed Flow Matching (PIFM)} graph reconstruction framework. 
    Starting from a partially observed adjacency matrix $\bbA^{\ccalO} = \xi \odot \bbA$, where $\xi$ denotes a mask, 
    we form an initialization $\bbA_0$ by combining the observed entries with prior predictions 
    $f_{\text{prior}}(\bbA^{\ccalO})$ obtained with an element-wise predictor.
    In dark red we denote the true edges that are masked, while in light red those masked positions  where there is no edge between nodes.
    A rectified flow then interpolates linearly from $\bbA_0$ to the ground-truth graph $\bbA_1 = \bbA$, learning global structural information from a coupling of all the edges.
    The intermediate states $\bbA_t$ improve on the prior-informed initialization, 
    enabling recovery of the missing edges.}}
    \label{fig:framework}
\end{figure}

In this work, we introduce \textbf{Prior-Informed Flow Matching (PIFM)}, a conditional flow model that treats a local link predictor as an informative source distribution and learns a global correction map to the clean graph distribution. Our approach combines the strengths of classical methods with modern generative frameworks; it uses learned structural graph priors (link predictors and latent graph models) while explicitly coupling edge predictions through a global, flow-based reconstruction framework. We formalize this design through the lens of the distortion-perception tradeoff \citep{blau2018perception}, which motivates a two-stage estimator \citep{freirich2021distortion, ohayon2025posteriormean}:
$(i)$ Approximate the Minimum Mean Squared Error (MMSE) estimator using local information. For this, we employ baseline models such as GraphSAGE \citep{GraphSAGE10.5555/3294771.3294869}, Node2Vec \citep{node2vec}, VGAE \citep{kipf2016variational}, or NCNC \citep{wangneural};
$(ii)$ Learn a rectified flow \citep{liuflow, albergo2023stochastic} that refines this initial estimate by explicitly modeling the dependencies among edge predictions.

We validate the advantages of PIFM through experiments on datasets with diverse structural characteristics, including both dense and sparse graphs. Our results demonstrate that PIFM effectively integrates local architectural priors with global flow-based modeling, offering a principled framework for graph reconstruction.

Our contributions are as follows:
\begin{itemize}[topsep=2pt,itemsep=1pt]

\item We formulate conditional graph reconstruction through the lens of the permutation-equivariant distortion--perception trade-off.

\item We propose \textsc{PIFM}, a two-stage reconstruction framework that first constructs a prior-informed source distribution using embeddings from latent graph models, and then refines these predictions with a flow-matching model.

\item We show that existing link prediction methods primarily rely on local edge-wise predictions and lack mechanisms to enforce global structural consistency. \textsc{PIFM} addresses this limitation by learning a generative model that integrates information across the full graph during reconstruction.

\item We empirically validate our approach on link prediction and two blind reconstruction settings, termed \textit{expansion} (recovering missing edges) and \textit{denoising} (removing spurious edges), showing consistent improvements over predictors based solely on local information.

\end{itemize}

\section{Related Work}

\paragraph{Flow/Diffusion models on graphs.}
Diffusion and flow-based graph generative models have shown impressive performance in recent years. 
Early models, namely EDP-GNN~\citep{niu2020permutation} and GDSS~\citep{jo2022score}, employ score-based \textit{continuous} diffusion over a relaxation of the graph structure. 
However, given that graphs are inherently discrete, subsequent work has explored discrete diffusion processes~\citep{austin2021structured}. 
Models like DiGress~\citep{vignac2023digress} demonstrated the effectiveness of this approach, which has been further advanced by discrete flow-based models like DeFoG~\citep{qin2024defog} and variational approaches like CatFlow~\citep{eijkelboom2024variational}.
A common point of these models is their reliance on a simple source distribution, such as Gaussian (continuous) or uniform (discrete) noise.
While effective for unconstrained generation, recent work on image-based inverse problems demonstrates the advantages of learning a data-dependent flow, using a prior-informed source distribution~\citep{albergo2023stochastic, delbracioinversion, ohayon2025posteriormean}. In the graph domain, concurrent work~\citep{wijesinghe2026flowette} also departs
from a simple source distribution by employing structural graphette priors, though it targets unconditional graph generation rather than reconstruction from partial observations.

\paragraph{Graph topology inference via flow/diffusion-based solvers.} 
Graph topology inference -- the task of recovering hidden edges from a partially observed graph -- is a long-standing inverse problem~\citep{segarra2017network, dong2016learning}. 
Several methods adapt diffusion for constrained graph generation, which is related to but distinct from topology inference. DiGress~\citep{vignac2023digress} introduced an inpainting mechanism, inspired by Repaint~\citep{lugmayr2022repaint}, to generate graph structures consistent with a partial observation. 
Similarly, in~\citet{trivedi2024editing} a similar mechanism is used for completing partially observed graphs. 
PRODIGY~\citep{sharma2024diffuse} enforces hard constraints by projecting the graph estimate onto a feasible set at each sampling step. 
More recently, GGDiff~\citep{tenorio2025graph} incorporates a guidance mechanism as a flexible alternative to inpainting.
However, all these methods are designed for \textit{constrained generation} (e.g., molecule generation with a given scaffold) rather than \textit{recovering masked edges from a partially observed graph}. 
Hence, to the best of our knowledge, designing a diffusion-based model explicitly for graph topology inference remains an open problem. 

\paragraph{Link prediction and Graph Autoencoders.}
Graph reconstruction from partial observations is closely related to link prediction, which aims to infer whether unobserved edges should exist in a partially observed graph~\citep{newman2001clustering, adamic2003friends, zhou2009predicting}. 
Classical approaches rely on topology based heuristics, while embedding based methods such as DeepWalk~\citep{DeepWalk10.1145/2623330.2623732} and node2vec~\citep{node2vec} learn node representations from random walks and score candidate edges from pairwise embedding features. 
Graph neural network (GNN) methods further improve link prediction by learning node representations through message passing. 
For example, GraphSAGE learns inductive neighborhood aggregation functions that can generalize across nodes or graphs~\citep{GraphSAGE10.5555/3294771.3294869}, while variational graph autoencoders (VGAE) encode the observed graph into latent node embeddings and decode pairwise edge probabilities~\citep{kipf2016variational}. 
Other methods model pairwise or path based structure more 
directly, such as NBFNet, which formulates link prediction 
as neural path aggregation through a Bellman Ford style 
recurrence~\citep{NBFNets10.5555/3540261.3542517}.

\section{Background}
\label{sec:background}

We represent an undirected graph $\ccalG = \{\ccalV, \ccalE\}$, where $\ccalV$ denotes the nodes and $\ccalE$ the edges, by its binary symmetric adjacency matrix $\bbA \in \reals^{N\times N}$. 
Table~\ref{tab:notation} summarizes the main notation used throughout the method.

\begin{table}[h!]
\centering
\small
\caption{\small Notation used in PIFM.}
\label{tab:notation}
\setlength{\tabcolsep}{5pt}
\renewcommand{\arraystretch}{1.05}
\begin{tabular}{ll}
\toprule
Symbol & Meaning \\
\midrule
$\bbA_1$ & Clean ground-truth adjacency matrix \\
$\bbA^{\ccalO}$ & Observed partial adjacency matrix \\
$\xi$ & Binary observation mask \\
$\hat{\bbA}^{*}$ & Prior estimate or approximate MMSE estimate \\
$\bbA_0$ & Prior-informed source sample for flow matching \\
$\bbA_t$ & Intermediate state along the flow path \\
$\widehat{\bbA}$ & Final reconstructed adjacency matrix \\
\bottomrule
\end{tabular}
\end{table}

\subsection{Continuous flow matching for graph generation}
\label{sec:cont-FM}

Flow matching~\citep{albergo2023stochastic, lipmanflow} is a family of generative models that defines a continuous-time transport map from samples $\bbA_0$ drawn from a source distribution $p_0$ to samples $\bbA_1$ from a target distribution $p_1$.
It is governed by the ODE
\begin{equation}
    d\bbA_t = v(\bbA_t, t)\,dt,
\end{equation}
where $v(\cdot, t)$ is a velocity field and $\bbA_t$ denotes a forward process, also known as stochastic interpolant, for $t \in [0, 1]$. 
Typically, $p_0$ is a tractable distribution (e.g., a Gaussian distribution), while $p_1$ corresponds to the data distribution.
To generate new samples, one must specify both $\bbA_t$ and $v$.
A common choice for the forward process is $\bbA_t = \alpha_t \bbA_0 + \beta_t \bbA_1$, where $\alpha_t$ and $\beta_t$ are differentiable functions such that $\alpha_0 = 1$, $\beta_0 = 0$ and $\alpha_1 = 0$, $\beta_1 = 1$. 
Differentiating this path gives a velocity $v(\bbA_t, t) = \dot{\alpha}_t \bbA_0 + \dot{\beta}_t \bbA_1$. 
Despite its closed-form, this expression depends explicitly on $\bbA_1$, making it impractical since the target is unknown at inference/sampling.
To circumvent this, we instead consider
$v(\bbA_t, t) = \mathbb{E}_{\bbA_0, \bbA_1}[\new{\dot{\alpha}_t \bbA_0 + \dot{\beta}_t \bbA_1} \mid \bbA_t]$, the conditional expectation of the velocity given $\bbA_t$~\citep{albergo2023stochastic},
which is then approximated with a neural network $v_\theta$.
The network is trained using a mean squared error loss:
\begin{equation}
\label{eq:loss_flow}
\mathbb{E}_{t, \bbA_0, \bbA_1} \left[ \left\| v_\theta(\bbA_t, t) - (\new{\dot{\alpha}_t \bbA_0 + \dot{\beta}_t \bbA_1}) \right\|_2^2 \right].
\end{equation}
\why{Swapped the $\bbA_0/\bbA_1$ indices in both the conditional expectation and the loss target. Differentiating $\bbA_t=\alpha_t\bbA_0+\beta_t\bbA_1$ gives $\dot{\alpha}_t\bbA_0+\dot{\beta}_t\bbA_1$ (matching the velocity expression two lines above). For rectified flow ($\alpha_t{=}1{-}t$, $\beta_t{=}t$) this reduces to $\bbA_1-\bbA_0$, which is the target the Euler-step update $\hat{\bbA}\leftarrow\hat{\bbA}+\tfrac{1}{K}v_{\theta^*}$ in Alg.~\ref{alg:train&sample} needs in order to integrate from $\bbA_0$ to $\bbA_1$ rather than the wrong way.}
In particular, this formulation does not require $\bbA_0$ and $\bbA_1$ to be independent; in fact, they might be sampled from a joint distribution, allowing for richer transport plans in cases where paired data is available.
This has been exploited to solve inverse problems on images~\citep{ohayon2025posteriormean, interpolant, delbracioinversion}, and is directly related to our proposed method, as described later.

Throughout this work, we consider the \textit{rectified flow} case~\citep{liuflow}, where $\alpha_t = 1 - t$ and $\beta_t = t$. 
As shown in~\citet{tongimproving}, the velocity field associated with this linear path approximates the optimal transport vector field when the joint distribution $p(\bbA_0, \bbA_1)$ closely resembles the optimal coupling between the marginals $p(\bbA_0)$ and $p(\bbA_1)$.
We defer to Appendix~\ref{app:background-diffusion-inverse} a more detailed background on generative models on graphs beyond continuous flow matching, including diffusion-based models, as well as related works.

\subsection{Distortion--perception formulation for conditional graph reconstruction}
\label{subsec:dist-perc}

Our goal is to reconstruct the ground-truth adjacency matrix $\bbA$ of a graph $\ccalG$ from a partially observed version, denoted by $\bbA^{\ccalO}$. 
Similarly to image restoration, there are two complementary criteria to assess the quality of reconstructed graphs: $(i)$ the average closeness of the reconstruction to the ground truth, measured through a distortion metric, and $(ii)$ the similarity in distribution between reconstructed and real graphs, namely, how well the reconstructed graph resembles a sample from the underlying graph distribution.

These two objectives are generally at odds and can be formalized through the perception--distortion trade-off introduced in~\citet{blau2018perception}:
\begin{equation}
\label{eq:dist-per}
D(P) =
\min_{p(\hat{\bbA}|\bbA^{\ccalO})}
\left\{
\mathbb{E}_{p(\bbA,\hat{\bbA})}
\left[
\Delta(\bbA,\hat{\bbA})
\right]
\;:\;
d(p_{\bbA}, p_{\hat{\bbA}}) \leq P
\right\},
\end{equation}
where $\hat{\bbA}$ denotes an estimator of $\bbA$ conditioned on the observation $\bbA^{\ccalO}$, $\Delta(\cdot,\cdot)$ is a distortion metric (e.g., MSE), and $d(p_{\bbA}, p_{\hat{\bbA}})$ measures the divergence between the distribution of clean graphs $p_{\bbA}$ and the distribution of reconstructed graphs $p_{\hat{\bbA}}$.

The function in~\eqref{eq:dist-per} has been extensively studied in the image domain, where different values of $P$ yield estimators with different trade-offs between reconstruction fidelity (distortion) and perceptual realism. We adapt this framework to graph reconstruction by estimating the adjacency matrix $\bbA$ while accounting for the symmetry constraints induced by the permutation invariance of graph representations.

\paragraph{MSE distortion.}
We first consider the distortion metric
$
\Delta(\bbA,\hat{\bbA}) = \|\bbA - \hat{\bbA}\|_F^2,
$
corresponding to the mean squared error (MSE). 
This setting was studied in~\citet{freirich2021distortion}, particularly for the two extreme cases $P=\infty$ and $P=0$. When $P=\infty$, the problem reduces to minimizing distortion without imposing perceptual constraints. The optimal estimator is the posterior mean
\[
\hat{\bbA}^* = \mathbb{E}[\bbA \mid \bbA^{\ccalO}],
\]
which minimizes the expected MSE. Although optimal in terms of distortion, this estimator may produce unrealistic graphs whose structural properties deviate from those of the underlying graph distribution.

At the other extreme, $P=0$ enforces perfect perceptual consistency, meaning that the reconstructed graphs follow exactly the same distribution as the ground-truth graphs, i.e., $p_{\hat{\bbA}} = p_{\bbA}$. 
As shown in~\citet{freirich2021distortion}, the corresponding estimator can be characterized through the optimal transport problem
\begin{equation}
\label{eq:dist-per-p0}
p_{\hat{\bbA},\hat{\bbA}^*}^*
=
\argmin_{p \in \Pi(p_{\bbA}, p_{\hat{\bbA}^*})}
\mathbb{E}
\left[
\|\hat{\bbA} - \hat{\bbA}^*\|_F^2
\right],
\end{equation}
where $\Pi(p_{\bbA}, p_{\hat{\bbA}^*})$ denotes the set of couplings with marginals $p_{\bbA}$ and $p_{\hat{\bbA}^*}$.

For the special case of squared-error distortion with the Wasserstein-2 perception index, \citet{freirich2021distortion} show that the estimator at any point on the distortion--perception curve can be obtained by interpolating the two extremes, $\hbA_p = (1-\tfrac{P}{P_{\infty}})\hbA + \tfrac{P}{P_{\infty}}\hbA^*$, where $\hbA$ is the perfect-perception ($P=0$) estimator, $\hbA^*$ is the MMSE estimator, and $P_{\infty}$ is the perception index attained by $\hbA^*$.
Therefore, obtaining the estimator associated with $D(0)$ amounts to solving an optimal transport problem between the distribution of clean graphs $p_{\bbA}$ and the distribution of the MMSE estimator $p_{\hat{\bbA}^*}$. 
In practice, we approximate a $P=0$ estimator by $(i)$ computing the posterior mean $\hat{\bbA}^* = \mathbb{E}[\bbA \mid \bbA^{\ccalO}]$, and $(ii)$ learning a transport from $p_{\hat{\bbA}^*}$ to $p_{\bbA}$ with the rectified flow described below. The learned map is trained to push $p_{\hat{\bbA}^*}$ toward $p_{\bbA}$, i.e., to approach perfect perception ($P=0$); this is only approximate, both because of finite model capacity and because a continuous, invertible flow cannot exactly match the discrete distribution of binary adjacency matrices. We likewise do not claim it recovers the exact optimal-transport coupling of~\eqref{eq:dist-per-p0}. Intuitively, this procedure refines the coarse estimate $\hat{\bbA}^*$ into a graph that is consistent with the underlying data distribution.

% While the solution corresponding to $P=\infty$ typically achieves lower MSE, it may generate graphs that violate important structural constraints. This limitation is particularly critical in applications such as conditional molecular generation, where reconstructed molecules must satisfy strict chemical validity conditions. In such settings, the solution with $P=0$ is preferable, since it guarantees that generated samples remain consistent with the target graph distribution. For this reason, our work focuses primarily on the case $P=0$.

We focus on the $P=0$ case: while $P=\infty$ is MSE-optimal, it may produce graphs that violate structural constraints (e.g., chemical validity in molecule generation), whereas $P=0$ guarantees distributional consistency.

\paragraph{Cross-entropy distortion.}
We also consider a cross-entropy-based distortion, which is particularly natural for graph reconstruction since adjacency matrices are discrete-valued objects. Specifically, letting $p_\theta^d(k \mid \bbA_t,t)$ denote the categorical distribution induced by the flow model for edge variable $d$, we define
\[
\Delta(\bbA,\hat{\bbA})
=
-
\sum_{d=1}^{D}
\sum_{k=1}^{K^d}
\mathbb{I}[A^d = k]\,
\log p_\theta^d(k \mid \bbA_t,t),
\]
which corresponds to the cross-entropy between the ground-truth adjacency matrix and the predicted edge distributions, and where $D = \frac{N(N-1)}{2}$ denotes the number of edge variables.
Unlike the MSE case, this formulation does not admit the same closed-form characterization of the perception--distortion trade-off. Nevertheless, as we demonstrate experimentally, it exhibits qualitatively similar behavior, yielding a comparable trade-off between reconstruction fidelity and perceptual consistency.

\section{Method}
\label{sec:method}

PIFM can be viewed as a two-stage refinement framework for conditional graph reconstruction, motivated by the distortion--perception formulation introduced in Section~\ref{subsec:dist-perc}. 
First, a structural prior estimates the conditional mean 
$\mathbb{E}[\bbA \mid \bbA^\ccalO]$ from the observed subgraph 
$\bbA^{\ccalO}$ by predicting marginal probabilities for the masked edges. 
While this estimate minimizes distortion locally, it does not necessarily produce a globally consistent graph. 
To address this, PIFM uses the resulting adjacency estimate as the source state of a rectified flow that learns to refine it toward the true graph distribution, recovering higher-order structural dependencies not captured by edge-wise predictions alone. 
The flow is trained on paired samples $(\bbA_0,\bbA_1)$ constructed from the same underlying graph, where $\bbA_0$ combines observed edges with the prior predictions on the masked region. 
At inference time, only $\bbA^{\ccalO}$ and $\xi$ are available: the prior produces the initial estimate, and the flow integrates the learned velocity field to generate the final graph reconstruction.

\subsection{Approximating the posterior mean}
\label{subsec:graphons-mmse}

As discussed in Section~\ref{subsec:dist-perc}, our goal is to approximate the conditional mean $\mathbb{E}[\bbA \mid \bbA^\ccalO]$ with a \textit{permutation equivariant} estimator. 
Before moving to particular parameterizations of the conditional mean, we introduce two assumptions.

{\assumption{We assume each edge in 
$\bbA \in \{0,1\}^{n \times n}$ follows a Bernoulli distribution whose 
probabilities depend on latent node variables 
$\bbz_1, \ldots, \bbz_n \in \ccalZ$ such that:
\begin{equation}
A_{ij} \;\sim\; \text{Bernoulli}\big(f(\bbz_i, \bbz_j)\big), 
\qquad 1 \leq i < j \leq n.
\end{equation}
The function $f$ maps 
pairs of latent variables to edge probabilities, i.e., $f(\bbz_i, \bbz_j) = P(A_{ij}|\bbz_i, \bbz_j) = p_{ij}$.}}

{\assumption{We assume that the edges are conditionally independent given the latent structure, i.e., given the latent structure $Z = \{\bbz_1,\ldots,\bbz_n\}$, we have:
\begin{equation}
P(\bbA \mid Z) = \prod_{1 \leq i < j \leq n} 
P(A_{ij} \mid \bbz_i, \bbz_j).
\end{equation}}}

Under these two assumptions, and assuming access to the mapping $\bbz^{-1}:\bbA \rightarrow \ccalZ$ that recovers the latent coordinates $Z = \bbz^{-1}(\bbA^{\ccalO})$, the posterior mean can be computed element-wise: on observed entries it equals the known value $A^{\ccalO}_{ij}$, while on masked entries $\left[\mathbb{E}[\bbA\mid\bbA^{\ccalO}]\right]_{ij} = P(A_{ij}=1 \mid \bbz_i, \bbz_j) = f(\bbz_i,\bbz_j)$.
% We adopt two different types of priors: $(i)$ inductive methods, represented by \textit{graphons}~\citep{graphon, avella2018centrality}, which are bounded, symmetric and measurable functions $\ccalW : [0,1]^2 \rightarrow [0,1]$, and \textit{GraphSAGE}~\citep{GraphSAGE10.5555/3294771.3294869}, a GNN-based estimator and $(ii)$ transductive ones, obtained from \textit{node2vec}~\citep{node2vec}, which provides an instance-level learned probabilistic model.

We adopt two different types of priors: $(i)$ inductive methods, represented by \textit{GraphSAGE}~\citep{GraphSAGE10.5555/3294771.3294869}, a GNN-based estimator, and $(ii)$ transductive ones, obtained from \textit{node2vec}~\citep{node2vec}, which provides an instance-level learned probabilistic model.

\paragraph{Posterior mean using inductive methods (dataset-informed).}

% A \emph{graphon}, defined as a symmetric function $\ccalW:[0,1]^2 \to [0,1]$, serves as a generative model for a family of graphs:
% \begin{align}
% \label{eq:stochastic_sampling}
%     z_i &\sim \operatorname{Uniform}[0,1], \quad i=1,\dots,n, \\ \nonumber
%     A_{ij} &\sim \operatorname{Bernoulli}\!\left(\ccalW(z_i,z_j)\right),
%     \quad 1 \leq i < j \leq n .
% \end{align}
% Graphons provide a functional representation of exchangeable random graphs where the conditional edge probability is $\left[\mathbb{E}[\bbA\mid \bbz]\right]_{ij} = \ccalW(z_i,z_j)$. This offers a natural, permutation-equivariant framework for estimating the posterior mean, though it requires access to the inverse mapping $z_i = [\bbz^{-1}(\bbA^{\ccalO})]_i$.
% Since $\ccalW$ is unknown, we estimate it using Scalable Implicit Graphon Learning (SIGL)~\citep{sigl}, which combines a graph neural network (GNN) encoder with an implicit neural representation (INR). SIGL operates in three steps: (1) a GNN-based sorting step to estimate latent node positions $\bbz$; (2) a histogram approximation of the sorted adjacency matrices; and (3) learning a graphon parameterization $f_{\phi}$ by minimizing its error against the histograms. A key feature of SIGL is its ability to recover the inverse mapping $\bbz^{-1}$, making it uniquely suitable for our model~\citep{ignr}. 

We approximate the posterior mean using a dataset-informed, inductive approach based on \emph{GraphSAGE}~\citep{GraphSAGE10.5555/3294771.3294869}. We train the model on the partially observed graphs in the dataset to produce node embeddings $\{\mathbf{z}_i\}_{i=1}^N$. From these embeddings, we train a single logistic predictor on Hadamard edge features ($\mathbf{z}_i \odot \mathbf{z}_j$) to estimate edge probabilities. The resulting conditional mean is parameterized as $\left[\mathbb{E}[\bbA \mid \bbA^{\ccalO}]\right]_{ij} \approx f_{\phi}(\mathbf{z}_i \odot \mathbf{z}_j)$.

% \hc{We also use a \emph{variational graph autoencoder (VGAE)}~\citep{kipf2016variational} 
% as a per-graph transductive prior. A GCN encoder maps the observed subgraph 
% to a Gaussian posterior over per-node latents $q(\mathbf{z}_i \mid \bbA^{\ccalO}) 
% = \mathcal{N}(\boldsymbol{\mu}_i, \mathrm{diag}(\boldsymbol{\sigma}_i^2))$, 
% and an inner-product decoder yields the conditional mean 
% $\left[\mathbb{E}[\bbA \mid \bbA^{\ccalO}]\right]_{ij} \approx 
% \sigma(\boldsymbol{\mu}_i^\top \boldsymbol{\mu}_j)$. Training maximizes the standard ELBO with a standard-normal latent prior.}

% \paragraph{Posterior mean using transductive methods (instance-specific).}
% For a transductive approach, we use \emph{node2vec} to learn an instance-specific embedding and predictor for each graph. Similar to the GraphSAGE method, we first train node2vec on a partially observed graph to obtain node embeddings $\{\mathbf{z}_i\}_{i=1}^N$. However, in contrast to the single predictor used for GraphSAGE, we fit a distinct, \emph{per-graph} logistic link predictor on Hadamard edge features with balanced negative sampling. This yields the same conditional mean parameterization, $\left[\mathbb{E}[\bbA \mid \bbA^{\ccalO}]\right]_{ij} = f_{\phi}(\mathbf{z}_i \odot \mathbf{z}_j)$, but with a predictor $f_{\phi}(\cdot)$ that is unique to each graph instance. At inference time, this instance-specific model is used to evaluate all masked pairs to compute the posterior mean.

\paragraph{Posterior mean using transductive methods (instance-specific).}
We use three per-graph priors that fit an embedding and a predictor specific to 
each graph instance from the partially observed graph $\bbA^{\ccalO}$. 
\emph{node2vec}~\citep{node2vec} learns embeddings $\{\mathbf{z}_i\}_{i=1}^N$ 
from biased random walks; a per-graph logistic predictor on Hadamard edge 
features then gives 
$\left[\mathbb{E}[\bbA \mid \bbA^{\ccalO}]\right]_{ij} \approx
f_{\phi}(\mathbf{z}_i \odot \mathbf{z}_j)$. 
\emph{VGAE}~\citep{kipf2016variational} uses a GCN encoder to produce a 
Gaussian posterior over per-node latents 
$q(\mathbf{z}_i \mid \bbA^{\ccalO}) = 
\mathcal{N}(\boldsymbol{\mu}_i, \mathrm{diag}(\boldsymbol{\sigma}_i^2))$ and 
an inner-product decoder, yielding 
$\left[\mathbb{E}[\bbA \mid \bbA^{\ccalO}]\right]_{ij} \approx 
\sigma(\boldsymbol{\mu}_i^\top \boldsymbol{\mu}_j)$, trained by ELBO 
maximization with a standard-normal prior. 
\emph{NCNC}~\citep{wangneural} is a neural common-neighbor link predictor 
that combines a GCN encoder with a pooled sum over softly-completed 
common-neighbor representations, giving 
$\left[\mathbb{E}[\bbA \mid \bbA^{\ccalO}]\right]_{ij} \approx
\sigma(\mathrm{NCNC}(i, j;\, \bbA^{\ccalO}, \bbX))$.

\subsection{Learning the flow model}
\label{subsec:flowmodel}

% \nz{We should check how the CE loss fits here.}

We now learn a flow model that approximates the joint distribution (coupling) \( p(\bbA, \hat{\bbA}^*) \) of the clean graph and its MMSE estimate.
As explained in Section~\ref{sec:background}, we need to specify the forward path $\bbA_t$ and the velocity field $v$.
Inspired by~\citet{ohayon2025posteriormean}, we incorporate prior information as the initialization of the forward path; with slight abuse of notation, 
we denote $f_{\text{prior}}$ as the prediction of the full graph 
(i.e., $f_{\text{prior}} (\bbA^{\ccalO}) \approx \mathbb{E}[\bbA \mid \bbA^{\ccalO}]$).
Specifically, we compute the sample $\bbA_0$ from the source distribution as follows:
\begin{equation}
    \bbA_0 = \bbA^{\ccalO}  + (1 - \xi) \odot \left(f_{\text{prior}}(\bbA^{\ccalO}) + \bbepsilon_s \right),
    \label{eq:A0_equation}
\end{equation}
where $\bbA$ is the ground-truth graph, $\xi$ is the corresponding mask (taking value $1$ for the observed pairs of nodes and $0$ otherwise), $\bbA^{\ccalO} = \xi \odot \bbA$ is the observed graph, and
$f_{\text{prior}}(\bbA^{\ccalO})$ is our approximate MMSE estimator for the masked edges.  
We also add a small amount of noise $\bbepsilon_s \sim \ccalN(0,\sigma_s^2)$ following~\citet{interpolant}.
We define $\bbA_1 = \bbA$ for the target distribution.

\paragraph{Distortion instantiations.}
Given the source distribution, we can learn the flow by minimizing a certain distortion $\Delta(\hbA_\theta, \bbA)$.
We consider two instances following the description in Section~\ref{subsec:dist-perc}: $(i)$ MSE loss, which boils down to the standard rectified flow-matching loss in~\eqref{eq:loss_flow}; and $(ii)$ CE loss, where the velocity field $v_\theta(\bbA_t,t)$ parameterizes a categorical distribution over edge values, similarly to~\citep{eijkelboom2024variational, dieleman2022continuous} and recent approaches that adapt continuous models to discrete data~\citep{lee2026flow, roos2026categorical}.

% , $\hat{y}_\theta = \sigma(v_\theta(\bbA_t, t))$,  $y = \bbA_1$, and $\Delta(\hat{y}, y)$ is an element-wise binary cross-entropy  evaluated on the upper triangle of the masked region. For our CE experiments we set $\sigma_s = 0$.
% \begin{itemize}[leftmargin=*]
%     \setlength\itemsep{-0.2em}
%     \item \textbf{Squared Frobenius (MSE):} 
%     $\hat{y}_\theta = v_\theta(\bbA_t, t)$, 
%     $y = \bbA_1 - \bbA_0$, and 
%     $\Delta(\hat{y}, y) = \|\hat{y} - y\|_F^2$. 
%     This gives the standard rectified flow-matching loss in~\eqref{eq:loss_flow}.

%     \item \textbf{Cross-entropy (CE):} This gives a formulation similar to~\citep{eijkelboom2024variational}, where $v_\theta(\bbA_t, t)$ outputs edge logits instead, $\hat{y}_\theta = \sigma(v_\theta(\bbA_t, t))$,  $y = \bbA_1$, and $\Delta(\hat{y}, y)$ is an element-wise binary cross-entropy  evaluated on the upper triangle of the masked region. For our CE experiments we set $\sigma_s = 0$.
%     % To address the heavy class imbalance of sparse adjacency matrices, we apply  a per-batch positive-class weight $w = \min(n_{\text{neg}}/n_{\text{pos}}, 50)$.  
% \end{itemize}

\paragraph{Training.}
For the velocity field $v$, we use the architecture from~\citet{jo2022score}, a GNN-based network that yields a permutation-equivariant parameterization (see Appendix~\ref{app:experimental_details} for details).
For each training graph $\bbA_1$, we sample a mask $\xi$ and time $t \sim U[0,1]$, construct $\bbA_0$ via Eq.~\eqref{eq:A0_equation}, and minimize the flow-matching objective with the chosen distortion $\Delta$ over the resulting paired samples  $(\bbA_0, \bbA_1)$.
The full training procedure is described in Alg.~\ref{alg:train&sample} (lines 1--6).

\paragraph{Inference.}
Given a new observed graph $\bbA^{\ccalO}$ and mask $\xi$, we initialize $\hat{\bbA}$ from the prior prediction and integrate the learned velocity field forward using $K$ Euler steps (Alg.~\ref{alg:train&sample}, lines 7--12).
Setting $K=1$ yields the lowest distortion and the highest AUC-ROC, while larger $K$ improves perceptual quality as discussed in Section~\ref{subsec:blind}.
The two distortion instantiations differ only in how the velocity field is obtained from the model. Under MSE, the network directly predicts the velocity field, which is integrated at each step. Under CE, the network predicts categorical distributions over edge values, and the corresponding velocity field is obtained from these probabilities before updating the observed entries (Alg.~\ref{alg:train&sample}, line~9).

\paragraph{Permutation invariance.}
Combined with the posterior-mean parameterization of Section~\ref{subsec:graphons-mmse}, the permutation-equivariant velocity field endows PIFM with two related symmetry guarantees.
\new{First, the reconstruction map is permutation-equivariant: relabeling the observation and mask $(\bbA^{\ccalO},\xi)$, with $\bbA^{\ccalO}=\xi\odot\bbA$ the observed part of the graph, produces a correspondingly relabeled reconstruction $\tilde{\bbA}_1$ (the sampler output at $t=1$). This holds for the full sampler of Alg.~\ref{alg:train&sample} (both the MSE and CE branches, for any number of steps $K$) and needs only equivariance and well-posedness of the dynamics; it is formalized in Theorem~\ref{thm:equiv}.
Second, in the idealized continuous-time flow the conditional density of the reconstruction is itself invariant under simultaneous relabeling of $(\tilde{\bbA}_1,\bbA^{\ccalO},\xi)$; this stronger statement is formalized in Proposition~\ref{prop:density}.}
Since graphs are exchangeable, the model's output should not depend on node ordering, making these symmetries a desirable inductive bias.\why{Split of the previous single theorem into two. The practically relevant claim is the equivariance of the reconstruction map (Theorem~\ref{thm:equiv}), which holds for the algorithm as implemented under both distortions; the CNF density-invariance statement (Proposition~\ref{prop:density}) is the stronger but more fragile claim and holds only for the idealized continuous projected MSE flow. The earlier single statement conflated the two and overpromised the density formula for the CE branch.}

\begin{theorem}[Permutation equivariance of the reconstruction]
\label{thm:equiv}
\new{Assume:
\begin{itemize}[leftmargin=*,itemsep=0pt,topsep=2pt]
\item[(a)] the prior $f_{\text{prior}}: \reals^{N\times N} \to \reals^{N\times N}$ and the velocity field $v_\theta: \reals^{N\times N}\times[0,1] \to \reals^{N\times N}$ are permutation-equivariant;
\item[(b)] $v_\theta$ is locally Lipschitz in its first argument (e.g., a smooth neural network on a bounded domain), so the sampling dynamics are well-posed.
\end{itemize}
For an observed graph $\bbA^{\ccalO}$ supported on the observed entries (i.e., $(1-\xi)\odot\bbA^{\ccalO}=0$) and mask $\xi$, let $\mathrm{PIFM}(\bbA^{\ccalO},\xi)$ be the reconstruction returned by Alg.~\ref{alg:train&sample} under either the MSE or the CE branch, for any number of Euler steps $K\ge 1$. Then, for every permutation matrix $\mathbf{P}$, the reconstruction is permutation-equivariant in distribution,
\begin{equation}
\mathrm{PIFM}(\mathbf{P}^\top\bbA^{\ccalO}\mathbf{P},\, \mathbf{P}^\top\xi\mathbf{P}) \;\stackrel{d}{=}\; \mathbf{P}^\top \mathrm{PIFM}(\bbA^{\ccalO},\xi)\, \mathbf{P},
\end{equation}
with equality holding pointwise either under the coupled noise draw $\mathbf{P}^\top\bbepsilon_s\mathbf{P}$ or when $\sigma_s=0$ (as in our CE experiments).}
\end{theorem}

\new{Theorem~\ref{thm:equiv} is the symmetry that matters in practice: it certifies that PIFM's reconstruction does not depend on node ordering for the algorithm exactly as implemented (both distortions, any $K$, including the sigmoid, clipping, and re-imposition steps of the CE branch), and it is independent of the choice of distortion $\Delta(\cdot,\cdot)$. In the idealized continuous projected flow (with a $C^1$, symmetry-preserving velocity and $\sigma_s>0$), this equivariance lifts to invariance of the reconstruction's conditional density on the masked edge coordinates; we state and prove this as Proposition~\ref{prop:density} in Appendix~\ref{app:perm-invariance-proof}.}

The proof is in Appendix~\ref{app:perm-invariance-proof}.
\new{Since the reconstruction task is exchangeable in node ordering (relabeling the input observation and mask should produce a correspondingly relabeled output), Theorem~\ref{thm:equiv} guarantees that PIFM has this inductive bias built in.}\why{Updated to point at the split theorems. The ``independent of the distortion'' note now lives next to Theorem~\ref{thm:equiv}, since it is true only for the equivariance result and not for the density formula (which excludes the CE branch).}

% \vspace{-0.1in}
\paragraph{Final algorithm.}
In Alg.~\ref{alg:train&sample}, we describe our training and sampling algorithms. In essence, PIFM is a general framework that learns a global graph structure to enhance simple, conditionally independent edge-wise priors.

\begin{figure*}[t]
\centering
\begin{subfigure}[t]{0.32\textwidth}
  \centering
  \includegraphics[width=\linewidth]{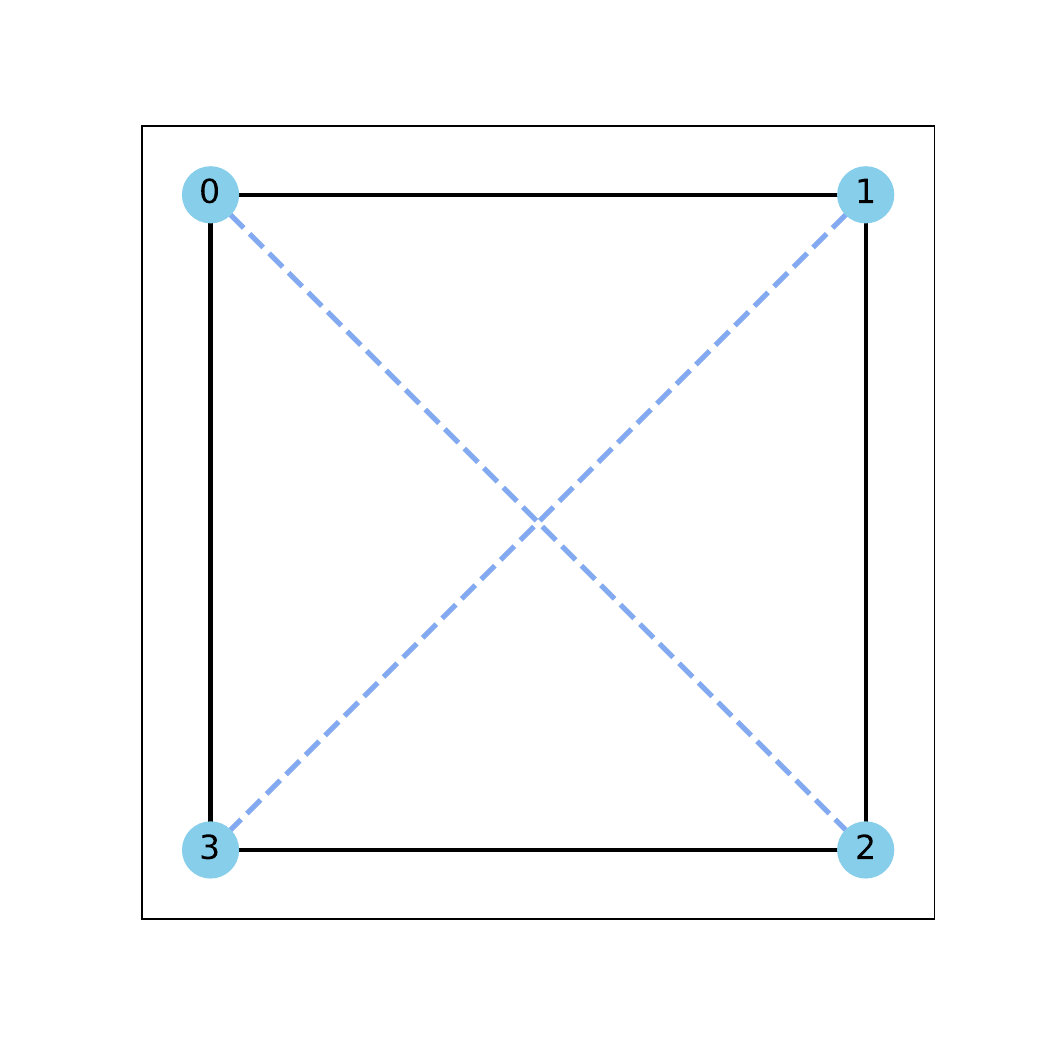}
  \caption{}
  \label{fig:graph_toy_exp}
\end{subfigure}\hfill
\begin{subfigure}[t]{0.32\textwidth}
  \centering
  \includegraphics[width=\linewidth]{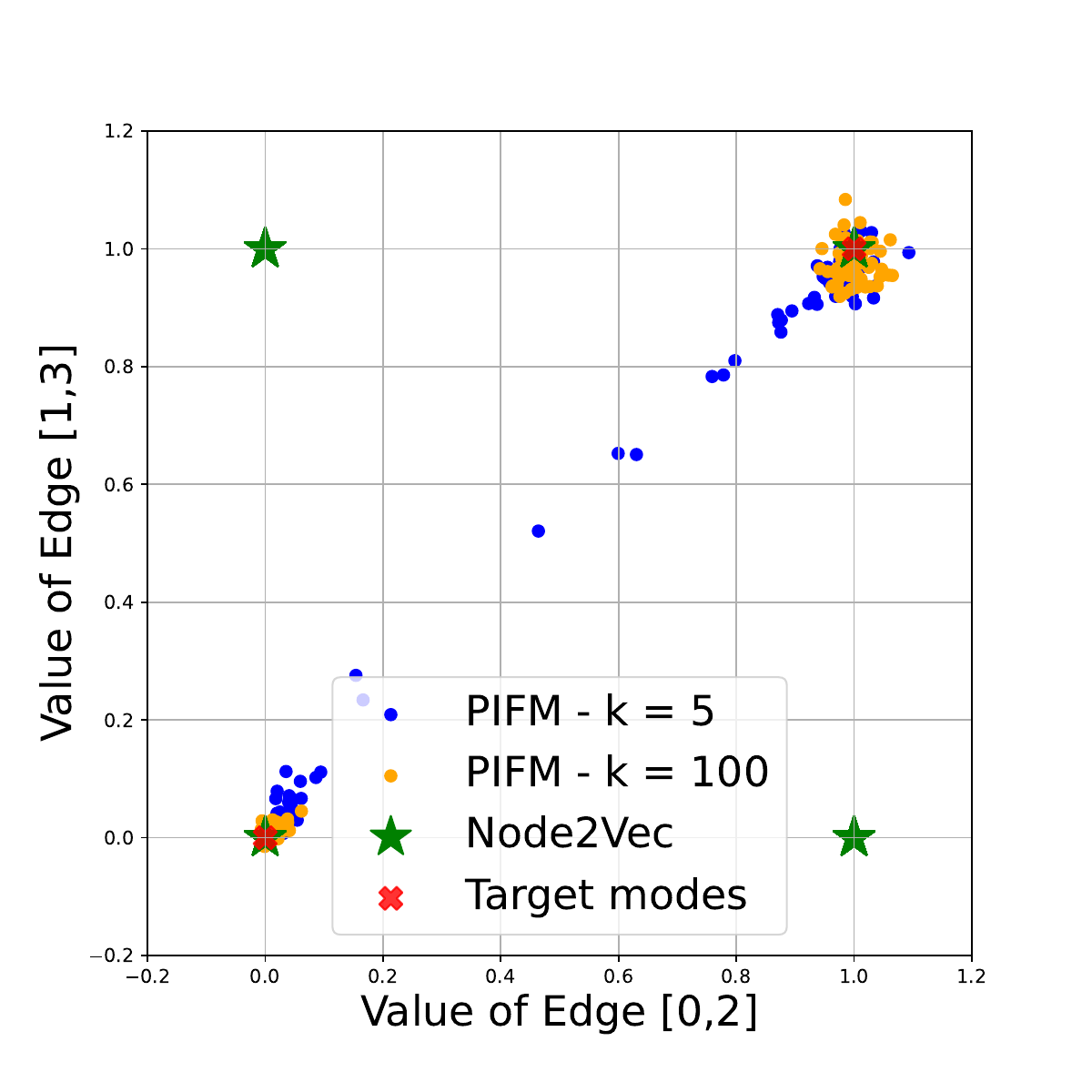}
  \caption{}
  \label{fig:results_toy_experiments}
\end{subfigure}\hfill
\begin{subfigure}[t]{0.32\textwidth}
  \centering
  \includegraphics[width=\linewidth]{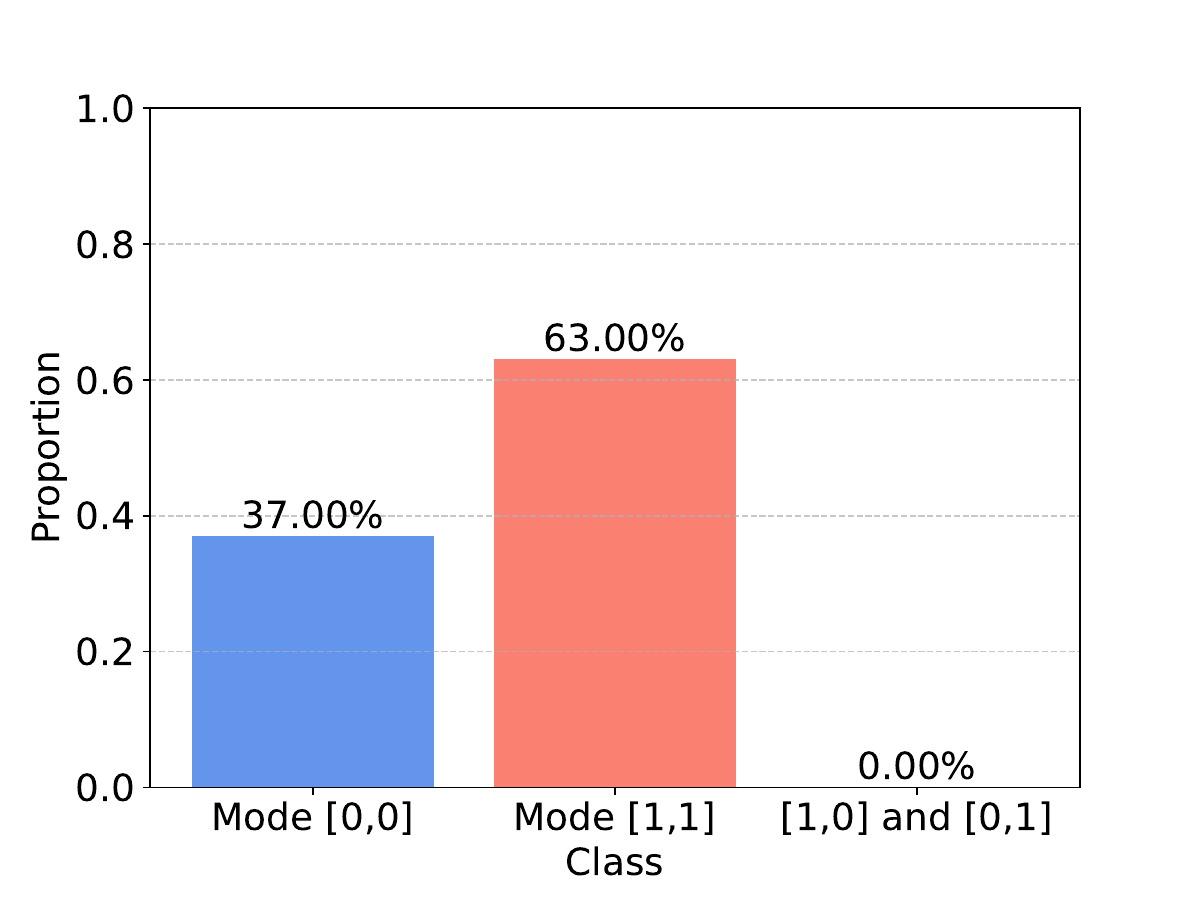}
  \caption{}
  \label{fig:proportions}
\end{subfigure}

\caption{\small Toy experiment showcasing the advantage of PIFM (in this case, for link prediction). a) Graph $\ccalG$ with four nodes, where the hidden edges are $e_{02}$ and $e_{13}$. b) Generated samples by using node2vec and PIFM (our proposed method): clearly, our method learns a probabilistic coupling, rendering a model that generates only the two valid modes. c) Proportions of samples generated with PIFM from each mode; remarkably, the method also learns a good approximation of the probability of each mode.}
\label{fig:toy_experiment}
\vspace{-0.15in}
\end{figure*}

To illustrate what we mean by \textit{learning a global and dependent predictor}, we now describe a toy experiment. Consider a four-node graph $\mathcal{G}$ (see Fig.~\ref{fig:toy_experiment} (a)) where the goal is to predict the diagonal edges under a specific constraint: the only valid outcomes are either both edges are present or both are absent, i.e., $\mathcal{E} = \{[e_{02} = 1, e_{13} = 1], [e_{02} = 0, e_{13} = 0]\}$.
Moreover, we assume that the probability of observing the first case is $0.6$, while the second one is $0.4$.

We first train an edge-wise prior using node2vec, which yields a probability of $0.6$ for each diagonal edge. 
Crucially, because node2vec models each edge prediction independently, this prior is misspecified. 
A standard predictor based on this prior would always predict $[1,1]$ if used as conditional mean or, if sampling were to be performed, could generate invalid predictions such as $[1,0]$.

We then train a flow model using this node2vec prior to construct the initial state $\mathbf{A}_0$ as in~\eqref{eq:A0_equation}.
After training (see Appendix~\ref{app:experimental_details} for details), we generate $200$ samples, illustrated in Fig.~\ref{fig:toy_experiment}(b); the proportion of each mode is shown in Fig.~\ref{fig:toy_experiment}(c).
The results clearly demonstrate that  the flow model $(i)$ successfully leverages global information, learning a \textit{probabilistic coupling} between the edges, to generate samples only from the two valid states, and $(ii)$ learns the probability of each mode.

\begin{algorithm}[t]
    \small
    \caption{Training and Sampling}\label{alg:train&sample}
    \begin{algorithmic}[1]
        \STATE \textbf{Training}
        \STATE Sample $\bbA_1 \sim p(\bbA)$, a mask $\xi$, and time $t \sim U[0,1]$.
        \STATE Train approximate MMSE estimator: $f_{\text{prior}}(\bbA^\ccalO)$
        \STATE Compute \\ $
        \bbA_0 = \bbA^{\ccalO} + (1-\xi)\odot \left(f_{\text{prior}}(\bbA^{\ccalO})
        + \bbepsilon_{s}\right), \quad
        \bbepsilon_{s} \sim \cN(0,\sigma_{s}^2)
        $
        \STATE Compute $\bbA_t = (1-t)\bbA_0 + t\bbA_1$.
        
        \STATE Train flow model: \\ 
        $\theta^* = \argmin_{\theta}\, \EE_{\bbA_1,\bbA_0,\xi,t}\!\big[\Delta\big(\hat{y}_\theta(\bbA_t,t),\ y(\bbA_0,\bbA_1)\big)\big]$ 
        \STATE \textbf{Sampling (Reconstruction)}
        \STATE 
        Initialize
        $\hat{\bbA} \gets \xi \odot \bbA^{\ccalO} + (1-\xi)\odot f_{\text{prior}}(\bbA^{\ccalO})
        + (1-\xi)\odot \bbepsilon_{s}, \quad
        \bbepsilon_{s} \sim \cN(0,\sigma_{\text{samp}}^2).
        $
        \FOR {$i \leftarrow 0, \dots, K-1$}
            \IF{MSE distortion}
                \STATE $\hat{\bbA} \leftarrow \hat{\bbA} + \frac{1}{K}\, v_{\theta^*}\!\left(\hat{\bbA}, \tfrac{i}{K}\right)$
            \ELSE
                \STATE \textit{// CE distortion}
                \STATE $\mu \leftarrow \sigma\!\left(v_{\theta^*}(\hat{\bbA}, \tfrac{i}{K})\right)$
                \STATE $\hat{\bbA} \leftarrow \mathrm{clip}\!\left(\hat{\bbA} + \tfrac{1}{K}\cdot\dfrac{\mu - \hat{\bbA}}{1 - i/K + \varepsilon},\; 0,\; 1\right)$
                \STATE $\hat{\bbA} \leftarrow \xi \odot \bbA^{\ccalO} + (1-\xi)\odot\hat{\bbA}$
            \ENDIF
        \ENDFOR
        \STATE $\hat{\bbA} \leftarrow \xi \odot \bbA^{\ccalO} + (1-\xi)\odot\hat{\bbA}$ \hfill\textit{// re-impose observed entries}
        \STATE Return $\hat{\bbA}$
    \end{algorithmic}
\end{algorithm}

% \newpage

% \vspace{-0.in}

% \paragraph{Computational cost.}
% We remark that the flow model training time of PIFM is the \textbf{same} as flow model training time of PIFM with Gaussian prior, meaning that there is no additional cost when training the flow model; the only difference comes from the additional training of the prior, if it is unavailable. The inference cost of PIFM scales linearly with the number of Euler steps $K$. 
% For the link prediction results in Table~\ref{tab:results_50_percent_mask_final}, we use $K=1$, so the flow refinement requires only one additional velocity evaluation after the prior prediction. 
% On an NVIDIA A100 80GB, the inference takes roughly 0.04--0.05s per graph, compared to 0.0002--0.002s for the standalone priors alone.
% When larger $K$ is used for perception-oriented sampling, inference becomes more expensive, with cost increasing approximately linearly in $K$ (e.g., ${\sim}2.8$s at $K{=}100$). 
% Detailed training and inference time measurements are reported in Appendix~\ref{app:training-time} and \ref{app:inference-time}.

\paragraph{Computational cost.}
We remark that the flow model training time of PIFM is the \textbf{same} as flow model training time of PIFM with Gaussian prior, meaning that there is no additional cost when training the flow model; the only added expense is training the prior when one is not already available. Inference is roughly 0.04--0.05s per graph at $K=1$ on an NVIDIA A100 (vs.\ 0.0002--0.002s for the priors alone) and scales linearly in $K$ (e.g., ${\sim}2.8$s at $K{=}100$); full timings are in Appendix~\ref{app:training-time} and~\ref{app:inference-time}.

\paragraph{Source distribution and prior model.}
PIFM is agnostic to the prior architecture: any model that 
produces edge-wise marginal probabilities on the masked region 
can serve as the source distribution.
In practice, we recommend validating candidate priors independently before flow training  and selecting the strongest available edge predictor for the 
target domain.

\section{Experiments}
\label{sec:numerical}

\subsection{Setup}
We evaluate our method on IMDB-B, PROTEINS, and ENZYMES~\citep{TUDataset}, three inductive graph datasets (training and testing on disjoint graphs), and on CORA~\citep{mccallum2000automating, yang2016revisiting}, which is transductive and large-scale, adding additional challenges for training the flow-matching module.
Thus, we focus on families of graphs that are diverse to show that our model learns a general predictor.
For the inductive datasets, each one is split into 85\% train, 10\% validation, and 5\% test graphs, and we evaluate 
reconstruction quality under two masking levels (10\% and 50\% of edges, with masks generated uniformly at random). 
The implementation details are provided in Appendix~\ref{app:experimental_details}.

% \vspace{-0.3cm}
\paragraph{Evaluation metrics.}
Performance is measured exclusively on masked edges. We report both threshold-dependent 
classification metrics (FPR, FNR) and threshold-independent metrics (ROC-AUC, AP). 
Threshold-dependent metrics are computed by binarizing predictions at a fixed cutoff of $0.5$, 
so entries with predicted probability $\geq 0.5$ are treated as edges and those $< 0.5$ as non-edges. 
In addition, we use maximum mean discrepancy (MMD)~\citep{o2021evaluation} to compute the distance between the generated graphs and the ground truth, serving as a proxy for computing the perception quality.
More details on these metrics are deferred to Appendix~\ref{app:metrics_calc}.
% \vspace{-0.2cm}
\paragraph{Baselines.}
We compare PIFM against several baselines, including diffusion-based. Recall that PIFM is composed of a one-shot prediction used as prior followed by a flow model. We compare PIFM to the accuracy of the one-shot prediction (without the flow) and with a flow with a random starting point:
\begin{itemize}[leftmargin=*]
    \setlength\itemsep{-0.2em}
    \item \textbf{Node2Vec Prior~\citep{node2vec}/GraphSAGE Prior~\citep{GraphSAGE10.5555/3294771.3294869}/NCNC Prior~\citep{wangneural}}: one-shot predictions using the structural prior directly.
    \item \textbf{Flow with Gaussian prior}: flow model initialized from Gaussian $\ccalN(0.5, 1)$ noise on masked entries. 
    \item \textbf{DiGress + RePaint~\citep{vignac2023digress}}: unconditional DiGress combined with RePaint-style resampling~\citep{lugmayr2022repaint}.
    \item \textbf{GDSS + RePaint~\citep{jo2022score}}: unconditional GDSS combined with RePaint-style resampling~\citep{lugmayr2022repaint}. 
    \item \textbf{VGAE}~\citep{kipf2016variational}: a graph autoencoder that learns latent node embeddings from the observed graph and reconstructs edges using a probabilistic decoder. 
\end{itemize}
% \vspace{-0.3cm}
Algorithmic details of the baselines are provided in 
Appendix~\ref{app:alg}. 
% {Lastly, we consider an additional experiment on a transductive case (CORA~\cite{yang2016revisiting}), where we compared with traditional baselines~\cite{li2023evaluating}.}

\subsection{Link Prediction}

\begin{table*}[h!]
\caption{
    \small{Graph reconstruction performance with \textbf{50\% of edges masked}. 
    We report results for PIFM trained with both the squared Frobenius distortion (MSE loss) 
    and the cross-entropy distortion (CE loss).
    We report AUC, Average Precision (AP$\uparrow$), False Positive Rate (FPR$\downarrow$), 
    and False Negative Rate (FNR$\downarrow$), all in percent (\%). 
    The best result for each metric is in \textbf{\textcolor{blue}{blue}} and the second best is \textcolor{green!50!black}{green}.}
}
\vspace{-0.1in}
\label{tab:results_50_percent_mask_final}
\centering
\resizebox{\textwidth}{!}{%
\begin{tabular}{l|cccc|cccc|cccc}
\toprule
\multicolumn{13}{c}{\textbf{Mask Rate: 50\% (0.5 Drop)}} \\
\midrule
& \multicolumn{4}{c|}{\textbf{ENZYMES}} & \multicolumn{4}{c|}{\textbf{PROTEINS}} & \multicolumn{4}{c}{\textbf{IMDB-B}} \\
\cmidrule(lr){2-5} \cmidrule(lr){6-9} \cmidrule(lr){10-13}
\textbf{Method} 
& AP $\uparrow$ & AUC $\uparrow$ & FNR $\downarrow$ & FPR $\downarrow$ 
& AP $\uparrow$ & AUC $\uparrow$ & FNR $\downarrow$ & FPR $\downarrow$ 
& AP $\uparrow$ & AUC $\uparrow$ & FNR $\downarrow$ & FPR $\downarrow$ \\
\midrule
\multicolumn{13}{l}{\textit{Baselines}} \\
\quad Node2Vec
& 19.14 & 55.22 & 46.37 & 44.29
& 23.51 & 53.83 & 51.37 & 44.36
& 54.20 & 52.22 & 48.41 & 47.51 \\
% \quad SIGL
% & 16.88 & 49.30 & 72.01 & 27.85
% & 22.90 & 52.55 & 100.00 & \textbf{\textcolor{blue}{0.00}}
% & 50.05 & 45.41 & 87.68 & 18.80 \\
\quad GraphSAGE
& 22.79 & 57.77 & 40.02 & 52.16
& 27.71 & 53.99 & \textcolor{green!50!black}{32.16} & 66.86
& 75.74 & 75.54 & 18.18 & 44.86 \\
\quad VGAE
& 25.35 & 63.50 & \textcolor{green!50!black}{36.45} & 41.24
& 32.22 & 64.20 & 39.42 & 41.90
& 67.37 & 67.30 & 39.07 & 35.98 \\
\quad NCNC
& 28.22 & 61.11 & 95.54 & 1.11
& 36.28 & 62.22 & 92.47 & \textcolor{green!50!black}{0.99}
& 80.49 & 74.85 & 74.65 & \textbf{\textcolor{blue}{1.70}} \\
\quad DiGress + RePaint
& 17.34 & 55.22 & 77.95 & 11.62
& 23.65 & 55.45 & 71.46 & 17.65
& 56.47 & 58.89 & 73.00 & 10.27 \\
\quad GDSS + RePaint
& 16.43 & 49.65 & 69.45 & 30.46
& 22.33 & 51.42 & 66.44 & 32.23
& 53.39 & 51.20 & 69.35 & 29.22 \\
\quad Flow w/ Gaussian prior
& 17.43 & 51.84 & 98.49 & \textcolor{green!50!black}{1.07}
& 26.40 & 55.55 & 93.21 & 5.25
& 78.72 & 79.76 & 41.56 & 14.62 \\
\midrule
\multicolumn{13}{l}{\textit{Ours (MSE loss)}} \\
\quad PIFM (Node2Vec)
& 22.95 & 59.14 & 90.71 & 3.53
& 27.57 & 59.68 & 87.05 & 8.98
& 84.46 & 85.71 & 32.95 & 15.03 \\
\quad PIFM (GraphSAGE)
& 25.44 & 61.36 & 95.62 & 1.86
& 35.50 & 60.61 & 85.05 & 10.23
& \textbf{\textcolor{blue}{93.13}} & \textbf{\textcolor{blue}{93.84}} & 17.52 & \textcolor{green!50!black}{7.61} \\
% \quad PIFM (SIGL)
% & 17.08 & 49.15 & 86.06 & 12.28
% & 28.38 & 59.58 & 61.20 & 20.38
% & 59.83 & 58.11 & 38.90 & 36.76 \\
\quad PIFM (VGAE)
& 27.47 & 65.69 & 98.23 & \textbf{\textcolor{blue}{0.85}}
& 35.42 & 67.43 & 82.37 & \textbf{\textcolor{blue}{0.81}}
& 89.39 & 87.10 & 19.43 & 14.98 \\
\quad PIFM (NCNC)
& 28.61 & 64.94 & 89.09 & 3.50
& \textbf{\textcolor{blue}{37.30}} & \textbf{\textcolor{blue}{68.56}} & 78.40 & 7.94
& 89.22 & 88.62 & 26.81 & 11.24 \\
\midrule
\multicolumn{13}{l}{\textit{Ours (CE loss)}} \\
\quad PIFM (GraphSAGE)
& 25.23 & 60.27 & 42.41 & 45.47
& 29.75 & 59.13 & 32.58 & 57.18
& 89.80 & \textcolor{green!50!black}{90.90} & \textbf{\textcolor{blue}{11.60}} & 22.10 \\
\quad PIFM (VGAE)
& \textcolor{green!50!black}{28.70} & \textcolor{green!50!black}{66.90} & \textbf{\textcolor{blue}{22.80}} & 59.50
& 35.10 & 65.80 & \textbf{\textcolor{blue}{5.90}} & 88.70
& \textcolor{green!50!black}{91.70} & 90.10 & \textcolor{green!50!black}{13.50} & 18.90 \\
\quad PIFM (NCNC)
& \textbf{\textcolor{blue}{29.21}} & \textbf{\textcolor{blue}{66.91}} & 55.96 & 20.70
& \textcolor{green!50!black}{37.12} & \textcolor{green!50!black}{68.12} & 44.03 & 27.39
& 88.81 & 87.34 & 19.62 & 17.78 \\
\bottomrule
\end{tabular}
}
\end{table*}

Table~\ref{tab:results_50_percent_mask_final} reports results for $50\%$ masking; $10\%$ results are deferred to Appendix~\ref{app:lp_10}. Overall, PIFM improves the AUC-ROC of all base priors (node2vec, GraphSAGE, VGAE, and NCNC), highlighting the benefit of the flow module in capturing global graph structure beyond local edge-wise predictions. Informed priors consistently outperform the Gaussian-prior flow, validating the use of prior predictions as informative initializations. We further observe that different priors dominate different graph regimes: PIFM(NCNC) achieves the strongest AUC on the sparse bioinformatics datasets (ENZYMES and PROTEINS), while PIFM(GraphSAGE) remains strongest on the dense social-network dataset (IMDB-B). The standalone NCNC prior is itself competitive on these benchmarks, but the flow refinement still yields additional gains.
% \coauthor{Two issues against Table~\ref{tab:results_50_percent_mask_final}. (1) The claims that PIFM ``improves the AUC-ROC of all base priors'' and that ``informed priors consistently outperform the Gaussian-prior flow'' are contradicted by SIGL: on ENZYMES, SIGL AUC ($49.30$) exceeds PIFM(SIGL) ($49.15$), and the Gaussian-prior flow beats PIFM(SIGL) on ENZYMES ($51.84$ vs $49.15$) and IMDB-B ($79.76$ vs $58.11$); the abstract's ``consistently enhances classical embeddings'' has the same counterexample. Please qualify these claims (e.g.\ ``improves most priors,'' noting SIGL is a weak prior here) or drop SIGL. (2) The best/second-best highlighting needs a full audit: at least one cell is mismarked (e.g.\ PROTEINS FNR marks $32.58$ as second-best though the GraphSAGE baseline has $32.16$).}

All PIFM results in this section use $K=1$, which yields the lowest distortion and highest AUC-ROC (consistent with Section~\ref{subsec:dist-perc}); perceptual quality improves with more integration steps as shown in Section~\ref{subsec:blind} and Appendix~\ref{app:ablation}. 
Even with $K=1$, PIFM outperforms the underlying priors despite their approximate MMSE optimality, indicating that the flow refinement captures global dependencies missed by the local predictors.

Notice we consider the two distortion instantiations introduced in Section~\ref{subsec:flowmodel}. 
While MSE and CE achieve comparable AUC-ROC, they exhibit different thresholding behavior on sparse graphs. In particular, CE significantly reduces the false negative rate (e.g., $95.62 \to 42.41$ on ENZYMES with GraphSAGE; $98.23 \to 22.80$ with VGAE), at the expense of a higher false positive rate. This behavior stems from the loss itself: MSE regression toward sparse binary targets tends to collapse predictions toward zero, pushing many true edges below the decision threshold, whereas CE with positive-class weighting yields better calibrated probabilities. As a result, MSE is preferable when prioritizing ranking metrics such as AUC/AP, while CE provides more balanced binary reconstructions on sparse datasets. Additional ablations are provided in Appendix~\ref{app:bce-ablation}. 

Finally, these experiments illustrate that the importance of the prior strongly depends on the difficulty of the reconstruction problem. The gap between PIFM and the Gaussian-prior baseline increases substantially with the masking rate (much larger at $50\%$ than at $10\%$ on IMDB), consistent with the standard ill-posedness principle that priors become more important as observations become scarcer (Appendix~\ref{app:transferability}). More expressive priors also lead to consistently stronger reconstructions, further highlighting the complementary role between local predictive representations and global flow-based refinement.

\paragraph{Large-scale graph reconstruction.}
We additionally evaluate PIFM on \textsc{Cora}.
Following~\cite{Limnios2023SaGess}, we use a subgraph-based training approach where each training example is constructed by sampling an edge-centered $k$-hop ego-subgraph (with a maximum node cap). 
Within the subgraph, we treat all other training edges as observed context and mask the centered edge as the supervision target. 
Then, PIFM is initialized using NCN~\citep{wangneural} as structural prior and trained
with the squared Frobenius distortion (MSE loss); experimental details are deferred to Appendix~\ref{app:cora_exp}.
As shown in Table~\ref{tab:cora_main}, PIFM outperforms the structural prior on held-out edges, albeit by a narrower margin than observed in the previous inductive datasets. This suggests that because the initial prior already achieves high performance on Cora, the marginal utility of the additional flow is somewhat diminished in this specific context.

\begin{table}[H]
\centering
% \vspace{-0.1cm}
\caption{\small{Link prediction on \textsc{Cora} (test edges split) with an NCN structural prior. We report AUC and Average Precision (AP) on the test edge split. Best results of each metrics are in \textbf{bold}. Refer to Table \ref{table:cora_experiment_appendix} for details about hyperparameters in each PIFM setting. PIFM uses MSE loss on \textsc{Cora} experiments. }}
\label{tab:cora_main}
\setlength{\tabcolsep}{4pt}
\renewcommand{\arraystretch}{1.05}
\scalebox{0.9}{
\begin{tabular}{lcccc}
\toprule
Method & AUC$\uparrow$ & AP$\uparrow$ & FPR$\downarrow$ & FNR$\downarrow$ \\
\midrule
NCN~\cite{wangneural}  & 93.78 & \textbf{93.89} & 23.50 & \textbf{6.71} \\
\midrule
PIFM (1-hop subgraphs) & \textbf{93.89} & 93.71 & \textbf{20.95} & 7.69 \\ 
PIFM (2-hop subgraphs) & 93.82 & 93.75 & 22.04 & 6.98 \\
\bottomrule
\end{tabular}}
\end{table}

\subsection{Blind graph reconstruction}
\label{subsec:blind}

We focus on two blind versions of link prediction, namely expansion and denoising.
In the expansion case, we only get to observe a subset of the edges (but no non-edges), and we need to determine which other entries correspond to existing edges.
Conversely, for denoising, we observe a set of edges that includes some spurious ones (but no confirmed non-edges), and we need to determine which of the observed edges are spurious and should be removed.
These cases are more challenging than link prediction since transductive priors like node2vec cannot be trained on the masked graphs (since we do not have positive and negative edges).
We report expansion here; denoising is in Appendix~\ref{app:denoising}.

% \hc{We focus on two blind versions of link prediction. \emph{Expansion} observes a subset of edges (no non-edges) and predicts the remaining edges; \emph{denoising} is the reverse: we observe a subset of non-edges (but no actual edge) and determine which other entries correspond to non-edges. Both preclude transductive priors that require both positive and negative training samples. We report expansion here; denoising is in Appendix~\ref{app:denoising}.}

% \vspace{-0.2in}
\paragraph{Expansion.} The goal in expansion is to predict a set of hidden edges $\ccalE_M$ given $\bbA^{\ccalO}$, such that the edge set of the ground truth is $\ccalE = \ccalE_M \cup\ccalE_O$.
Therefore, defining $\bbA_1 = \bbA$, the initialization becomes $\bbA_0 = \bbA^{\ccalO} + (1 - \bbA^{\ccalO}) \odot (f_{prior}(\bbA^{\ccalO}) + \bbepsilon_s)$. Throughout the blind-reconstruction experiments (expansion here and denoising in
Appendix~\ref{app:denoising}), we use PIFM with the MSE loss.
The results for a drop rate of 50\% are shown in Table~\ref{tab:results_expansion}.
Among all baselines, PIFM (GraphSAGE) attains the top AUC on all datasets and the top AP on PROTEINS and IMDB-B (remaining a close second to the GraphSAGE prior on ENZYMES AP), improving over the diffusion baselines.
Compared to a Gaussian start, the informed prior is crucial to improve AUC and AP, indicating effective global coupling beyond local scores.
Overall, PIFM serves as a better reconstructor in this challenging case, with $K$ providing a tunable perception–distortion trade-off (cf. Appendix~\ref{app:mmd}).

\begin{table*}[h!]
\caption{\small Performance for the \textbf{expansion} task with \textbf{50\% of edges masked (0.5 Drop)} (see Table~\ref{tab:results_10_percent_mask_final} for definitions). PIFM uses MSE loss for expansion tasks.}
\label{tab:results_expansion}
\centering
\small
\setlength{\tabcolsep}{3.5pt}
\renewcommand{\arraystretch}{1.05}

\resizebox{\textwidth}{!}{%
\begin{tabular}{lcccccccccccc}
\toprule
\multicolumn{13}{c}{\textbf{Mask Rate: 50\% (0.5 Drop)}} \\
\midrule
& \multicolumn{4}{c}{\textbf{ENZYMES}}
& \multicolumn{4}{c}{\textbf{PROTEINS}}
& \multicolumn{4}{c}{\textbf{IMDB-B}} \\
\cmidrule(lr){2-5} \cmidrule(lr){6-9} \cmidrule(lr){10-13}
\textbf{Method}
& AP $\uparrow$ & AUC $\uparrow$ & FNR $\downarrow$ & FPR $\downarrow$
& AP $\uparrow$ & AUC $\uparrow$ & FNR $\downarrow$ & FPR $\downarrow$
& AP $\uparrow$ & AUC $\uparrow$ & FNR $\downarrow$ & FPR $\downarrow$ \\
\midrule
\multicolumn{13}{l}{\textit{Baselines}} \\
GraphSAGE
& \textbf{\textcolor{blue}{13.95}} & \textcolor{green!50!black}{57.54} & \textbf{\textcolor{blue}{40.29}} & 52.74
& \textcolor{green!50!black}{18.91} & 53.91 & \textbf{\textcolor{blue}{31.62}} & 67.18
& \textcolor{green!50!black}{67.18} & \textcolor{green!50!black}{74.92} & \textbf{\textcolor{blue}{19.04}} & 45.20 \\
DiGress + RePaint
& 2.41 & 54.25 & 84.03 & \textcolor{green!50!black}{7.54}
& 4.52 & \textcolor{green!50!black}{56.87} & 81.10 & \textcolor{green!50!black}{13.00}
& 22.93 & 56.37 & 81.05 & \textbf{\textcolor{blue}{6.69}} \\
GDSS + RePaint
& 9.21 & 49.63 & \textcolor{green!50!black}{69.45} & 30.80
& 14.66 & 51.03 & \textcolor{green!50!black}{66.44} & 32.06
& 39.43 & 50.68 & 69.35 & 30.00 \\
Flow w/ Gaussian prior
& 9.45 & 50.40 & 90.64 & 9.35
& 14.71 & 50.31 & 82.32 & 17.28
& 49.46 & 62.28 & 71.41 & 13.64 \\
\midrule
\multicolumn{13}{l}{\textit{Ours}} \\
PIFM (GraphSAGE)
& \textcolor{green!50!black}{13.17} & \textbf{\textcolor{blue}{60.09}} & 100.00 & 0.00
& \textbf{\textcolor{blue}{21.70}} & \textbf{\textcolor{blue}{62.34}} & 94.75 & \textbf{\textcolor{blue}{4.54}}
& \textbf{\textcolor{blue}{83.49}} & \textbf{\textcolor{blue}{87.28}} & \textcolor{green!50!black}{29.74} & \textcolor{green!50!black}{11.27} \\
\bottomrule
\end{tabular}%
}
\end{table*}

% \vspace{-0.2in}
\subsection{Distortion-perception trade-off.} 
 While a single-step reconstruction ($K=1$) yields the lowest distortion and the highest AUC-ROC, we assess if more steps improve perceptual quality. To quantify perceptual quality, we adopt the standard graph-generation protocol of measuring the maximum mean discrepancy (MMD) between graph statistics of generated and reference graphs. 
 This protocol was used by GraphRNN~\citep{you2018graphrnn}, and it remains the common perception metric for both diffusion-based~\citep{jo2022score} and flow-based~\citep{qin2024defog} graph generators, which report MMD distance to measure how closely generated graphs match the data distribution. 
 Sweeping this metric as a function of $K$ therefore traces the perception axis of our distortion--perception trade-off.
 
\begin{figure*}[h!]
    \centering
    % --- First Subfigure ---
    \includegraphics[width=0.95\linewidth]{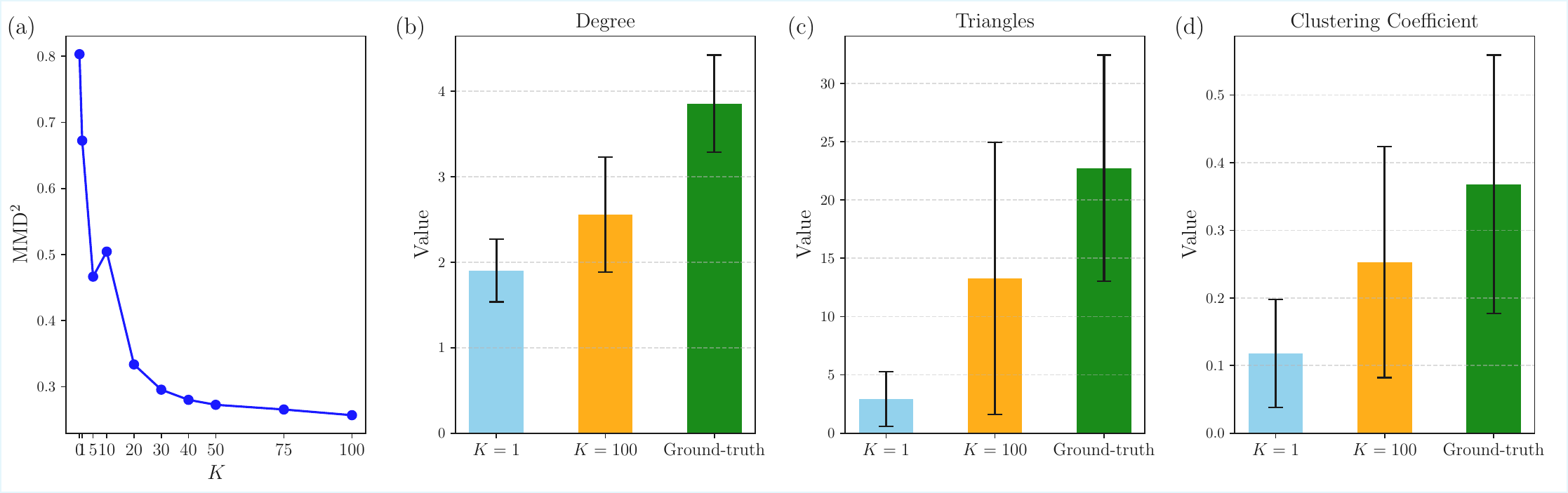}
    % --- Overall Figure Caption ---
    \vspace{-0.2in}
    \caption{\small {Evaluation of the distortion-perception trade-off. (a) The MMD score, measuring the distance to the true data distribution, decreases as $K$ increases. (b-d) This result is corroborated by key graph statistics, where the average degree, number of triangles, and clustering coefficient for graphs generated with $K=100$ more closely match the ground-truth distribution compared to those generated with $K=1$. 
    Error bars indicate the standard deviation over $300$ samples ($10$ samples for each of the $30$ test graphs).}}
    \label{fig:mmd-05-expansion}
    \vspace{-0.2in}
\end{figure*}

 We measure the MMD$^2$ score between the generated and ground-truth graph distributions on the ENZYMES dataset as a function of the number of steps, $K$.
As shown in Fig.~\ref{fig:mmd-05-expansion}(a), the MMD$^2$ score decreases as $K$ increases, signifying a closer match to the true data distribution and thus higher realism. 
We further validate this by comparing graph statistics (degree, triangles, clustering coefficients), which also show that a larger $K$ more closely matches the ground-truth. From the Distortion-perception trade-off perspective, decreasing MMD$^2$ indicates that the refinement stage moves the reconstruction closer to the clean graph distribution, and hence toward better perceptual quality.
Additional results and details are in Appendices~\ref{app:mmd} and \ref{app:raw_graphs}.

\section{Conclusions}
\label{sec:concl}

% In this paper, we introduced {Prior-Informed Flow Matching (PIFM)}, a method for graph reconstruction that learns global structural information by integrating local edge predictors within a flow-based generative model. 
% PIFM formulates graph topology inference as a distortion-perception problem, learning an optimal transport map from a local estimator to the ground-truth graph distribution.
% We evaluate PIFM using two types of local estimators, inductive (graphons and graphSAGE) and transductive (node2vec), which induce different reconstruction behaviors. 
% Experiments on multiple benchmark datasets show that PIFM consistently outperforms both classical embedding methods and recent flow-based baselines, demonstrating the significant value of learning global edge correlations.

% Our current formulation is limited to the homogeneous graph.
% Extending PIFM to heterogeneous graphs by defining the process in the probability simplex~\citep{eijkelboom2024variational} or using discrete flow models~\cite{qin2024defog} is another promising direction for future research.

In this paper, we introduced \textbf{Prior-Informed Flow Matching (PIFM)}, a flow-based 
estimator that learns global structural information by refining a local 
edge-prediction prior. PIFM is grounded in the distortion--perception 
framework: it transports a local posterior-mean estimate (obtained from 
GraphSAGE, node2vec, VGAE, or NCNC) toward the clean graph 
distribution, under either a mean squared error or cross-entropy distortion.
Experiments on diverse benchmarks show that PIFM consistently outperforms 
both classical embedding methods and recent diffusion/flow-based baselines. The current formulation handles only homogeneous graphs; extending it to 
heterogeneous data via simplex-based~\citep{eijkelboom2024variational} or 
discrete~\citep{qin2024defog} flow formulations is another promising direction for future research.

% \section*{Acknowledgement}
% This research was sponsored by the Army Research Office under Grant Number W911NF-17-S-0002 and by the National Science Foundation under awards CCF-2340481 and EF-2126387. 
% The views and conclusions contained in this document are those of the authors and should not be interpreted as representing the official policies, either expressed or implied, of the Army Research Office, the U.S. Army, or the U.S. Government. 
% The U.S. Government is authorized to reproduce and distribute reprints for Government purposes, notwithstanding any copyright notation herein.

\newpage
\section*{Impact statement}

This paper presents work whose goal is to advance the field of machine learning. There are many potential societal consequences of our work, none of which we feel must be specifically highlighted here.
\section*{Acknowledgement}
This research was sponsored by the Army Research Office under Grant Number W911NF-17-S-0002 and by the National Science Foundation under awards CCF-2340481 and EF-2126387. 
NZ was partially supported by a Ken Kennedy Institute 2024–25 Ken Kennedy-HPE Cray Graduate Fellowship.
The views and conclusions contained in this document are those of the authors and should not be interpreted as representing the official policies, either expressed or implied, of the Army Research Office, the U.S. Army, or the U.S. Government. 
The U.S. Government is authorized to reproduce and distribute reprints for Government purposes, notwithstanding any copyright notation herein.

\bibliography{citations}
\bibliographystyle{tmlr}
%%%%%%%%%%%%%%%%%%%%%%%%%%%%%%%%%%%%%%%%%%%%%%%%%%%%%%%%%%%%%%%%%%%%%%%%%%%%%%%
%%%%%%%%%%%%%%%%%%%%%%%%%%%%%%%%%%%%%%%%%%%%%%%%%%%%%%%%%%%%%%%%%%%%%%%%%%%%%%%
% APPENDIX
%%%%%%%%%%%%%%%%%%%%%%%%%%%%%%%%%%%%%%%%%%%%%%%%%%%%%%%%%%%%%%%%%%%%%%%%%%%%%%%
%%%%%%%%%%%%%%%%%%%%%%%%%%%%%%%%%%%%%%%%%%%%%%%%%%%%%%%%%%%%%%%%%%%%%%%%%%%%%%%
% \newpage
\appendix
\onecolumn % ICML appendix is usually one column

% % 1. Force LaTeX to start recording TOC entries again
\makeatletter
\addtocontents{toc}{\protect\setcounter{tocdepth}{3}} 
\renewcommand{\addcontentsline}[3]{%
  \addtocontents{#1}{\protect\contentsline{#2}{#3}{\thepage}{}}%
}
\makeatother

% % 2. Define the Index appearance
\etocsettocstyle{\section*{Contents}}{}
\etocsetnexttocdepth{subsection}

% 3. The Title
\section*{Appendix}
\addcontentsline{toc}{section}{Appendix}

% % 4. Generate the Index
% \localtableofcontents
% \vspace{1cm}
% \hrule
% \vspace{1cm}

% \section{Technical Appendices and Supplementary Material}
% Technical appendices with additional results, figures, graphs and proofs may be submitted with the paper submission before the full submission deadline (see above), or as a separate PDF in the ZIP file below before the supplementary material deadline. There is no page limit for the technical appendices.

\section{Algorithm}
\label{app:alg}

In this section, we describe the algorithms that we use as baselines.
Each method serves distinct purposes: the priors alone test whether the flow model provides meaningful improvement beyond the one-shot estimates given by the priors, uniform + flow evaluates whether the prior methods used (i.e. GraphSAGE/Node2Vec/VGAE/NCNC) are good structural priors for effective denoising, and DiGress + RePaint compares our model to standard modified unconditional generation models.

\subsection{Uniform + flow Baseline}

\label{app:uni+diffusion_alg}

This baseline ablates the structural prior by initializing the flow from a state where unknown entries are filled with uniform noise. The model then learns to denoise from this less-informed starting point.

\begin{algorithm}[h]
\small
\caption{Uniform + flow Training and Sampling}\label{alg:uniform_train&sample}
\begin{algorithmic}[1]

\STATE \textbf{Training}
\STATE Sample $\bbA_1$, a mask $\xi$, and time $t \sim U[0,1]$.

\STATE \parbox[t]{0.96\linewidth}{Define initial state with a uniform fill plus a small Gaussian perturbation on the masked region:
$\bbA_0 = \bbA^{\ccalO} + (1-\xi)\odot \mathcal{U}(0,1)^{N\times N}
+ (1-\xi)\odot \bbepsilon_{\text{train}}$, where
$\bbepsilon_{\text{train}} \sim \cN(0,\sigma_{\text{train}}^2)$.}

\STATE Define interpolant $\bbA_t = (1-t)\bbA_0 + t\bbA_1$.
\STATE Solve $\theta^* = \argmin_{\theta}\, \EE_{\bbA_1,\xi,t}\!\left\|\,v_\theta(\bbA_t,t) - (\bbA_1-\bbA_0)\,\right\|_F^2$.

\STATE \textbf{Sampling (Reconstruction)}
\STATE \parbox[t]{0.96\linewidth}{Given observed graph $\bbA^{\ccalO}$, define the initial state with masked noise:
$\hat{\bbA} \gets \xi \odot \bbA^{\ccalO} + (1-\xi)\odot \mathcal{U}(0,1)^{N\times N}
+ (1-\xi)\odot \bbepsilon_{\text{samp}}$, where
$\bbepsilon_{\text{samp}} \sim \cN(0,\sigma_{\text{samp}}^2)$.}

\FOR{$i \leftarrow 0$ {\bf to} $K-1$}
  \STATE $\hat{\bbA} \leftarrow \hat{\bbA} + \frac{1}{K}\, v_{\theta^*}\!\left(\hat{\bbA}, \tfrac{i}{K}\right)$
\ENDFOR
\STATE $\hat{\bbA} \leftarrow \xi \odot \bbA^{\ccalO} + (1-\xi)\odot\hat{\bbA}$ \hfill\textit{// re-impose observed entries}
\STATE \textbf{Return} $\hat{\bbA}$

\end{algorithmic}
\end{algorithm}

\subsection{DiGress + RePaint Baseline}
\label{app:digressRepaint_alg}

\paragraph{Training (Unconditional)}
The model $p_\theta$ is trained unconditionally on complete graphs $\bbA_0 \sim p_{\text{data}}$ to reverse a discrete forward noising process $q$. The forward process is a fixed Markov chain $q(\bbA_t|\bbA_{t-1})$ that corrupts the graph over $T$ steps. The training objective is to learn the denoising distribution $p_\theta(\bbA_0 | \bbA_t)$, modeled as a categorical prediction task for each node and edge.

\begin{algorithm}[H]
\caption{DiGress Unconditional Training}
\label{alg:digress-train-math}
\small
\begin{algorithmic}[1]
\STATE Sample $t \sim \mathcal{U}\{1,\ldots,T\}$ and $\bbA_t \sim q(\bbA_t \mid \bbA_0)$. \hfill\textit{/* Forward process */}
\STATE Predict $\bbA_0$ from $(\bbA_t,t)$ with $p_\theta(\cdot,t)$. \hfill\textit{/* Denoising objective */}
\STATE Minimize expected cross-entropy:
\vspace{-0.25em}
\[
\theta^{*} = \arg\min_{\theta}\;
\mathbb{E}_{\bbA_0 \sim p_{\text{data}},\, t \sim \mathcal{U}\{1,\ldots,T\}}
\big[\mathcal{L}_{\text{CE}}(\bbA_0,\; p_\theta(\bbA_t,t))\big].
\]
\vspace{-0.75em}
\end{algorithmic}
\end{algorithm}

\paragraph{Sampling (Conditional Reconstruction via RePaint)} At inference, given an observed graph $\bbA^{\ccalO} = \xi \odot \bbA_0$, the unconditionally trained model $p_{\theta^*}$ generates the missing entries. This is achieved by iteratively re-imposing the known (unmasked) information during the reverse diffusion process~\citep{lugmayr2022repaint}.

\begin{algorithm}[H]
\small
\caption{DiGress + RePaint Sampling}
\label{alg:digress-repaint-sample-math}
\begin{algorithmic}[1]
    \STATE \textbf{Input:} Observed graph $\bbA^{\ccalO}$, mask $\xi$, trained model $p_{\theta^*}$, steps $T$.
    \STATE \textbf{Output:} Reconstructed graph $\hat{\bbA}_0$.

    \STATE Initialize $\hat{\bbA}_T \sim p_{\text{prior}}(\cdot)$, where $p_{\text{prior}}$ is a random graph distribution.
    \FOR{$t = T, T-1, \dots, 1$}
        \STATE \textit{// Predict clean graph from current state}
        \STATE $\tilde{\bbA}_0 = p_{\theta^*}(\hat{\bbA}_t, t)$.
        \STATE
        \STATE \textit{// Impose known data by noising it to the next step}
        \STATE $\bbA_{t-1}^{\text{known}} \sim q(\bbA_{t-1} | \bbA^{\ccalO})$.
        \STATE
        \STATE \textit{// Sample the unknown region by noising the prediction to the next step}
        \STATE $\bbA_{t-1}^{\text{unknown}} \sim q(\bbA_{t-1} | \tilde{\bbA}_0)$.
        \STATE
        \STATE \textit{// Combine known and unknown parts for the next state}
        \STATE $\hat{\bbA}_{t-1} = \xi \odot \bbA_{t-1}^{\text{known}} + (1 - \xi) \odot \bbA_{t-1}^{\text{unknown}}$.
    \ENDFOR
    \STATE Return $\hat{\bbA}_0$
\end{algorithmic}
\end{algorithm}

% \newpage

\subsection{Node2Vec Prior (Per-Graph Classifier)}
\label{app:node2vec_prior}

This baseline learns a \emph{per-graph} edge-probability model from the observed subgraph.
We (i) fit node2vec embeddings on the observed topology and (ii) train a logistic classifier on Hadamard edge features to produce probabilities on the masked pairs.

\begin{algorithm}[t]
\small
\caption{Node2Vec Prior: Training and Inference}
\label{alg:node2vec_prior}
\begin{algorithmic}[1]
\setlength{\itemsep}{0pt}

\STATE \parbox[t]{0.96\linewidth}{\textbf{Inputs:} Full adjacency $\bbA_1$, mask $\xi$ ($\xi_{ij}=1$ if observed), Node2Vec hyperparams (dim $d$, walk length $L$, walks/node $R$, window $w$, $p,q$), negatives/positive ratio $k$.}
\STATE \parbox[t]{0.96\linewidth}{\textbf{Outputs:} Probabilities $\hat{P}$ on masked entries, i.e., $f_{\text{prior}}(\bbA^{\ccalO})$.}

\STATE \underline{\textbf{Training (per graph)}}
\STATE Construct observed graph $\bbA^{\ccalO} \leftarrow \xi \odot \bbA_1$.
\STATE Train Node2Vec on $\bbA^{\ccalO}$ to obtain embeddings $\{\mathbf{z}_i\}_{i=1}^N \in \mathbb{R}^d$.

\STATE \parbox[t]{0.96\linewidth}{Build labeled edge set on observed pairs ($i<j$):}
\STATE \parbox[t]{0.96\linewidth}{$\mathcal{P}^+ = \{(i,j): \xi_{ij}=1,\, A_{ij}=1\}$, \;
$\mathcal{P}^- \sim$ $k$-to-1 balanced samples from $\{(i,j): \xi_{ij}=1,\, A_{ij}=0\}$.}

\STATE \parbox[t]{0.96\linewidth}{Features: $\mathbf{x}_{ij} \leftarrow \mathbf{z}_i \odot \mathbf{z}_j$ (Hadamard); \;
Labels: $y_{ij}\in\{0,1\}$.}
\STATE Fit logistic classifier $g_\phi(\mathbf{x})=\sigma(\mathbf{w}^\top\mathbf{x}+b)$ (L2; class-balanced).

\STATE \underline{\textbf{Inference (per graph)}}
\STATE For each masked pair $(i,j)$ with $\xi_{ij}=0$, compute $\mathbf{x}_{ij} \leftarrow \mathbf{z}_i \odot \mathbf{z}_j$.
\STATE Predict $\hat{P}_{ij} \leftarrow g_\phi(\mathbf{x}_{ij})$ and set $\hat{P}_{ji} \leftarrow \hat{P}_{ij}$.
\STATE Return $\hat{P}$ as $f_{\text{prior}}(\bbA^{\ccalO})$ (used in Eq.~\eqref{eq:A0_equation}).
\end{algorithmic}
\end{algorithm}

\noindent\textit{Notes.} (i) We train embeddings \emph{only} on $\bbA^{\ccalO}$ to avoid leakage. 
(ii) The Hadamard feature works well and is symmetric; concatenation can be used but breaks symmetry unless sorted.
(iii) Thresholding at $0.5$ yields hard reconstructions; we use scores $\hat{P}$ directly in PIFM.

\section{Extended background}

\subsection{Graphons and graphon estimation}
\label{app:background_graphon}

% A graphon is defined as a bounded, symmetric, and measurable function $\ccalW : [0,1]^2 \rightarrow [0,1]$~\citep{graphon}.
% By construction, a graphon acts as a \emph{generative model for random graphs}, allowing the sampling of graphs that exhibit similar structural properties.
% To generate an undirected graph $\ccalG$ with $N$ nodes from a given graphon $\ccalW$, the process consists of two main steps: (1) assigning each node a latent variable drawn uniformly at random from the interval $[0,1]$, and (2) connecting each pair of nodes with a probability given by evaluating $\ccalW$ at their respective latent variable values.

A \emph{graphon}, defined as a symmetric function $\ccalW:[0,1]^2 \to [0,1]$, serves as a generative model for a family of graphs:
\begin{align}
\label{eq:stochastic_sampling}
    z_i &\sim \operatorname{Uniform}[0,1], \quad i=1,\dots,n, \\ \nonumber
    A_{ij} &\sim \operatorname{Bernoulli}\!\left(\ccalW(z_i,z_j)\right),
    \quad 1 \leq i < j \leq n .
\end{align}
Graphons provide a functional representation of exchangeable random graphs where the conditional edge probability is $\left[\mathbb{E}[\bbA\mid \bbz]\right]_{ij} = \ccalW(z_i,z_j)$. This offers a natural, permutation-equivariant framework for estimating the posterior mean, though it requires access to the inverse mapping $z_i = [\bbz^{-1}(\bbA^{\ccalO})]_i$.
Since $\ccalW$ is unknown, we estimate it using Scalable Implicit Graphon Learning (SIGL)~\citep{sigl}, which combines a graph neural network (GNN) encoder with an implicit neural representation (INR). SIGL operates in three steps: (1) a GNN-based sorting step to estimate latent node positions $\bbz$; (2) a histogram approximation of the sorted adjacency matrices; and (3) learning a graphon parameterization $f_{\phi}$ by minimizing its error against the histograms. A key feature of SIGL is its ability to recover the inverse mapping $\bbz^{-1}$, making it uniquely suitable for our model~\citep{ignr}. 

The generative process in~\eqref{eq:stochastic_sampling} can also be viewed in reverse: given a collection of graphs (represented by their adjacency matrices) \(\ccalD = \{\bbA_t\}_{t=1}^{M}\) that are sampled from an \emph{unknown} graphon \(\ccalW\), estimate \(\ccalW\).
Several methods have been proposed for this task~\citep{sas, sassbm, gwb, ignr, sigl}.
We focus on SIGL~\citep{sigl}, a resolution-free method that, in addition to estimating the graphon, also \emph{infers the latent variables $\bbeta$}.
This method parameterizes the graphon using an implicit neural representation (INR) \citep{siren}, a neural architecture defined as \(f_{\phi}(x, y): [0,1]^2 \rightarrow [0,1]\) where the inputs are coordinates from $[0,1]^2$ and the output approximates the graphon value $\ccalW$ at a particular position. 
In a nutshell, SIGL works in three steps: (1) a sorting step using a GNN \(g_{\phi'}(\bbA)\) that estimates the latent node positions or representations $\bbeta$; 
(2) a histogram approximation of the sorted adjacency matrices; 
and (3) learning the parameters $\phi$ by minimizing the mean squared error between $f_{\phi}(x, y)$ and the histograms (obtained in step 2).

\subsection{Node2vec}
\label{app:node2vec_background}

\textit{node2vec}~\citep{node2vec} is a scalable model for learning continuous node representations in graphs. This method is \textit{transductive}, meaning that it generates an embedding for each graph.
It extends the Skip-gram model from natural language processing to networks by sampling sequences of nodes through biased random walks. Node2vec introduces two hyperparameters $(p,q)$ that interpolate between breadth-first and depth-first exploration. This flexibility allows embeddings to capture both \emph{homophily} (nodes in the same community) and \emph{structural equivalence} (nodes with similar roles, e.g., hubs), which frequently coexist in real-world graphs.

The embeddings are learned via stochastic gradient descent with negative sampling to maximize the likelihood of preserving sampled neighborhoods. Once learned, node embeddings can be combined through simple binary operators (e.g., Hadamard product) to form edge features, enabling applications such as link prediction. Empirically, node2vec has been shown to outperform prior unsupervised embedding methods across tasks like classification and link recovery, while remaining computationally efficient and scalable to large graphs~\citep{node2vec}.

\subsection{GraphSAGE}

GraphSAGE~\citep{GraphSAGE10.5555/3294771.3294869} is an \textit{inductive} technique for link prediction based on graph neural networks (GNN) framework designed to generate embeddings for nodes in large, evolving graphs. 
It consists of two steps: for a target node, it first samples a fixed-size neighborhood of adjacent nodes, and then it aggregates feature information from these sampled neighbors.
By learning aggregation functions (such as a mean, pool, or LSTM aggregator) rather than embeddings for every single node, GraphSAGE can efficiently generate predictions for nodes that were not part of the training set, making it highly scalable and effective for real-world applications like social networks and recommendation systems.

\subsection{NCN and NCNC}
\label{app:ncnc_background}

\emph{Neural Common Neighbor} (NCN)~\citep{wangneural} is a link predictor that combines a message-passing encoder with a structural feature: node features $\{\mathbf{h}_i\}_{i=1}^N$ are produced by a single MPNN pass on the graph, and the score for an edge $(i,j)$ is a learned function of the Hadamard term $\mathbf{h}_i \odot \mathbf{h}_j$ and a pooled sum over common-neighbor representations $\sum_{u \in N(i) \cap N(j)} \mathbf{h}_u$. Because the MPNN runs only once per graph rather than once per target link, NCN is significantly more scalable than methods that apply MPNN to a subgraph for each candidate edge, while remaining expressive enough to distinguish links by the identity of their common neighbors.

\emph{NCN with Completion} (NCNC) extends NCN to address graph incompleteness: when the observed graph is partial, true common neighbors may be unobserved, distorting the structural feature on which NCN relies. NCNC softly completes the common-neighbor set by extending the pooled sum to all candidate intermediaries $u \in N(i) \cup N(j)$ and weighting each by a learned link-existence probability ($1$ if $u$ is already an observed common neighbor; the predicted probability of the unobserved endpoint-to-$u$ edge if exactly one endpoint connects to $u$; and $0$ otherwise). In our work we use the depth-1 NCNC predictor (a single round of soft completion before re-applying NCN), corresponding to the \texttt{IncompleteCN1Predictor} module of the official implementation.

\subsection{Graph Diffusion Models}
\label{app:background-diffusion}

Diffusion models are generative frameworks composed of two processes: a \textbf{forward process} that systematically adds noise to data until it becomes pure noise, and a \textbf{reverse process} that learns to reverse this, generating new data by starting from noise and progressively denoising it.
While these models exist for both discrete~\citep{vignac2023digress} and continuous domains~\citep{jo2022score}, we describe the continuous case which is the most related to our method. 
Here, a graph $\bbG_0$ is defined by its node features $\bbX_0 \in \mathbb{R}^{N\times F}$ and its weighted adjacency matrix $\bbA_0 \in \mathbb{R}^{N\times N}$. Following the GDSS framework, the forward process is described by a stochastic differential equation (SDE) that gradually perturbs the graph data over a time interval $t \in [0, T]$:
\begin{equation*}
    \text{d}\bbG_t = -\frac{1}{2}\beta(t)\bbG_t\,\mathrm{d}t + \sqrt{\beta(t)}\,\mathrm{d}\bbW_t
\end{equation*}
In this equation, $\bbW_t$ represents standard Brownian motion (i.e., noise), and $\beta(t)$ is a noise schedule that typically increases over time. This process is designed so that by the final time $T$, the noised graph $\bbG_T$ is indistinguishable from a standard Gaussian.

The generative reverse process is defined by another SDE that traces the path from noise back to data. This process relies on the \textbf{score function}, $\nabla_{\bbG_t}\log p(\bbG_t)$, which is the gradient of the log-density of the noisy data at time $t$. Since the true score function is unknown, it must be approximated. This is done using a neural network, or \textbf{score network}, which is trained to predict the score. For graphs, separate networks are often used for the adjacency matrix and node features: $\bbepsilon_{\bbtheta_A}(\bbA_t, t)$ and $\bbepsilon_{\bbtheta_X}(\bbX_t, t)$. These networks are trained by minimizing the denoising score-matching loss.

Once trained, these score networks can be plugged into the reverse SDE. New graphs are then generated by solving this SDE numerically using standard samplers like DDPM or DDIM.

\subsubsection{Diffusion-based inverse problems solver for graphs}
\label{app:background-diffusion-inverse}

We now expand on diffusion-based solvers for graph inverse problems.
Given a condition $\ccalC$ and a reward function $r(\bbG_0)$ that quantifies how close the sample $\bbG_0$ is to meeting $\ccalC$, the objective is to generate graphs $G_0$ that maximize the reward function.
From a Bayesian perspective, this problem boils down to sampling from the posterior $p(\bbG_0|\ccalC) \propto p(\ccalC|\bbG_0)p(\bbG_0)$ where $p(\ccalC|\bbG_0) \propto \exp{(r(\bbG_0))}$ is a likelihood term and $p(\bbG_0)$ is a prior given by the pre-trained diffusion model.
We now describe previous works for both differentiable and non-differentiable reward functions.

\paragraph{Guidance with Differentiable Reward Functions.}
Several approaches have been developed to guide generative models when the objective can be expressed as a \textbf{differentiable reward function}, particularly for inverse problems in imaging. These methods typically leverage the differentiability of the reward -- often a likelihood tied to a noisy measurement -- to calculate a \textit{conditional score} using Bayes' rule:
\begin{equation*}
    \nabla_{\bbG_t}\log p(\bbG_t|\ccalC) = \nabla_{\bbG_t}\log p(\ccalC|\bbG_t) + \nabla_{\bbG_t}\log p(\bbG_t)
\end{equation*}
In this formulation, the diffusion model naturally serves as the prior ($p(\bbG_t)$), while the likelihood term ($p(\ccalC|\bbG_t)$) provides the guidance. However, directly computing the score of the likelihood term is intractable because it requires integrating over all possible clean data: $p(\ccalC|\bbG_t) = \int p(\ccalC|\bbG_0)p(\bbG_0|\bbG_t)d\bbG_0$.

To overcome this, a common technique is to approximate the posterior distribution $p(\bbG_0|\bbG_t)$ with a Gaussian centered at the MMSE denoiser. This denoised estimate can be calculated efficiently using \textbf{Tweedie's formula}:
\begin{equation*}
    \mathbb{E}[\bbG_0|\bbG_t] = \frac{1}{\alpha_t}\left(\bbG_t + \sigma_t^2 \nabla_{\bbG_t} \log p(\bbG_t, t)\right)
\end{equation*}
While this framework is established for images, its application to graph-based inverse problems is less explored. This is primarily because most interesting properties and constraints in graphs are \textbf{not differentiable}. 
Some graph-specific methods, like DiGress~\citep{vignac2023digress}, implement guidance by training an auxiliary model, similar to classifier-free guidance, which introduces additional complexity.

\paragraph{Guidance with Non-Differentiable Reward Functions.}
For the more common scenario of non-differentiable constraints in graph generation, alternative strategies have emerged. 
The \textbf{PRODIGY} method, for instance, operates by repeatedly applying a two-step process at each denoising step: generation followed by projection.

First, it uses the unconditional diffusion model to produce a candidate sample $\hbG_{t-1}$. Second, it projects this candidate onto the set of valid solutions using a projection operator: $\Pi_{\ccalC}(\hbG_{t-1}) = \arg\min_{\bbZ \in \ccalC}\|\bbZ - \hbG_{t-1}\|_2^2$. Since applying the full projection at every step can destabilize the generation process, PRODIGY uses a partial update to balance constraint satisfaction with the learned diffusion trajectory:
\begin{equation*}
    \bbG_{t-1} \leftarrow (1-\gamma_t) \hbG_{t-1} + \gamma_t \Pi_{\ccalC}(\hbG_{t-1})
\end{equation*}
This approach has two main limitations. First, it is only practical for simple constraints where the projection operator $\Pi_{\ccalC}(\cdot)$ has an efficient, closed-form solution. Second, it applies the projection directly to the noisy intermediate sample $\bbG_{t-1}$, whereas the constraint $\ccalC$ is defined on the clean data $\bbG_0$, creating a domain mismatch.
Recently, in~\citet{tenorio2025graph}, the authors use zeroth-order optimization to build a guidance term, improving over PRODIGY in challenging tasks.

\subsubsection{Flow-based inverse solvers}

More recently, two flow-based generative models for graphs have been proposed.
Catflow, introduced in~\citet{eijkelboom2024variational}, formulates flow matching as a variational inference problem, allowing to build a model for categorical data.
The key difference between Catflow and traditional flow matching is that in the former, the objective is to approximate the posterior probability path, which is a distribution over possible end points of a trajectory. 
Compared to discrete diffusion, this formulation defines a path in the probability simplex, building a continuous path.
This formulation boils down to a cross-entropy loss.
Another recent work is DeFoG, introduced in~\citet{qin2024defog}.
This method is inspired by discrete flow matching~\citep{campbell2024generative}, where a discrete probability path is used.
Similarly, the loss is a cross-entropy.

\section{Proofs}

\subsection{Proof of Theorem~\ref{thm:equiv} and Proposition~\ref{prop:density}}
\label{app:perm-invariance-proof}

\begin{proposition}[Permutation invariance of the reconstruction density]
\label{prop:density}
\new{In addition to (a)--(b), assume:
\begin{itemize}[leftmargin=*,itemsep=0pt,topsep=2pt]
\item[(b$'$)] $v_\theta$ is continuously differentiable in its first argument (a sharpening of (b) needed for the change-of-variables Jacobian);
\item[(c)] $f_{\text{prior}}$ and $v_\theta(\cdot,t)$ map symmetric, zero-diagonal matrices to symmetric, zero-diagonal matrices, so the dynamics stay in the space of undirected graphs;
\item[(d)] the source noise is non-degenerate, $\sigma_s>0$.
\end{itemize}
The mask $\xi\in\{0,1\}^{N\times N}$ is symmetric with zero diagonal (it selects unordered edges), which is what lets $w_\theta=(1-\xi)\odot v_\theta$ preserve the symmetric, zero-diagonal subspace. We identify an undirected graph with its edge vector over the unordered pairs $\{i,j\}$, $i<j$; the masked edge coordinates are the pairs with $\xi_{ij}=0$, of which there are $m$. Define the \emph{projected} velocity field
\[
w_\theta(\bbA,t;\xi) = (1-\xi)\odot v_\theta(\bbA,t),
\]
which vanishes on observed coordinates by construction and is the idealized object the trained MSE field approximates. Define
\[
\begin{aligned}
\bbA_0 &= \bbA^{\ccalO} + (1-\xi) \odot \bigl(f_{\text{prior}}(\bbA^{\ccalO}) + \bbepsilon_s\bigr), \quad \bbepsilon_s \sim \mathcal{N}(0,\sigma_s^2 I_m) \text{ on masked edges}, \\
\bbA_t &\text{ solves the projected flow } \dot{\bbA}_t = w_\theta(\bbA_t,t;\xi),\ \bbA_t|_{t=0} = \bbA_0 \text{ (assumed to exist on } [0,1]\text{)}, \qquad \tilde{\bbA}_1 = \bbA_{t=1}.
\end{aligned}
\]
Then the conditional density of $\tilde{\bbA}_1$ on the masked edge coordinates, given by the CNF change-of-variables formula
\begin{equation}
\log p(\tilde{\bbA}_1 \mid \bbA^{\ccalO}, \xi) = \log p(\bbA_0 \mid \bbA^{\ccalO}, \xi) - \int_{0}^1 \tr_m\!\left(\frac{\partial w_\theta(\bbA_t,t;\xi)}{\partial \bbA_t}\right) dt,
\end{equation}
with $\tr_m$ the trace restricted to masked edge coordinates, is invariant under simultaneous relabeling: for every permutation matrix $\mathbf{P}$,
\begin{equation}
\log p(\mathbf{P}^\top\tilde{\bbA}_1\mathbf{P} \mid \mathbf{P}^\top \bbA^{\ccalO} \mathbf{P}, \mathbf{P}^\top \xi \mathbf{P}) = \log p(\tilde{\bbA}_1 \mid \bbA^{\ccalO}, \xi).
\end{equation}}\why{The idealized flow integrates the \emph{projected} field $w_\theta=(1-\xi)\odot v_\theta$, which our implemented MSE sampler does not. Flowing $w_\theta$ corresponds to the projected update $\hat{\bbA}\leftarrow \hat{\bbA} + \tfrac{1}{K}(1-\xi)\odot v_{\theta^*}(\hat{\bbA},i/K)$, whereas the MSE branch of Alg.~\ref{alg:train&sample} applies the unrestricted velocity $v_{\theta^*}$ to all coordinates, so observed entries drift during the flow. To enforce the observation constraints, the MSE sampler re-imposes the observed entries at the end (final line of Alg.~\ref{alg:train&sample}); this leaves the reported masked-region metrics unchanged, because the MSE training target $y=\bbA_1-\bbA_0$ is exactly zero on observed entries (there $\bbA_0=\bbA_1$), so $\xi\odot v_{\theta^*}\approx 0$ and the masked trajectory is essentially unaffected. The density statement of Proposition~\ref{prop:density} nonetheless describes the idealized projected flow $w_\theta$, since the trajectory still integrates the unrestricted $v_{\theta^*}$ rather than $w_\theta$ throughout. (The CE branch enforces the projection per step via its re-imposition, but its clipping makes the map non-invertible, so the density formula does not apply to CE either; CE is covered by Theorem~\ref{thm:equiv} only.)}
\end{proposition}

\begin{proof}
\new{Throughout, write $\bbA'^{\ccalO} = \mathbf{P}^\top \bbA^{\ccalO} \mathbf{P}$, $\xi' = \mathbf{P}^\top \xi \mathbf{P}$, and let
\[
\mu(\bbX,\eta) \;=\; \bbX + (1-\eta) \odot f_{\text{prior}}(\bbX)
\]
denote the deterministic center of the source state. For the density statement we work in undirected-edge coordinates: a symmetric, zero-diagonal matrix $\bbX$ is identified with its edge vector $\mathbf{e}(\bbX)\in\reals^{N(N-1)/2}$ over unordered pairs $\{i,j\}$, simultaneous relabeling $\bbX\mapsto\mathbf{P}^\top\bbX\mathbf{P}$ acts on $\mathbf{e}(\bbX)$ by an orthogonal permutation matrix $R_\mathbf{P}$ that maps masked edges to masked edges, $S_\xi\in\{0,1\}^{m\times N(N-1)/2}$ extracts the $m$ masked edge coordinates, and $\mathbf{M}_\xi = S_\xi^\top S_\xi$ is the corresponding diagonal projector. Assumption~(c) makes $w_\theta$ tangent to this subspace, so the projected flow remains in edge coordinates. We prove both results together: Steps~(i) and~(iii), together with the equivariance of the discrete sampler, establish the equivariance of the reconstruction (Theorem~\ref{thm:equiv}); Steps~(ii) and~(iv), combined with the change-of-variables conclusion, establish the density invariance (Proposition~\ref{prop:density}). The four steps are (i) equivariance of $\mu$; (ii) invariance of the conditional source density under simultaneous relabeling; (iii) equivariance of the ODE flow map; (iv) pointwise invariance of the masked-trace integrand.

\paragraph{Step (i): equivariance of the deterministic center $\mu$.} Using equivariance of $f_{\text{prior}}$ (assumption~(a)) together with the identity $(\mathbf{P}^\top X \mathbf{P}) \odot (\mathbf{P}^\top Y \mathbf{P}) = \mathbf{P}^\top (X \odot Y) \mathbf{P}$:
\begin{align*}
\mu(\bbA'^{\ccalO}, \xi')
&= \bbA'^{\ccalO} + (1-\xi') \odot f_{\text{prior}}(\bbA'^{\ccalO}) \\
&= \mathbf{P}^\top \bbA^{\ccalO} \mathbf{P} + (1 - \mathbf{P}^\top\xi\mathbf{P}) \odot f_{\text{prior}}(\mathbf{P}^\top \bbA^{\ccalO} \mathbf{P}) \\
&= \mathbf{P}^\top \bbA^{\ccalO} \mathbf{P} + \mathbf{P}^\top(1-\xi)\mathbf{P} \odot \mathbf{P}^\top f_{\text{prior}}(\bbA^{\ccalO}) \mathbf{P} \\
&= \mathbf{P}^\top \!\bigl[\bbA^{\ccalO} + (1-\xi) \odot f_{\text{prior}}(\bbA^{\ccalO})\bigr] \mathbf{P} \\
&= \mathbf{P}^\top \mu(\bbA^{\ccalO}, \xi) \mathbf{P}.
\end{align*}

\paragraph{Step (ii): invariance of the conditional source density.} In the masked edge coordinates the source is $\mathcal{N}\!\bigl(\mathbf{e}_\xi(\mu(\bbA^{\ccalO},\xi)),\,\sigma_s^2 I_m\bigr)$ on $\reals^m$, where $\mathbf{e}_\xi(\bbX) = S_\xi\,\mathbf{e}(\bbX)\in\reals^m$ extracts the masked edges (each edge counted once), so that
\[
p(\bbA_0 \mid \bbA^{\ccalO}, \xi) \;\propto\; \exp\!\left(-\tfrac{1}{2\sigma_s^2}\bigl\|\mathbf{e}_\xi\bigl(\bbA_0 - \mu(\bbA^{\ccalO},\xi)\bigr)\bigr\|_2^2\right).
\]
Evaluating at $\mathbf{P}^\top \bbA_0 \mathbf{P}$ under the permuted inputs $(\bbA'^{\ccalO},\xi')$ and using Step~(i): since relabeling maps masked edges to masked edges, $\mathbf{e}_{\xi'}(\mathbf{P}^\top\bbX\mathbf{P})$ is a permutation of the $m$ entries of $\mathbf{e}_\xi(\bbX)$, so its $\ell_2$ norm is unchanged,
\[
\bigl\|\mathbf{e}_{\xi'}\bigl(\mathbf{P}^\top \bbA_0 \mathbf{P} - \mu(\bbA'^{\ccalO},\xi')\bigr)\bigr\|_2^2
= \bigl\|\mathbf{e}_\xi\bigl(\bbA_0 - \mu(\bbA^{\ccalO},\xi)\bigr)\bigr\|_2^2.
\]
Hence
\[
p(\mathbf{P}^\top \bbA_0 \mathbf{P} \mid \bbA'^{\ccalO}, \xi') \;=\; p(\bbA_0 \mid \bbA^{\ccalO}, \xi).
\]

\paragraph{Step (iii): equivariance of the ODE flow map.} Let $g_\xi$ be a velocity field that is locally Lipschitz in $\bbA$ and satisfies the equivariance identity $\mathbf{P}^\top g_\xi(\bbA,t)\mathbf{P} = g_{\xi'}(\mathbf{P}^\top\bbA\mathbf{P},t)$. This covers $g_\xi=v_\theta$ (the implemented continuous sampler, which does not depend on $\xi$, so $g_{\xi'}=v_\theta$) and $g_\xi=w_\theta(\cdot,\cdot;\xi)$ (the projected flow of Proposition~\ref{prop:density}, for which the identity follows from equivariance of $v_\theta$ and $1-\xi'=\mathbf{P}^\top(1-\xi)\mathbf{P}$). Let $\Phi_t^\xi(\bbA_0)$ solve $\dot{\bbA}_t = g_\xi(\bbA_t,t)$ from $\bbA_0$, and set $\tilde{\bbA}_t := \mathbf{P}^\top \Phi_t^\xi(\bbA_0) \mathbf{P}$. Differentiating in $t$ and using the equivariance identity,
\[
\dot{\tilde{\bbA}}_t \;=\; \mathbf{P}^\top g_\xi(\Phi_t^\xi(\bbA_0), t) \mathbf{P} \;=\; g_{\xi'}(\mathbf{P}^\top \Phi_t^\xi(\bbA_0)\mathbf{P}, t) \;=\; g_{\xi'}(\tilde{\bbA}_t, t),
\]
with initial condition $\tilde{\bbA}_0 = \mathbf{P}^\top \bbA_0 \mathbf{P}$. By Picard-Lindel\"of (assumption~(b)), the $\xi'$-trajectory starting at $\mathbf{P}^\top \bbA_0 \mathbf{P}$ is exactly $\tilde{\bbA}_t$, so
\[
\Phi_t^{\xi'}(\mathbf{P}^\top \bbA_0 \mathbf{P}) \;=\; \mathbf{P}^\top \Phi_t^\xi(\bbA_0) \mathbf{P} \qquad \text{for all } t \in [0,1].
\]
In particular, the inverse flow satisfies $(\Phi_1^{\xi'})^{-1}(\mathbf{P}^\top \tilde{\bbA}_1 \mathbf{P}) = \mathbf{P}^\top \bbA_0 \mathbf{P}$.

\new{\paragraph{Equivariance of the reconstruction (Theorem~\ref{thm:equiv}).} Step~(i) gives equivariance of the deterministic center $\mu$; since $\bbepsilon_s$ is isotropic Gaussian, $\mathbf{P}^\top\bbepsilon_s\mathbf{P} \stackrel{d}{=} \bbepsilon_s$, so the source state obeys $\bbA_0(\bbA'^{\ccalO},\xi') \stackrel{d}{=} \mathbf{P}^\top \bbA_0(\bbA^{\ccalO},\xi)\mathbf{P}$ (pointwise under the coupled draw $\mathbf{P}^\top\bbepsilon_s\mathbf{P}$, or when $\sigma_s=0$). Step~(iii) gives equivariance of the continuous flow map. The remaining operations of the discrete sampler in Alg.~\ref{alg:train&sample} act entrywise with shared parameters (the Euler increment, the sigmoid $\sigma(\cdot)$, the clip to $[0,1]$, and the re-imposition $\xi\odot\bbA^{\ccalO}+(1-\xi)\odot\hat{\bbA}$ of the CE branch), so each commutes with the relabeling $X\mapsto\mathbf{P}^\top X\mathbf{P}$. Composing an equivariant source with the equivariant $K$-step sampler yields $\mathrm{PIFM}(\bbA'^{\ccalO},\xi') \stackrel{d}{=} \mathbf{P}^\top\mathrm{PIFM}(\bbA^{\ccalO},\xi)\mathbf{P}$, proving Theorem~\ref{thm:equiv}. This uses only assumptions (a)--(b) and holds for both distortions and every $K$.}

\paragraph{Step (iv): pointwise invariance of the masked-trace integrand.} Work in the edge coordinates of the preamble, on which simultaneous relabeling acts by the orthogonal permutation $R_\mathbf{P}$ (so $\mathbf{e}(\bbA'_t) = R_\mathbf{P}\,\mathbf{e}(\bbA_t)$). Let $J_\xi(\bbA_t)$ be the edge-coordinate Jacobian of the projected field $w_\theta(\cdot,t;\xi)$. Equivariance of $w_\theta$ and chain-rule differentiation give
\[
J_{\xi'}(\bbA'_t) \;=\; R_\mathbf{P}\,J_\xi(\bbA_t)\,R_\mathbf{P}^\top.
\]
Under simultaneous relabeling the masked-edge set maps to itself, so $\mathbf{M}_{\xi'} = R_\mathbf{P}\,\mathbf{M}_\xi\,R_\mathbf{P}^\top$. By cyclic trace invariance and $R_\mathbf{P}^\top R_\mathbf{P} = \mathbf{I}$,
\[
\tr_m\!\left(\tfrac{\partial w_\theta(\bbA'_t,t;\xi')}{\partial \bbA'_t}\right)
= \tr(\mathbf{M}_{\xi'} J_{\xi'}(\bbA'_t))
= \tr(R_\mathbf{P}\,\mathbf{M}_\xi\,J_\xi(\bbA_t)\,R_\mathbf{P}^\top)
= \tr(\mathbf{M}_\xi J_\xi(\bbA_t))
= \tr_m\!\left(\tfrac{\partial w_\theta(\bbA_t,t;\xi)}{\partial \bbA_t}\right).
\]
The integrand is pointwise invariant, hence so is its integral over $t$.

\paragraph{Conclusion (density invariance, Proposition~\ref{prop:density}).} Applying the CNF formula at the permuted endpoint $\mathbf{P}^\top \tilde{\bbA}_1 \mathbf{P}$, and chaining Steps~(ii)--(iv):
\[
\begin{aligned}
\log p(\mathbf{P}^\top \tilde{\bbA}_1 \mathbf{P} \mid \bbA'^{\ccalO}, \xi')
&= \log p\!\bigl((\Phi_1^{\xi'})^{-1}(\mathbf{P}^\top \tilde{\bbA}_1 \mathbf{P}) \,\big|\, \bbA'^{\ccalO}, \xi'\bigr)
- \int_0^1 \tr_m\!\left(\tfrac{\partial w_\theta(\bbA'_t,t;\xi')}{\partial \bbA'_t}\right) dt \\
&\stackrel{\text{(iii)}}{=} \log p(\mathbf{P}^\top \bbA_0 \mathbf{P} \mid \bbA'^{\ccalO}, \xi')
- \int_0^1 \tr_m\!\left(\tfrac{\partial w_\theta(\bbA'_t,t;\xi')}{\partial \bbA'_t}\right) dt \\
&\stackrel{\text{(ii)}}{=} \log p(\bbA_0 \mid \bbA^{\ccalO}, \xi)
- \int_0^1 \tr_m\!\left(\tfrac{\partial w_\theta(\bbA'_t,t;\xi')}{\partial \bbA'_t}\right) dt \\
&\stackrel{\text{(iv)}}{=} \log p(\bbA_0 \mid \bbA^{\ccalO}, \xi)
- \int_0^1 \tr_m\!\left(\tfrac{\partial w_\theta(\bbA_t,t;\xi)}{\partial \bbA_t}\right) dt \\
&= \log p(\tilde{\bbA}_1 \mid \bbA^{\ccalO}, \xi).
\end{aligned}
\]}\why{Proof reorganized for the theorem split and tightened for the density result. Steps~(i) and~(iii) plus the discrete-sampler argument prove the equivariance result (Theorem~\ref{thm:equiv}), which holds for the implemented algorithm under both distortions; Steps~(ii) and~(iv) plus this conclusion prove the density-invariance result (Proposition~\ref{prop:density}), scoped to the idealized projected flow $\dot{\bbA}_t = w_\theta(\bbA_t,t;\xi)$ with $\sigma_s>0$ and a $C^1$ velocity. The source density, the divergence trace, and the permutation action are all expressed in undirected-edge coordinates over unordered pairs $\{i,j\}$, where the Gaussian source is non-degenerate (each edge counted once) and the induced permutation $R_\mathbf{P}$ is orthogonal. This corrects the earlier draft, which used the full Frobenius norm (double-counting edges) and incorrectly treated the upper-triangular representation as permutation-invariant. Step~(iii) is stated for a generic equivariant field so that it applies both to $v_\theta$ (implemented sampler) and to $w_\theta$ (projected flow).}
\end{proof}

\new{\paragraph{Remark (which priors satisfy assumption~(a)).}
SIGL, GraphSAGE, VGAE, and NCNC are permutation-equivariant by construction: each is built from equivariant GNN layers, and the edge-prediction head reduces to a symmetric function of node embeddings (Hadamard product, inner product, or common-neighbor pooling). When inference is stochastic (VGAE latent sampling, GraphSAGE neighborhood sampling), this equivariance holds in distribution, under the coupled randomness obtained by relabeling the sampling seeds together with the nodes; it holds pointwise only when inference is fully deterministic (e.g., full-neighborhood aggregation with a deterministic readout and no neighborhood sampling), since a deterministic final readout alone does not remove randomness from the encoder. Node2vec is not strictly equivariant: it produces stochastic embeddings via SGD, and the PCA-based canonicalization we use to fix the embedding ordering relies on conventions (sign-fixing, tie-breaking under repeated eigenvalues) that are not in general permutation-invariant. For Node2vec, Theorem~\ref{thm:equiv} and Proposition~\ref{prop:density} should therefore be read as approximate guarantees that hold when the canonicalization is itself approximately invariant; the strictly equivariant priors give the unconditional guarantee.}\why{Per Finding~6: the previous in-proof claim that ``Node2vec enforces equivariance via PCA sign-fixing'' overpromised, since rules like ``first nonzero entry positive'' depend on node ordering. Honest demotion: the four GNN-based priors satisfy assumption~(a) strictly; Node2vec satisfies it only approximately, so the theorem's guarantee is correspondingly weaker for Node2vec. The per-prior list now lives outside the proof body so the proof itself is purely about the theorem hypothesis.}

\section{Experimental details}
\label{app:experimental_details}

\subsection{Details about the architecture}
\label{app:architecture}

Our model adopts a modified version of the adjacency score network architecture introduced in GDSS~\citep{jo2022score}.
The network is a permutation-equivariant graph neural network designed to approximate the scores $\nabla_{\bbA_t} \log p_t(X_t, \bbA_t)$ and $\nabla_{\bbx_t} \log p_t(\bbx_t, \bbA_t)$ at each diffusion step; in this paper, we use only score w.r.t. $\bbA_t$.
Concretely, the architecture consists of stacked message-passing layers followed by a multi-layer perceptron. Each layer propagates node and edge information through adjacency-based aggregation, ensuring equivariance under node relabeling. Time information $t$ is incorporated by scaling intermediate activations with the variance of the forward diffusion process, following the practice in continuous-time score models. Residual connections and normalization layers are used to stabilize training. 
The final output is an $N \times N$ tensor matching the adjacency dimension. This design provides the required permutation-equivariance and expressive power while remaining computationally tractable for mid-sized benchmark graphs.

The modification that incorporates is a module to build an embedding for the variable $t$ and a FiLM style modulation to incorporate noise conditioning.
In particular, we incorporate the following modules:

\begin{itemize}
    \item A positional encoding based on a sinusoidal embedding following~\citet{karras2022elucidating}
    \item An MLP layer with SiLU activation per attention layer
    \item A modulation at each attention layer, where we scale the hidden features by an adaptive RMS norm operation~\citep{crowson2024scalable}
\end{itemize}

\subsection{Details about the datasets}

In Table~\ref{tab:dataset_stats} we report the statistics of the datasets used in the main text.

\begin{table}[H]
\centering
\small
\caption{Statistics of the datasets used for evaluation.}
\label{tab:dataset_stats}
\begin{tabular}{lrrrrl}
\toprule
\textbf{Dataset} & \textbf{\# Graphs} & \textbf{Avg. Nodes} & \textbf{Avg. Edges} & \textbf{\# Classes} & \textbf{Domain} \\
\midrule
ENZYMES  & 600   & 32.63 & 62.14 & 6 & Bioinformatics \\
PROTEINS & 1{,}113 & 39.06 & 72.82 & 2 & Bioinformatics \\
IMDB-B   & 1{,}000 & 19.77 & 96.53 & 2 & Social Network \\
\bottomrule
\end{tabular}
\end{table}

\subsection{Hyperparameters}

\subsubsection{Flow-based baselines}
\label{app:flow_baselines_hparams}

We report the hyperparameters of our model, both the architecture and training. All three baselines use the same rectified-flow architecture and optimizer family; the only substantive differences are the prior settings.

\begin{table}[H]
\centering
\small
\begin{tabular}{l l}
\toprule
\multicolumn{2}{l}{\textbf{PIFM - Link Prediction, 10\% masked}} \\
\midrule
\texttt{batch\_size} & 64 (IMDB-B \& ENZYMES), 32 (PROTEINS) \\
\texttt{optimizer} & Adam \\
\texttt{learning\_rate} & 2e-4 \\
\texttt{dropout} & 0.2 \\
\texttt{hidden\_dim} & 32 \\
\texttt{num\_layers} & 5 \\
\texttt{num\_linears} & 2 \\
\texttt{channels} & \{c\_init: 2,\; c\_hid: 8,\; c\_final: 4\} \\
\texttt{train/val/test\_noise\_std} & 0.05 (IMDB-B \& PROTEINS), 0.1 (ENZYMES)\\
\texttt{ode\_steps} (Euler, $K$) & 1 to 100 \\
\texttt{prior} & GraphSAGE (default hyperparameters) \\
\midrule
\end{tabular}
\end{table}

\begin{table}[H]
\centering
\small
\begin{tabular}{l l}
\toprule
\multicolumn{2}{l}{\textbf{PIFM - Link Prediction, 50\% masked}} \\
\midrule
\texttt{batch\_size} & 64 (IMDB-B \& ENZYMES), 32 (PROTEINS) \\
\texttt{optimizer} & Adam \\
\texttt{learning\_rate} & 2e-4 \\
\texttt{dropout} & 0.2 \\
\texttt{hidden\_dim} & 32 \\
\texttt{num\_layers} & 5 \\
\texttt{num\_linears} & 2 \\
\texttt{channels} & \{c\_init: 2,\; c\_hid: 8,\; c\_final: 4\} \\
\texttt{train/val/test\_noise\_std} & 0.05 (IMDB-B \& PROTEINS), 0.1 (ENZYMES)\\
\texttt{ode\_steps} (Euler, $K$) & 1 to 100 \\
\texttt{prior} & GraphSAGE (default hyperparameters) \\
\midrule
\end{tabular}
\end{table}

\begin{table}[H]
\centering
\small
\begin{tabular}{l l}
\toprule
\multicolumn{2}{l}{\textbf{PIFM (VGAE) - Link Prediction, 50\% masked}} \\
\midrule
\texttt{batch\_size} & 64 (IMDB-B \& ENZYMES), 32 (PROTEINS) \\
\texttt{optimizer} & Adam \\
\texttt{learning\_rate} & 2e-4 \\
\texttt{dropout} & 0.2 \\
\texttt{hidden\_dim} & 32 \\
\texttt{num\_layers} & 5 \\
\texttt{num\_linears} & 2 \\
\texttt{channels} & \{c\_init: 2,\; c\_hid: 8,\; c\_final: 4\} \\
\texttt{train/val/test\_noise\_std} & 0.01 (IMDB-B), 0.00 (PROTEINS, ENZYMES) \\
\texttt{ode\_steps} (Euler, $K$) & 1 to 100 \\
\texttt{prior} & VGAE (epochs: 200, lr: 0.01, hidden: 32/16) \\
\bottomrule
\end{tabular}
\end{table}

\begin{table}[H]
\centering
\small
\begin{tabular}{l l}
\toprule
\multicolumn{2}{l}{\textbf{PIFM (NCNC) - Link Prediction, 50\% masked}} \\
\midrule
\texttt{batch\_size} & 64 (IMDB-B \& ENZYMES), 32 (PROTEINS) \\
\texttt{optimizer} & Adam \\
\texttt{learning\_rate} & 2e-4 \\
\texttt{dropout} & 0.2 \\
\texttt{hidden\_dim} & 32 \\
\texttt{num\_layers} & 5 \\
\texttt{num\_linears} & 2 \\
\texttt{channels} & \{c\_init: 2,\; c\_hid: 8,\; c\_final: 4\} \\
\texttt{train/val/test\_noise\_std} & 0.05 (IMDB-B), 0.01 (ENZYMES, PROTEINS) \\
\texttt{ode\_steps} (Euler, $K$) & 1 to 100 \\
\texttt{prior} & NCNC (default hyperparameters) \\
\bottomrule
\end{tabular}
\end{table}

\begin{table}[H]
\centering
\small
\begin{tabular}{l l}
\toprule
\textbf{Flow w/ Gaussian Prior} & \textbf{Value} \\
\midrule
\texttt{epochs} & 1000 \\
\texttt{batch\_size} & 64 (default), 32 (PROTEINS) \\
\texttt{optimizer} & Adam \\
\texttt{learning\_rate} & 2e-4 \\
\texttt{dropout} & 0.2 \\
\texttt{hidden\_dim} & 32 \\
\texttt{num\_layers} & 5 \\
\texttt{num\_linears} & 2 \\
\texttt{channels} & \{c\_init: 2,\; c\_hid: 8,\; c\_final: 4\} \\
\texttt{train\_noise\_std} (masked $t{=}0$) & 0.00 \\
\texttt{val\_noise\_std} (masked $t{=}0$) & 0.00 \\
\texttt{ode\_steps} (Euler, $K$) & 1 to 100 \\
\texttt{prior} & None (masked entries initialized from $\mathcal{N}(0.5,1)$) \\
\bottomrule
\end{tabular}
\end{table}

\subsection{Metrics Calculation}
\label{app:metrics_calc}

We evaluate performance only on the set of masked (unknown) edges in the upper triangle 
of the adjacency matrix. For each test graph, all metrics are computed on these entries 
and then averaged across graphs.

\paragraph{Metrics Used in Tables}
We report the following four metrics in the main results:

\begin{itemize}[leftmargin=*]

    \item \textbf{Area Under the ROC Curve (AUC).}  
    Computed on the raw predicted scores (when available). AUC measures the probability 
    that a randomly chosen true edge receives a higher predicted score than a randomly 
    chosen non-edge. Larger AUC indicates stronger ranking performance.

    \item \textbf{Average Precision (AP).}  
    Computed from the precision–recall curve induced by ranking the predictions. 
    AP summarizes how well the model recovers true edges across all possible thresholds, 
    with higher values indicating better precision–recall trade-offs.

    \item \textbf{False Positive Rate (FPR).}  
    After thresholding predictions at $0.5$, the FPR is defined as
    \[
    \mathrm{FPR} = \frac{\mathrm{FP}}{\mathrm{FP}+\mathrm{TN}},
    \]

    \item \textbf{False Negative Rate (FNR).}  
    After thresholding predictions at $0.5$, the FNR is defined as
    \[
    \mathrm{FNR} = \frac{\mathrm{FN}}{\mathrm{FN}+\mathrm{TP}},
    \]

    \item \textbf{MMD.} A kernel-based method that measures the difference between two probability distributions by embedding them in a feature space and finding the maximum difference between their mean embeddings.
\end{itemize}

\noindent AUC and AP are threshold-independent metrics (computed directly on the provided scores), 
while FPR and FNR are threshold-dependent error rates (obtained after binarizing 
at $0.5$). All values reported in the tables are averaged over test graphs and expressed in percentage.

\subsection{Training cost}
\label{app:training-time}

All models are trained on NVIDIA A100-SXM4-80GB GPUs. Once the prior is fixed, the flow model used in PIFM has essentially the same training loop and computational cost as the flow model initialized with a Gaussian prior. Thus, the main additional cost of PIFM comes from fitting the prior itself, not from the subsequent flow-matching stage. This observation is consistent with our empirical training runs, where we did not observe noticeable differences in optimization behavior between Gaussian-prior and prior-informed flow variants across edge-drop settings. Table~\ref{tab:training_time_100epochs} reports training time of the flow model for $100$ epochs on each dataset, with PROTEINS taking the most time while the other 2 datasets remain similar training costs.

\begin{table}[H]
\centering
\small
\caption{Training time of the flow model for $100$ epochs on each dataset under 50\% mask rate.}
\label{tab:training_time_100epochs}
\begin{tabular}{lr}
\toprule
\textbf{Dataset} & \textbf{Training Time (s)} \\
\midrule
\textsc{ENZYMES}  & 206.4 \\
\textsc{PROTEINS} & 646.7 \\
\textsc{IMDB-B}   & 243.7 \\
\bottomrule
\end{tabular}
\end{table}

\subsection{Inference cost}
\label{app:inference-time}

Since PIFM reconstructs graphs by integrating a learned velocity field over $K$ Euler steps, its runtime grows approximately linearly with the number of Euler steps integrated. We observe a cost of about $0.03$ seconds per step per graph, so increasing $K$ improves refinement quality at the expense of proportionally higher latency.

Table~\ref{tab:inference_steps_scaling} reports the per-graph inference time of PIFM as a function of the number of Euler steps. The results show near-linear scaling, from approximately $0.03$ seconds in the lowest-step setting to about $2.8$ seconds when using $100$ steps.

\begin{table}[H]
\centering
\small
\caption{Per-graph inference time of PIFM as a function of the number of Euler steps.}
\label{tab:inference_steps_scaling}
\begin{tabular}{lc}
\toprule
\textbf{Setting} & \textbf{Inference Time per Graph (s)} \\
\midrule
$K=1$          & 0.03 \\
$K=10$           & 0.28 \\
$K=50$           & 1.40 \\
$K=100$                & 2.80 \\
\bottomrule
\end{tabular}
\end{table}

To contextualize these costs, Table \ref{tab:inference_baselines_compare} compares PIFM against the standalone prior predictors on each dataset. As expected, the one-shot priors are faster because they do not require iterative refinement. However, PIFM with $K=1$ remains relatively inexpensive while retaining smallest reconstruction error, with per-graph inference times below $0.06$ seconds on all three benchmarks. This makes the single-step setting attractive in regimes where distortion is the main objective, while larger values of $K$ are better suited to settings where perceptual refinement is more important.

\begin{table}[H]
\centering
\small
\caption{Per-graph inference time comparison between standalone priors and PIFM.}
\label{tab:inference_baselines_compare}
\begin{tabular}{lccc}
\toprule
\textbf{Method} & \textbf{PROTEINS (s)} & \textbf{ENZYMES (s)} & \textbf{IMDB-B (s)} \\
\midrule
Node2Vec Prior   & 0.0016 & 0.0020 & 0.0010 \\
GraphSAGE Prior  & 0.0003 & 0.0003 & 0.0002 \\
PIFM ($K=1$)     & 0.0425 & 0.0532 & 0.0440 \\
PIFM ($K=100$)   & 2.8207 & 2.8419 & 2.8660 \\
\bottomrule
\end{tabular}
\end{table}

\newpage
\section{Additional results}

\subsection{Link prediction with 10 \% of edges masked}
\label{app:lp_10}

\begin{table}[H]
\caption{
    \small{Graph reconstruction performance with \textbf{10\% of edges masked (0.1 Drop)} (see Table~\ref{tab:results_50_percent_mask_final} for definitions).
   }
}
\vspace{-0.1in}
\label{tab:results_10_percent_mask_final}
\centering
\resizebox{\textwidth}{!}{%
\begin{tabular}{l|cccc|cccc|cccc}
\toprule
\multicolumn{13}{c}{\textbf{Mask Rate: 10\% (0.1 Drop)}} \\
\midrule
& \multicolumn{4}{c|}{\textbf{ENZYMES}} 
& \multicolumn{4}{c|}{\textbf{PROTEINS}} 
& \multicolumn{4}{c}{\textbf{IMDB-B}} \\
\cmidrule(lr){2-5} \cmidrule(lr){6-9} \cmidrule(lr){10-13}
\textbf{Method} 
& AP $\uparrow$ & AUC $\uparrow$ & FNR $\downarrow$ & FPR $\downarrow$ 
& AP $\uparrow$ & AUC $\uparrow$ & FNR $\downarrow$ & FPR $\downarrow$ 
& AP $\uparrow$ & AUC $\uparrow$ & FNR $\downarrow$ & FPR $\downarrow$ \\
\midrule
\multicolumn{13}{l}{\textit{Baselines}} \\
Node2Vec              & 24.62 & 59.60 & \textcolor{green!50!black}{51.39} & 37.96 & 33.24 & 64.40 & 48.56 & 35.37 & 65.00 & 56.36 & 50.27 & 41.68 \\
GraphSAGE             & 41.28 & 73.70 & \textbf{\textcolor{blue}{13.49}} & 60.59 & 46.36 & 74.58 & \textbf{\textcolor{blue}{11.00}} & 63.50 & 83.55 & 83.26 & 16.42 & 36.89 \\
DiGress + RePaint     & 33.39 & 67.86 & 58.92 & 5.19 & 40.34 & 72.39 & \textcolor{green!50!black}{47.82} & 6.00 & 59.25 & 58.63 & 76.44 & 7.68 \\
GDSS + RePaint        & 18.35 & 47.04 & 74.31 & 32.19 & 26.96 & 51.39 & 63.07 & 32.09 & 57.89 & 46.11 & 69.75 & 36.17 \\
Flow w/ Gaussian prior& 40.09 & 72.44 & 71.03 & 5.87 & \textcolor{green!50!black}{57.86} & 80.83 & 65.09 & \textcolor{green!50!black}{3.07} & \textcolor{green!50!black}{98.89} & \textcolor{green!50!black}{98.37} & 2.26 & \textbf{\textcolor{blue}{2.54}} \\
\midrule
\multicolumn{13}{l}{\textit{Ours}} \\
PIFM (Node2Vec)       & \textcolor{green!50!black}{41.67} & \textcolor{green!50!black}{76.86} & 72.09 & \textcolor{green!50!black}{5.11} & \textbf{\textcolor{blue}{58.25}} & \textbf{\textcolor{blue}{81.74}} & 59.37 & 6.34 & 97.60 & 97.28 & \textbf{\textcolor{blue}{1.37}} & 3.77 \\
PIFM (GraphSAGE)      & \textbf{\textcolor{blue}{47.21}} & \textbf{\textcolor{blue}{80.25}} & 72.85 & \textbf{\textcolor{blue}{2.40}} & 54.79 & \textcolor{green!50!black}{81.02} & 55.73 & 5.40 & \textbf{\textcolor{blue}{99.37}} & \textbf{\textcolor{blue}{98.79}} & \textcolor{green!50!black}{1.81} & \textcolor{green!50!black}{3.37} \\
\bottomrule
\end{tabular}
}
\end{table}

\subsection{PIFM with a graphon-based prior (SIGL)}
\label{app:sigl_prior_appendix}

The priors used in the main text (GraphSAGE, node2vec, VGAE, NCNC) are all
embedding-based link predictors. 
To test whether PIFM is agnostic to the \emph{family} of structural prior, we additionally evaluate a graphon-based prior: Scalable Implicit Graphon Learning (SIGL)~\citep{sigl}, which fits a graphon to the training graphs and predicts masked edges from the learned latent node coordinates
(background in Appendix~\ref{app:background_graphon}). Unlike the main-text priors, SIGL is a generative model of an exchangeable graph family rather than a discriminative edge predictor, making it a useful test on PIFM's
prior-agnosticism.

\begin{table}[H]
\centering
\small
\caption{\small PIFM with a graphon-based prior (SIGL) on link prediction, at 10\%
and 50\% edge drop. AP, AUC, FNR, FPR in \% (see
Table~\ref{tab:results_50_percent_mask_final} for definitions). \textbf{Bold} marks
the better of the two rows per ranking metric (AP, AUC) within each block.}
\label{tab:sigl_prior}
\resizebox{\textwidth}{!}{%
\begin{tabular}{l|cccc|cccc|cccc}
\toprule
& \multicolumn{4}{c|}{\textbf{ENZYMES}} & \multicolumn{4}{c|}{\textbf{PROTEINS}} & \multicolumn{4}{c}{\textbf{IMDB-B}} \\
\cmidrule(lr){2-5} \cmidrule(lr){6-9} \cmidrule(lr){10-13}
\textbf{Method}
& AP $\uparrow$ & AUC $\uparrow$ & FNR $\downarrow$ & FPR $\downarrow$
& AP $\uparrow$ & AUC $\uparrow$ & FNR $\downarrow$ & FPR $\downarrow$
& AP $\uparrow$ & AUC $\uparrow$ & FNR $\downarrow$ & FPR $\downarrow$ \\
\midrule
\multicolumn{13}{l}{\textit{Mask rate: 10\% (0.1 Drop)}} \\
\quad SIGL prior  & 18.17 & 48.04 & 69.33 & 25.43 & 26.77 & 48.91 & 100.00 & 0.00 & 58.91 & 50.61 & 88.44 & 16.33 \\
\quad PIFM (SIGL) & \textbf{26.93} & \textbf{59.48} & 71.33 & 11.33 & \textbf{42.21} & \textbf{60.76} & 60.75 & 7.48 & \textbf{85.60} & \textbf{83.21} & 16.37 & 18.41 \\
\midrule
\multicolumn{13}{l}{\textit{Mask rate: 50\% (0.5 Drop)}} \\
\quad SIGL prior  & 16.88 & \textbf{49.30} & 72.01 & 27.85 & 22.90 & 52.55 & 100.00 & 0.00 & 50.05 & 45.41 & 87.68 & 18.80 \\
\quad PIFM (SIGL) & \textbf{17.08} & 49.15 & 86.06 & 12.28 & \textbf{28.38} & \textbf{59.58} & 61.20 & 20.38 & \textbf{59.83} & \textbf{58.11} & 38.90 & 36.76 \\
\bottomrule
\end{tabular}}
\end{table}

\subsection{Denoising} 
\label{app:denoising}

Denoising is the complement of expansion, meaning that we seek to remove a set of spurious edges $\ccalE_S$ from $A^{\ccalO}$, such that the edge set of the ground truth is $\ccalE = \ccalE_O \setminus \ccalE_S$.
Hence, the initialization becomes $\bbA_0 = \bbA^{\ccalO} \odot (f_{prior}(\bbA^{\ccalO}) + \bbepsilon_s)$. As in the expansion task, PIFM here is trained with the squared Frobenius distortion (MSE loss).
We corrupt the observed graph by flipping 20\% of the upper-triangle zero-entries into spurious edges (denoted 0.2 Flip); the results are shown in Table~\ref{tab:results_denoising}.
% \coauthor{Please confirm this matches the actual experiment: spurious edges are created by flipping 20\% of the upper-triangle zero-entries to ones, not by removing 20\% of the true edges. The previous wording (``20\% of the edges are flipped'') would describe edge removal instead, which is inconsistent with a denoising (edge-removal) task.}
Similarly to expansion, PIFM (GraphSAGE) attains the best AUC/AP on all datasets, again surpassing the GraphSAGE prior and remaining baselines.
It \emph{strongly} reduces false positives from the given prior initialization, while FNR is low on dense IMDB-B ($2.67$) and higher on sparser sets.
Overall, PIFM removes spurious edges more reliably while also improving other metrics.

\begin{table}[H]
\caption{
    \small{Performance for the \textbf{denoising} task with \textbf{20\% of upper-triangle 0-entries flipped (0.2 Flip)}.
    We report AUC, Average Precision (AP↑), False Positive Rate (FPR↓), and False Negative Rate (FNR↓), all in percent (\%). 
    The best result for each metric is in \textbf{\textcolor{blue}{bold blue}} and the second best is \textcolor{green!50!black}{green}. The metrics are restricted on the upper-triangle 1-region of $A^{\ccalO}$, and compared against $\bbA_1$ on that region.}
}
\label{tab:results_denoising}
\centering
\resizebox{1\textwidth}{!}{
\begin{tabular}{l|cccc|cccc|cccc}
\toprule
\multicolumn{13}{c}{\textbf{Flip Rate: 20\% (0.2 Flip)}} \\
\midrule
& \multicolumn{4}{c|}{\textbf{ENZYMES}} & \multicolumn{4}{c|}{\textbf{PROTEINS}} & \multicolumn{4}{c}{\textbf{IMDB-B}} \\
\cmidrule(lr){2-5} \cmidrule(lr){6-9} \cmidrule(lr){10-13}
\textbf{Method} 
& AP $\uparrow$ & AUC $\uparrow$ & FNR $\downarrow$ & FPR $\downarrow$ 
& AP $\uparrow$ & AUC $\uparrow$ & FNR $\downarrow$ & FPR $\downarrow$ 
& AP $\uparrow$ & AUC $\uparrow$ & FNR $\downarrow$ & FPR $\downarrow$ \\
\midrule
\multicolumn{13}{l}{\textit{Baselines}} \\
\quad GraphSAGE 
& \textcolor{green!50!black}{68.19} & \textcolor{green!50!black}{73.89} & \textbf{\textcolor{blue}{16.14}} & 61.72 
& \textcolor{green!50!black}{73.79} & \textcolor{green!50!black}{76.70} & \textbf{\textcolor{blue}{12.68}} & 60.47 
& 92.54 & 77.29 & 16.77 & 53.37 \\
\quad DiGress + RePaint 
& 41.98 & 49.38 & 87.91 & \textbf{\textcolor{blue}{13.33}} 
& 49.36 & 51.20 & 78.44 & \textbf{\textcolor{blue}{18.19}} 
& 80.54 & 51.59 & 73.41 & 23.49 \\
\quad GDSS + RePaint  
& 44.59 & 50.86 & 69.18 & 29.70 
& 49.72 & 49.43 & 68.39 & 32.64 
& 82.36 & 53.23 & 69.05 & 26.00 \\
\quad Flow w/ Gaussian prior 
& 49.90 & 54.56 & 52.30 & 38.93 
& 57.18 & 58.70 & 62.84 & 24.32 
& \textcolor{green!50!black}{96.75} & \textcolor{green!50!black}{94.63} & \textcolor{green!50!black}{3.80} & \textcolor{green!50!black}{12.66} \\
\midrule
\multicolumn{13}{l}{\textit{Ours}} \\
\quad PIFM (GraphSAGE) 
& \textbf{\textcolor{blue}{69.40}} & \textbf{\textcolor{blue}{77.17}} & \textcolor{green!50!black}{45.66} & \textcolor{green!50!black}{18.14} 
& \textbf{\textcolor{blue}{77.43}} & \textbf{\textcolor{blue}{81.78}} & \textcolor{green!50!black}{32.41} & \textcolor{green!50!black}{20.91} 
& \textbf{\textcolor{blue}{98.46}} & \textbf{\textcolor{blue}{96.52}} & \textbf{\textcolor{blue}{2.67}} & \textbf{\textcolor{blue}{12.10}} \\
\bottomrule
\end{tabular}
}
\end{table}

\subsection{Loss-calibration analysis on sparse graphs}
\label{app:bce-ablation}

The MSE and CE instantiations of PIFM (Table~\ref{tab:results_50_percent_mask_final}) 
show a pronounced calibration difference on the sparser datasets, PROTEINS and 
ENZYMES. Here we analyze its cause.

Under the MSE distortion, PIFM is trained as a continuous 
regression toward the conditional mean. On a sparse graph, non-edges vastly 
outnumber edges, so the loss is dominated by the negative class and the 
fitted edge scores concentrate near zero (the conditional-mean target of the rare positive class is itself small). True 
edges are then treated as rare events and underweighted, and at the fixed 
decision threshold of $0.5$ almost all of them are missed, inflating the 
false-negative rate (FNR). This is a thresholded-calibration effect rather 
than a ranking failure: the threshold-independent metrics (AP, AUC) remain 
informative even where FNR is high.

The cross-entropy (CE) instantiation addresses this directly. Inspired by 
CatFlow~\citep{eijkelboom2024variational}, which establishes cross-entropy as 
an effective distortion for flow matching on graphs, the CE distortion is an 
element-wise binary cross-entropy on the upper triangle of the masked region. 
In contrast to CatFlow's unweighted categorical objective, we additionally 
apply a per-batch positive-class weight $w = \min(n_{\text{neg}}/n_{\text{pos}}, 50)$ 
to counter the class imbalance specific to our conditional, sparse-graph setting. We 
emphasize that the positive weighting, not the switch from MSE to 
cross-entropy alone, is what restores calibration: an \emph{unweighted} 
cross-entropy on these datasets would face the same negative-dominated 
gradient and leave the fitted probabilities below the $0.5$ threshold, so the thresholded predictions again collapse to no edges, as with MSE. The weighting 
rebalances the per-edge contributions so that the rare positive class is no 
longer ignored, sharply reducing FNR (Table~\ref{tab:results_50_percent_mask_final})
at a substantial cost in FPR (which can rise from near zero to $45$--$89\%$ on the sparsest datasets), while AP/AUC stay comparable. The clamp at $50$ caps 
the weight on the very sparsest graphs, where an unbounded ratio would 
destabilize training.

These findings suggest that there are two natural directions for handling sparse graphs more effectively. One option is to adopt a BCE-style objective, which improves threshold calibration for binary edges but departs from the classical closed-form distortion-perception formulation. Another option is to retain the continuous distortion-perception objective while replacing the Gaussian perturbation with a noise model better suited to sparse binary data. We leave this second direction for future work.

\subsection{Ablation}
\label{app:ablation}

Our method has two main hyperparameters: 

\begin{enumerate}
    \item $\sigma_s$, which is used for computing $\bbepsilon_{s}\sim \ccalN(0, \sigma_s^2)$ in $\bbA_0 = \bbA^{\ccalO} + (1-\xi)\odot \left(f_{\text{prior}}(\bbA^{\ccalO})
        + \bbepsilon_{s}\right)$
    \item $K$, which is the total number of steps in the Euler approximation
\end{enumerate}

We ablate these two hyperparameters.

\paragraph{Performance as a function of $\sigma_s$.}
We run an ablation of the performance of PIFM with GraphSAGE as a function of $\sigma_s$.
We focus on ENZYMES and IMDB, and we evaluate the ROC for the best value of $K$ for each noise level.

\begin{figure}[H]
    \centering
    \includegraphics[width=0.5\linewidth]{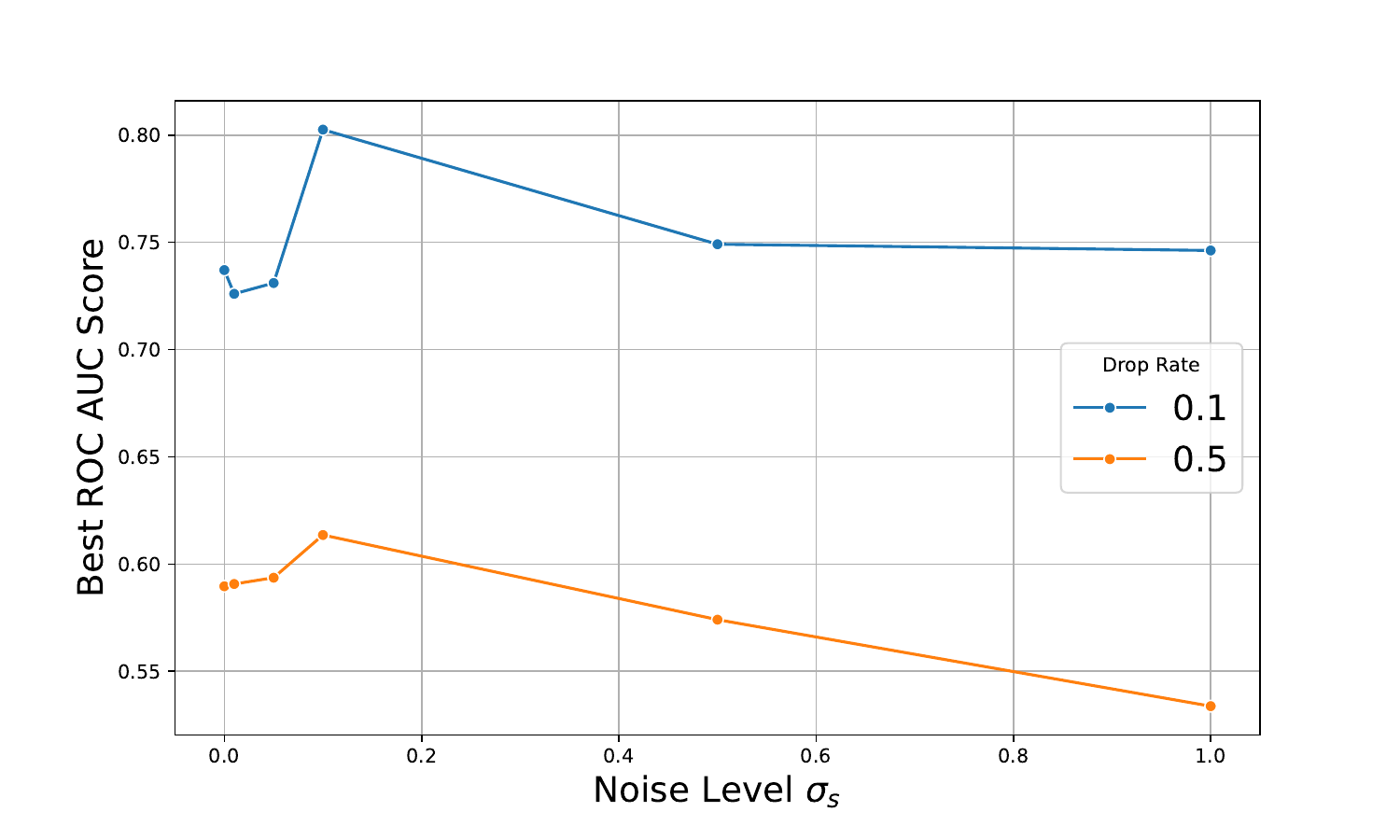}
    \caption{ROC as a function of the noise $\sigma_s$ in $p(\bbA_0)$.
    The impact of noise level $\sigma_s$ on model performance, measured by the best ROC AUC score. Results are shown for two different drop rates: 0.1 (blue) and 0.5 (orange). A small amount of noise improves performance for both configurations, after which increasing noise leads to performance degradation. }
    \label{fig:roc-sigma}
\end{figure}

The ablation is illustrated in Fig.~\ref{fig:roc-sigma}.
First, we observe that the gains of using PIFM are higher for a smaller drop rate, as expected; in particular, we observe that PIFM with $\sigma_s$ jumps from $\approx 0.73$ for $\sigma_s = 0$ to $\approx 0.81$ for $\sigma_s=0.1$.
Second, for both configurations, performance peaks not at zero noise, but at a small noise level of $\sigma_s=0.1$.
This suggests that a slight injection of noise benefits model generalization. 
Beyond this optimal point, increasing the noise level leads to a steady decline in performance, meaning that the effect of the prior decreases, as expected. 

\begin{figure}[H]
    \centering
    % --- First Subfigure ---
    \begin{subfigure}[b]{0.48\linewidth}
        \centering
    \includegraphics[width=1.1\linewidth]{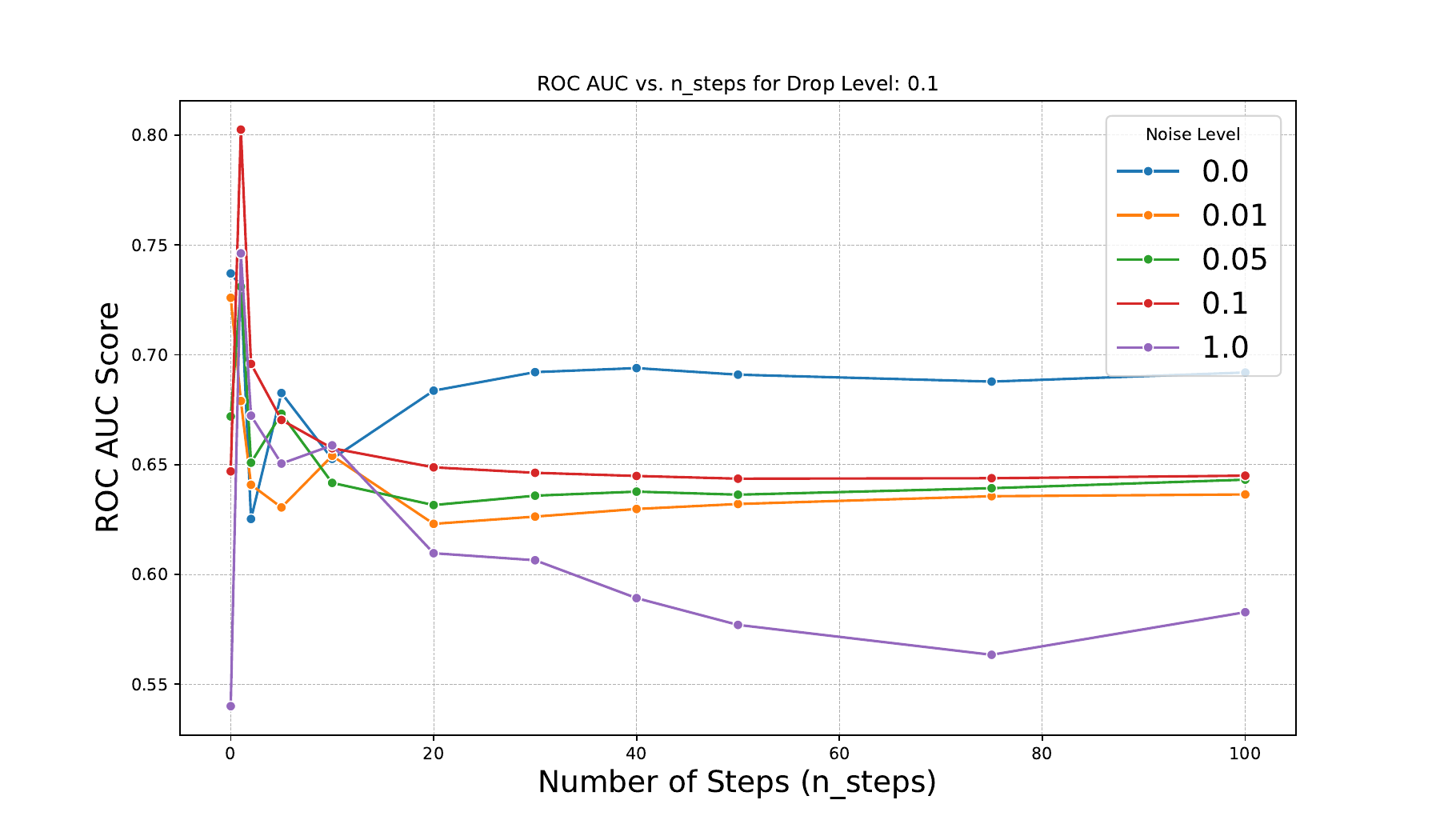}
    \end{subfigure}    
    % --- Second Subfigure ---
    \begin{subfigure}[b]{0.48\linewidth}
        \centering
        % Replace with your second figure's file path
        \includegraphics[width=1.1\linewidth]{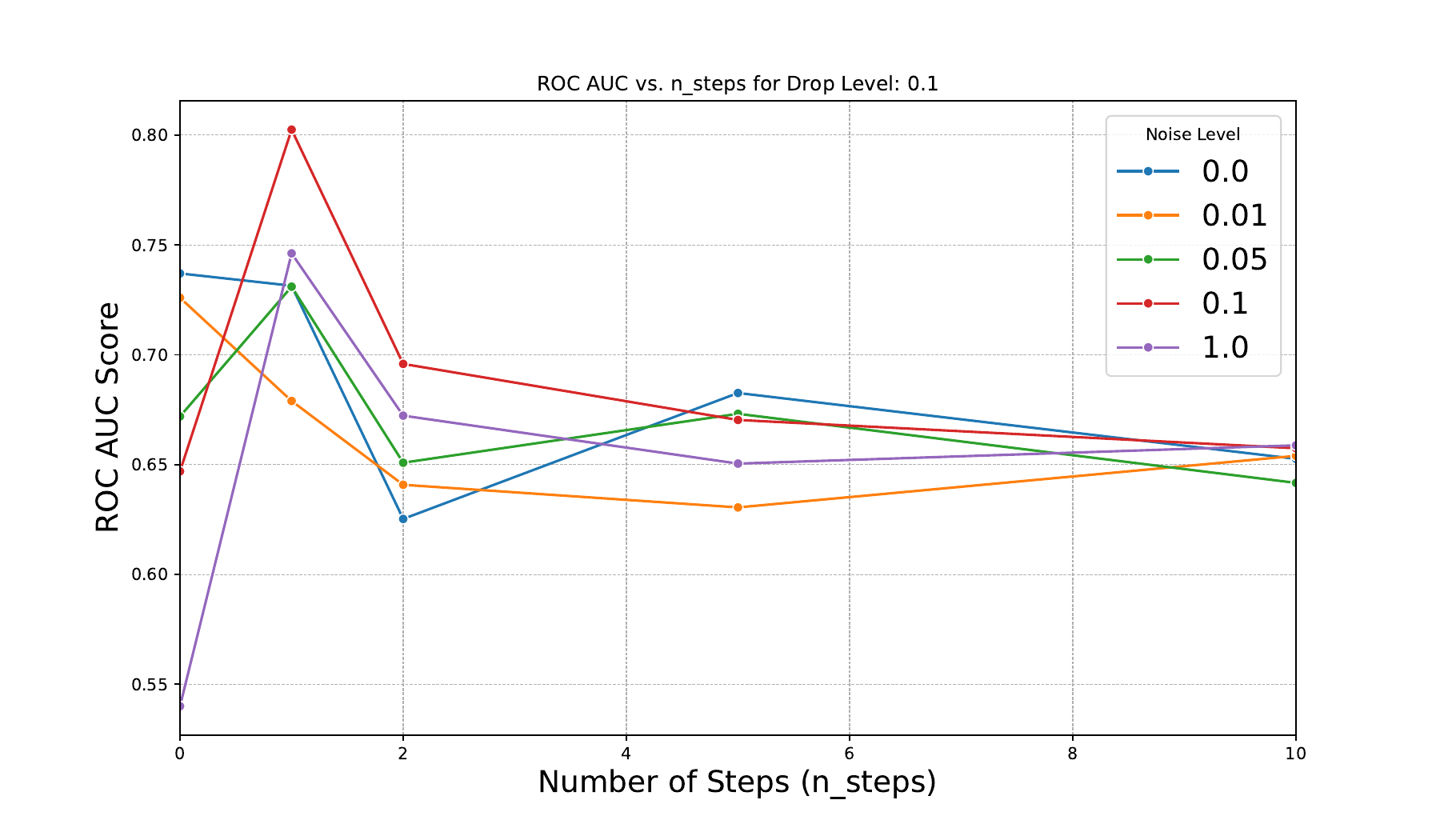} 
    \end{subfigure}
    % --- Overall Figure Caption ---
    \caption{An analysis of the ROC AUC score as a function of the number of processing steps (K) for a drop rate of 10\%. This experiment was conducted with a fixed drop rate of 0.1, while varying the noise level, $\sigma_s$. The results show that the optimal number of steps is small, typically under 10.}
    \label{fig:roc-k}
\end{figure}

\begin{figure}[H]
    \centering
    % --- First Subfigure ---
    \begin{subfigure}[b]{0.48\linewidth}
        \centering
    \includegraphics[width=1.1\linewidth]{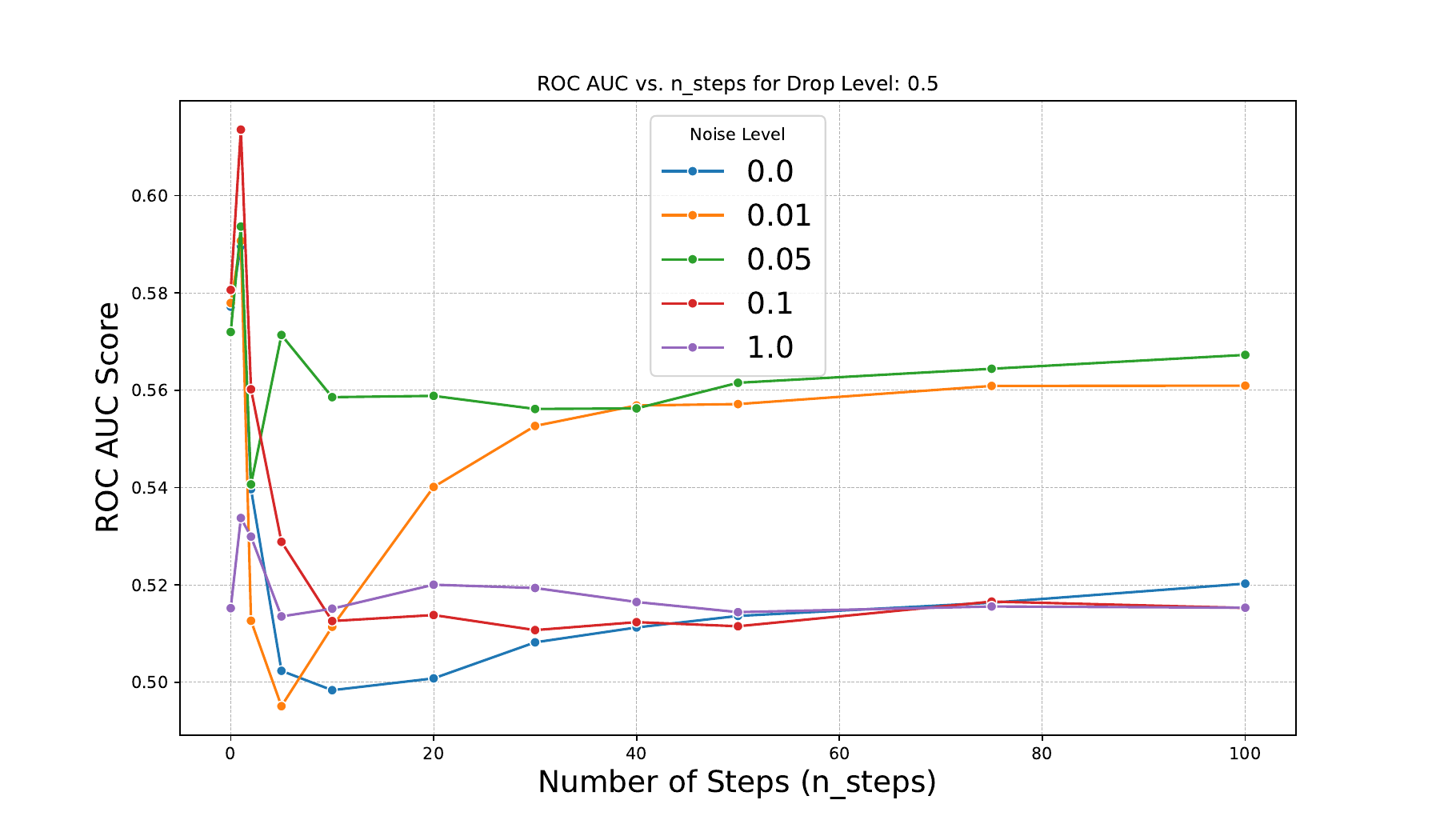}
    \end{subfigure}    
    % --- Second Subfigure ---
    \begin{subfigure}[b]{0.48\linewidth}
        \centering
        % Replace with your second figure's file path
        \includegraphics[width=1.1\linewidth]{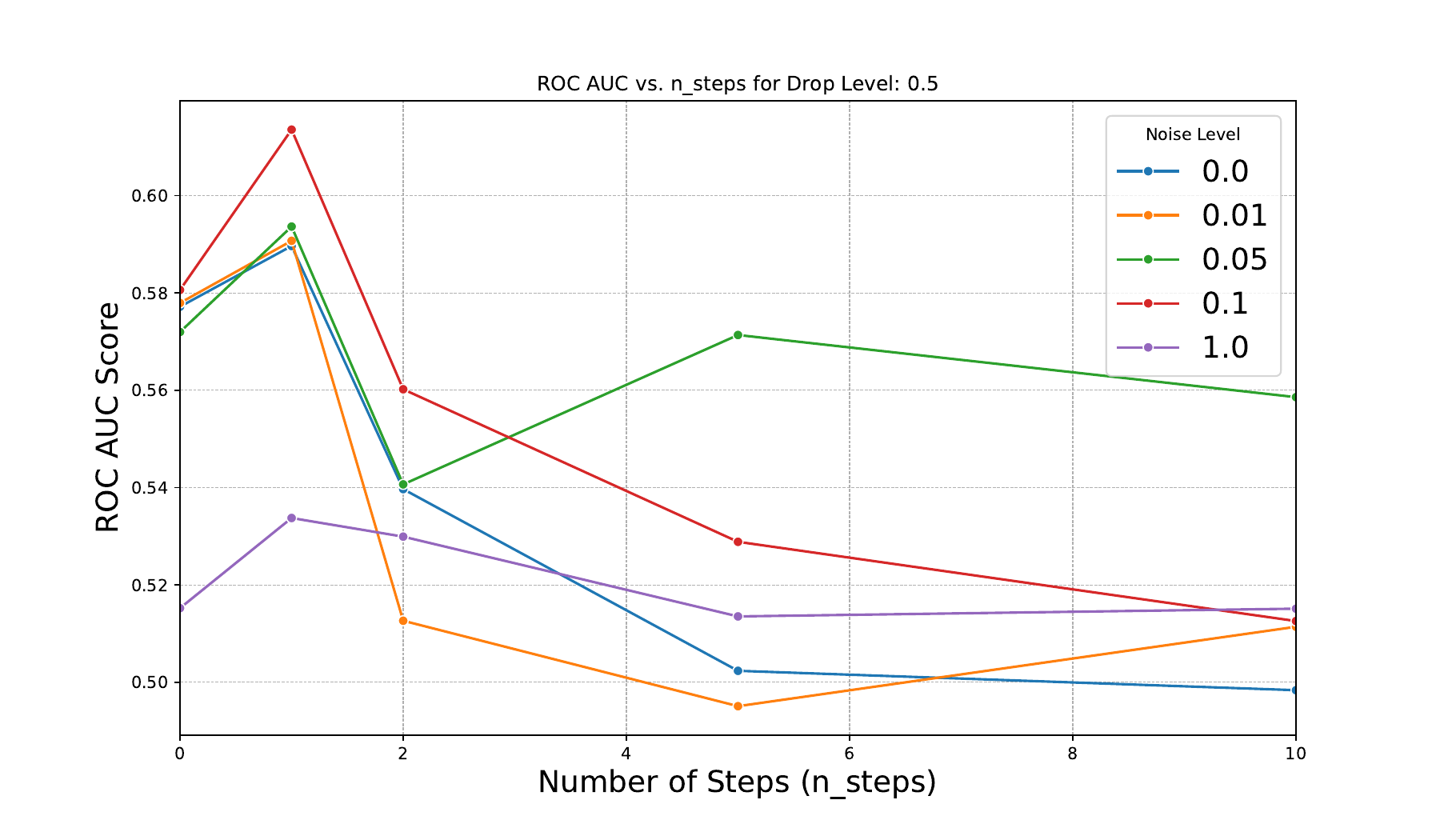} 
    \end{subfigure}
    % --- Overall Figure Caption ---
    \caption{An analysis of the ROC AUC score as a function of the number of processing steps (K) for a drop rate of 50\%. This experiment was conducted with a fixed drop rate of 0.5, while varying the noise level, $\sigma_s$. The results show that the optimal number of steps is small, typically under 10.}
    \label{fig:roc-k-05}
\end{figure}

% \vspace{-3cm}
\paragraph{Performance as a function of $K$.}
To determine the optimal number of processing steps, $K$, we evaluated model performance while varying this parameter from 1 to 100. 
Figures~\ref{fig:roc-k} and~\ref{fig:roc-k-05} shows the results for a fixed drop rate of 0.1 and 0.5 respectively, across five different noise levels.

A key observation is that peak performance, in terms of AUC-ROC, is achieved within a very small number of steps, typically for $K<10$. In particular, the introduction of a moderate noise level allows the model to achieve its highest overall score ($\approx 0.80$ ROC AUC) in a single step ($K=1$).
However, this advantage diminishes as the number of steps increases. 
The model without noise ($\sigma_s=0.0$) provides the most stable and consistently high performance for larger $K$. 
Conversely, a high noise level ($\sigma_s = 1$) consistently degrades performance regardless of the number of steps. This analysis suggests a trade-off: while noise can provide a significant boost for models with very few steps, a no-noise configuration is more robust for models with a larger number of steps.

% \begin{figure}[H]
%     \centering
%     \includegraphics[width=0.8\linewidth]{Figures/roc_vs_nsteps_drop_0.1_from_1.pdf}
%     \caption{An analysis of the ROC AUC score as a function of the number of processing steps (K) for a drop rate of 10\%. This experiment was conducted with a fixed drop rate of 0.1, while varying the noise level, $\sigma_s$. The results show that the optimal number of steps is small, typically under 10.}
%     \label{fig:roc-k}
% \end{figure}

\begin{figure}[H]
    \centering
    \includegraphics[width=\textwidth]{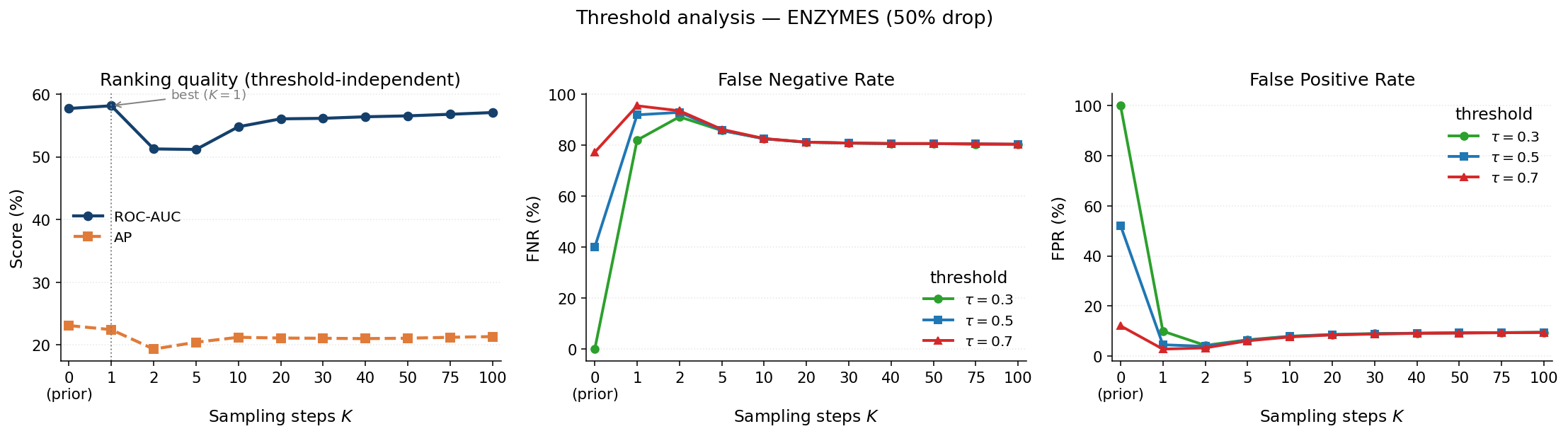}
    \caption{Threshold analysis on \textsc{ENZYMES} (link prediction, 50\% drop) versus the
    number of sampling steps $K$. \textbf{Left:} ROC-AUC and AP are computed on the raw scores
    and are therefore \emph{identical} for all binarization thresholds $\tau\in\{0.3,0.5,0.7\}$
    (threshold-independent), peaking at $K{=}1$. \textbf{Center/Right:} the false-negative
    (FNR) and false-positive (FPR) rates at $\tau=0.3,0.5,0.7$; the threshold shifts only these
    binarized error rates, and only appreciably at small $K$, where the outputs lie near the
    decision boundary. ``$K{=}0$'' is the prior initialization.}
    \label{fig:enzymes_threshold}
\end{figure}

\begin{figure}[H]
    \centering
    \includegraphics[width=\textwidth]{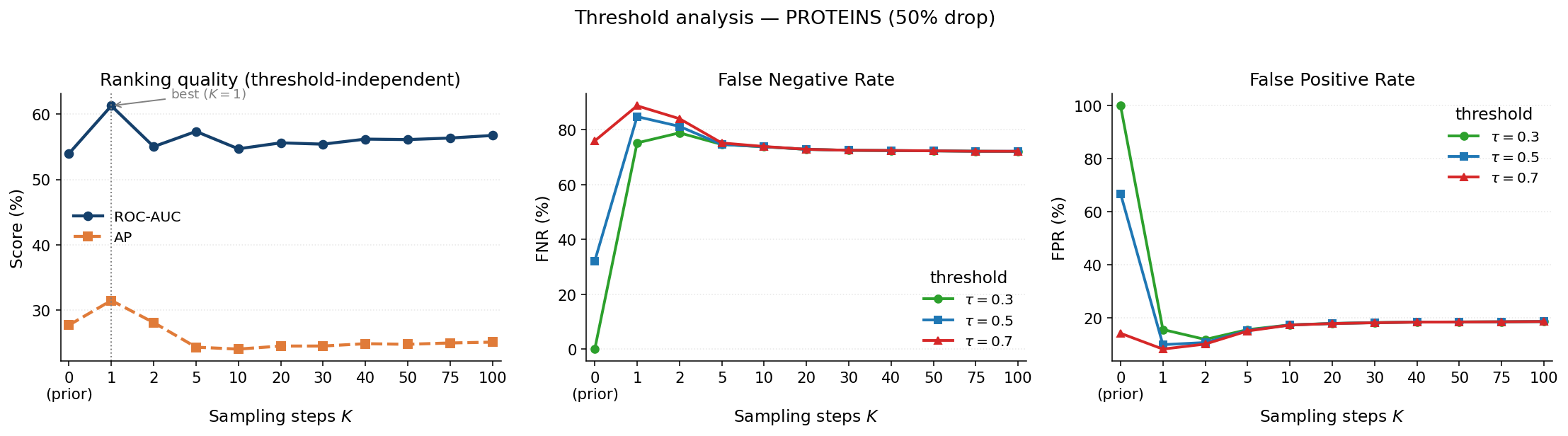}
    \caption{Threshold analysis on \textsc{PROTEINS} (link prediction, 50\% drop) versus the
    number of sampling steps $K$, with panels as in Fig.~\ref{fig:enzymes_threshold}.
    ROC-AUC and AP (left) coincide across thresholds, while lowering the threshold trades a
    lower FNR (center) for a higher FPR (right).}
    \label{fig:proteins_threshold}
\end{figure}

\paragraph{Threshold levels.}
{
We investigated the impact of alternative thresholding levels (0.3 and 0.7, beyond the
standard 0.5), with results for \textsc{ENZYMES} and \textsc{PROTEINS} in
Figs.~\ref{fig:enzymes_threshold} and~\ref{fig:proteins_threshold}, each reporting the ranking
metrics and the binarized error rates as a function of the number of steps $K$.
Since AP and AUC are computed from the raw scores, they are threshold-independent: the
ROC-AUC-vs-$K$ curves are the same across the three thresholds (left panels),
so thresholding does not change ranking quality. The threshold instead governs the binarized
error rates (center and right panels): lowering it reduces the FNR at the cost of a higher
FPR (e.g., on \textsc{ENZYMES} at $K{=}1$ the FNR drops from $91.9$ to $82.0$ as the FPR rises
from $4.5$ to $9.9$ when lowering from $\tau=0.7$ to $\tau=0.3$). This effect is appreciable only for small $K$ (and for the prior at
$K{=}0$), where the model's outputs lie nearer the decision boundary; for larger $K$ the outputs
are pushed towards the binary extremes (0 or 1) and the three thresholds nearly coincide.
Crucially, across steps the FNR and FPR merely trade off, so an alternative threshold yields no
significant improvement in overall performance; the principled rebalancing of these error rates
is instead provided by the cross-entropy distortion (Sec.~\ref{subsec:flowmodel}). }

\subsection{Distortion-perception trade-off.} 
\label{app:mmd}

\begin{figure}[H]
    \centering
    % --- First Subfigure ---
    \begin{subfigure}[b]{0.48\linewidth}
        \centering
    \includegraphics[width=1.1\linewidth]{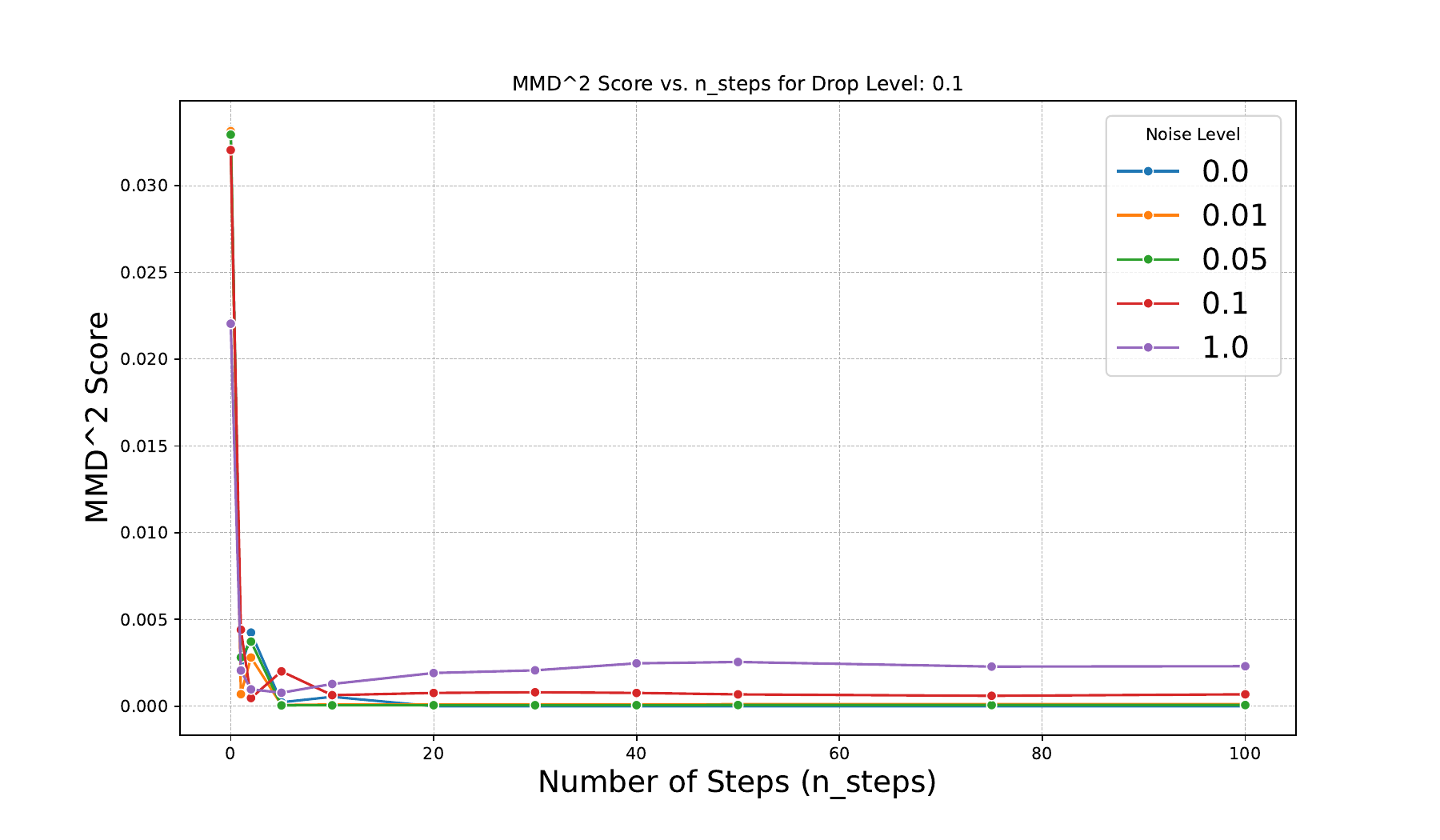}
    \end{subfigure}    
    % --- Second Subfigure ---
    \begin{subfigure}[b]{0.48\linewidth}
        \centering
        % Replace with your second figure's file path
    \includegraphics[width=1.1\linewidth]{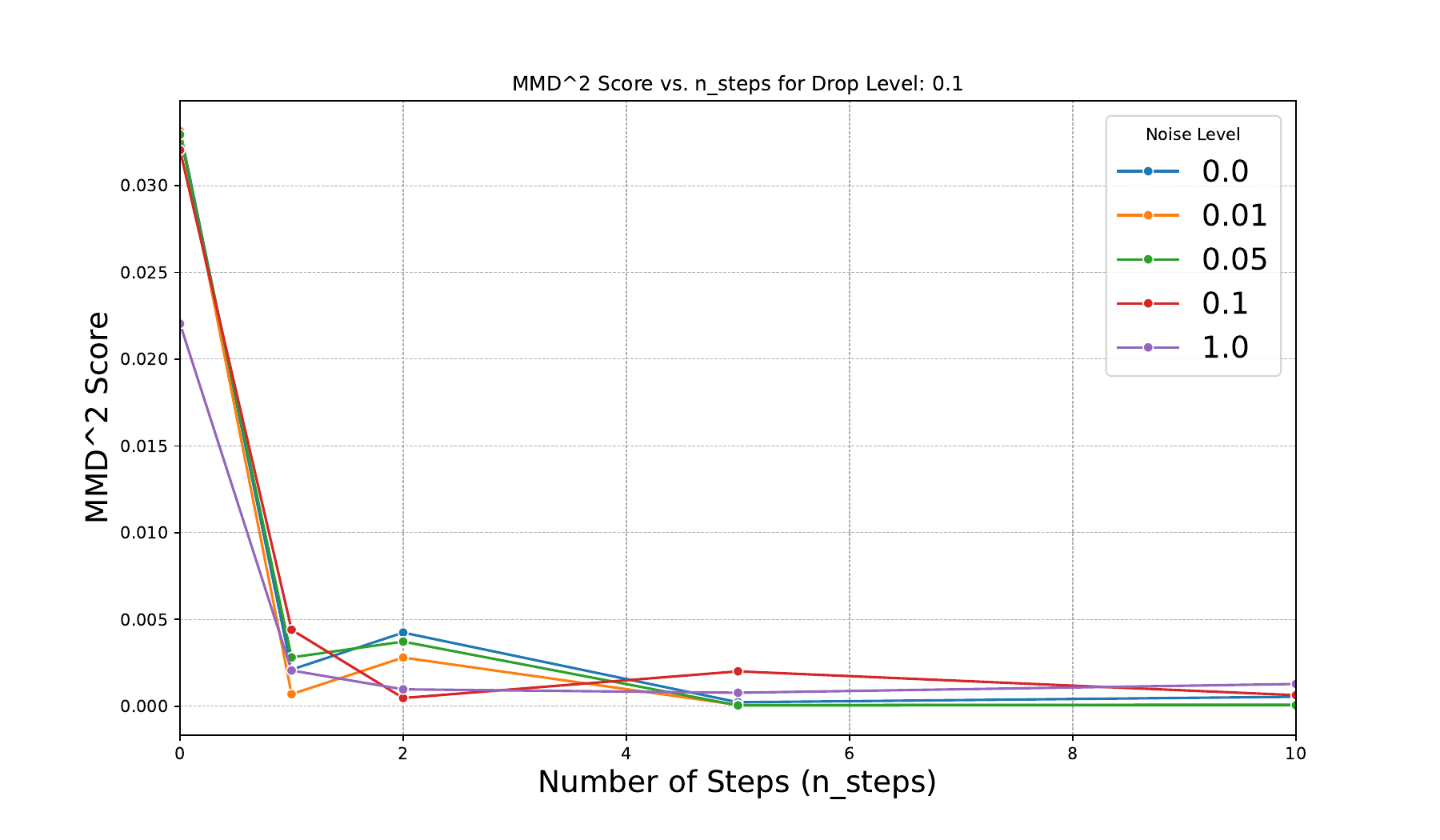}
    \end{subfigure}
    % --- Overall Figure Caption ---
    \caption{ Analysis of the perception component of the distortion-perception trade-off. The plot shows the MMD² score (where lower is better) versus the number of steps, K, for a fixed drop rate of 0.1. Each line represents a different noise level $\sigma_s$. }
    \label{fig:mmd-01}
\end{figure}

\begin{figure}[H]
    \centering
    % --- First Subfigure ---
    \begin{subfigure}[b]{0.48\linewidth}
        \centering
    \includegraphics[width=1.1\linewidth]{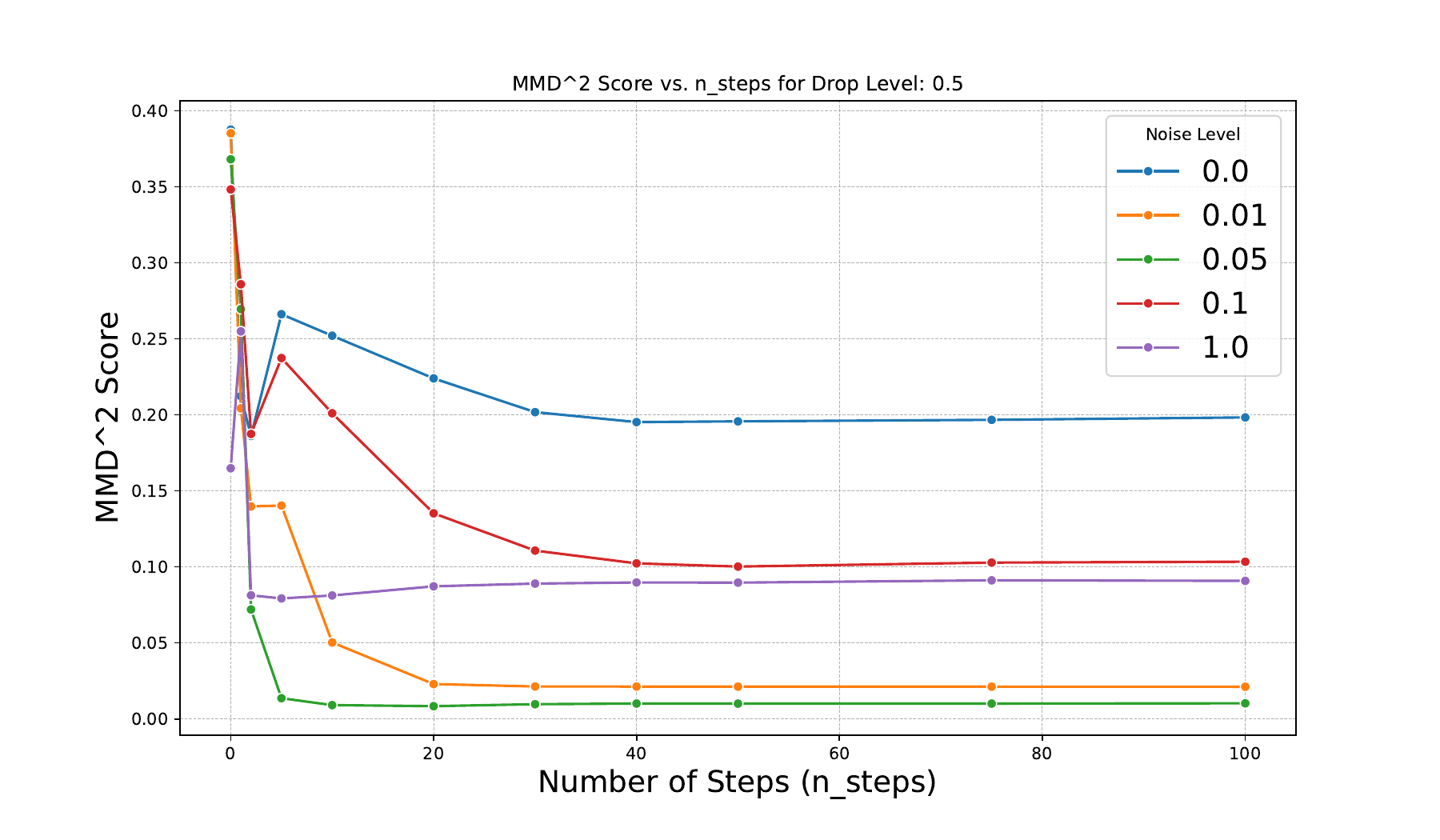}
    \end{subfigure}    
    % --- Second Subfigure ---
    \begin{subfigure}[b]{0.48\linewidth}
        \centering
        % Replace with your second figure's file path
    \includegraphics[width=1.1\linewidth]{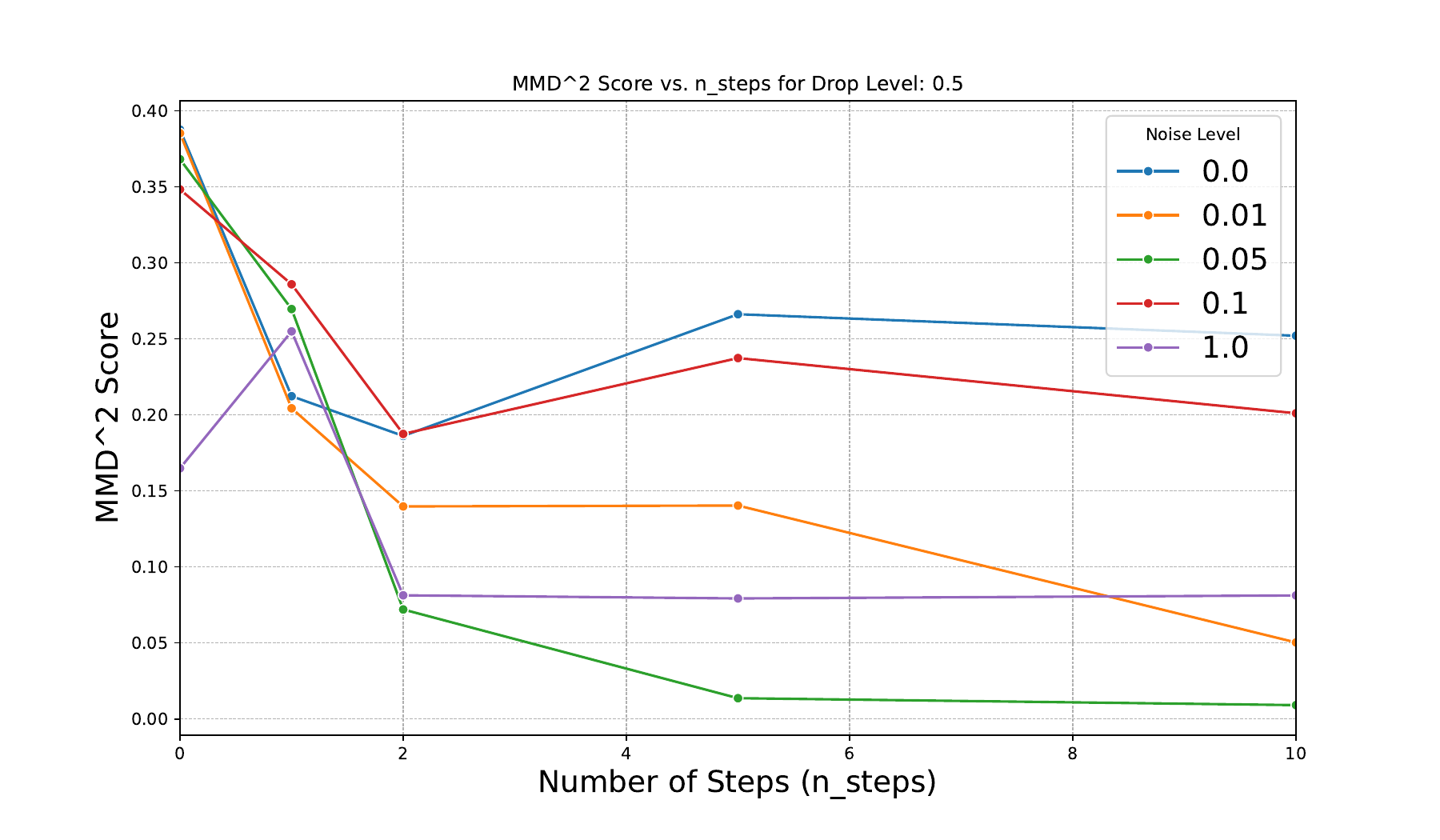}
    \end{subfigure}
    % --- Overall Figure Caption ---
    \caption{ Analysis of the perception component of the distortion-perception trade-off. The plot shows the MMD² score (where lower is better) versus the number of steps, K, for a fixed drop rate of 0.5. Each line represents a different noise level $\sigma_s$.}
    \label{fig:mmd-05}
\end{figure}

Here we expand on the distortion-perception trade-off by computing the MMD as a function of the number of steps
The results are shown in Figures~\ref{fig:mmd-01} and~\ref{fig:mmd-05}. 
Again, both figures show the MMD² distance dropping sharply over the first few steps (most noticeably for $0 < \sigma_s \leq 0.1$), rather than decreasing monotonically in $K$: for some noise levels it plateaus or rebounds after its early minimum. 
In other words, if we aim for a high-quality perceptual reconstruction, we should consider $\sigma_s = 0.01$ or $0.05$.
However, if we are aiming for high reconstruction quality in terms of AUC-ROC, we should use $\sigma_s = 0.1$ (see Fig.\ref{fig:roc-sigma}).
In other words, the choice of the $K$ is heavily dependent on the downstream task.

\subsection{Evaluation using additional metrics}

\begin{figure}[H]
\centering
\setlength{\tabcolsep}{6pt} % horizontal spacing between columns
\renewcommand{\arraystretch}{1.0}

\begin{tabular}{cc}
\begin{minipage}{0.48\linewidth}
\centering
\includegraphics[width=\linewidth]{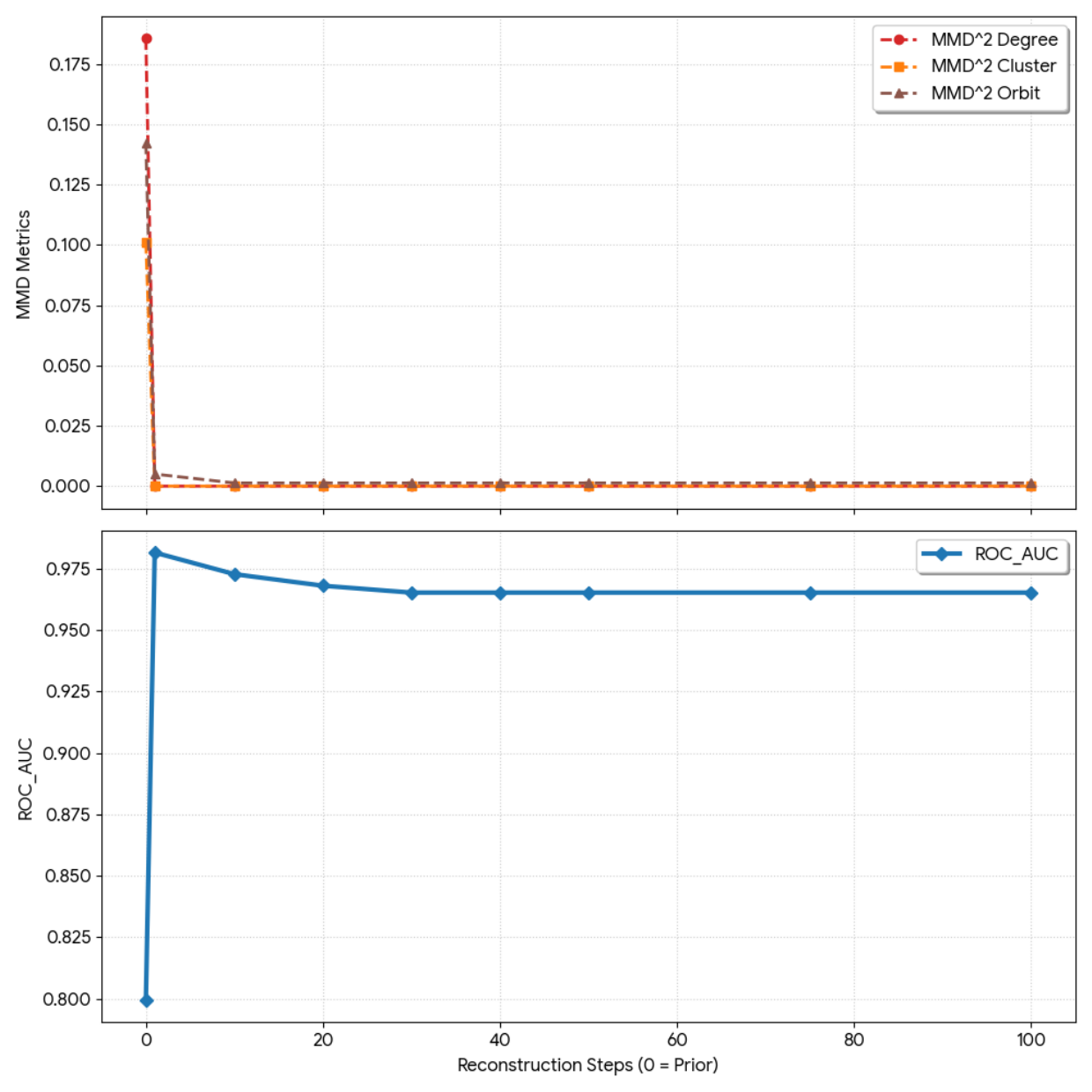}
\caption{IMDB dataset, expansion task, 10\% drop rate}
\label{fig:imdb_01}
\end{minipage}
&
\begin{minipage}{0.48\linewidth}
\centering
\includegraphics[width=\linewidth]{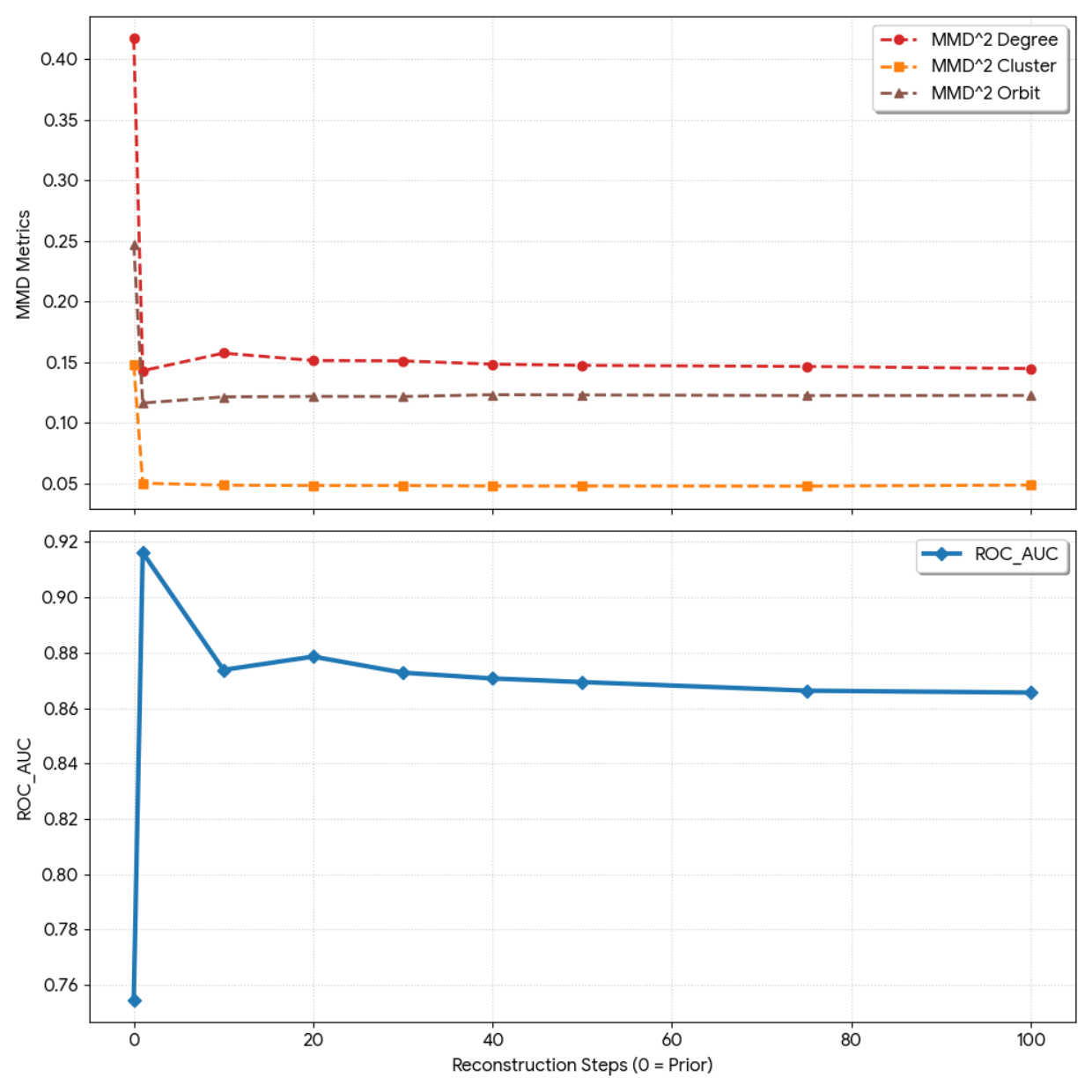}
\caption{IMDB dataset, expansion task, 50\% drop rate}
\label{fig:imdb_05}
\end{minipage}
\\[4pt]
\begin{minipage}{0.48\linewidth}
\centering
\includegraphics[width=\linewidth]{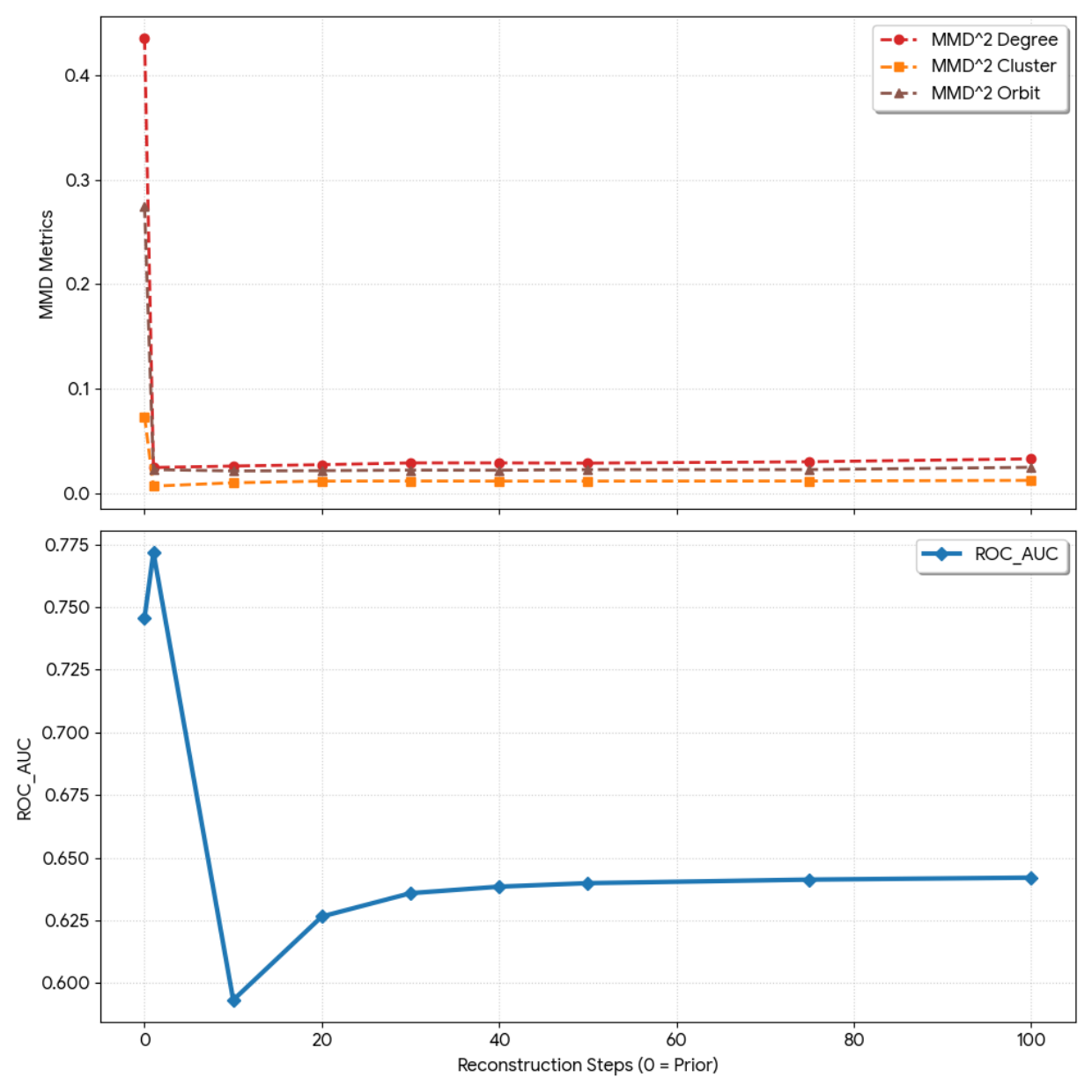}
\caption{PROTEINS dataset, expansion, 10\% drop rate}
\label{fig:proteins_01}
\end{minipage}
&
\begin{minipage}{0.48\linewidth}
\centering
\includegraphics[width=\linewidth]{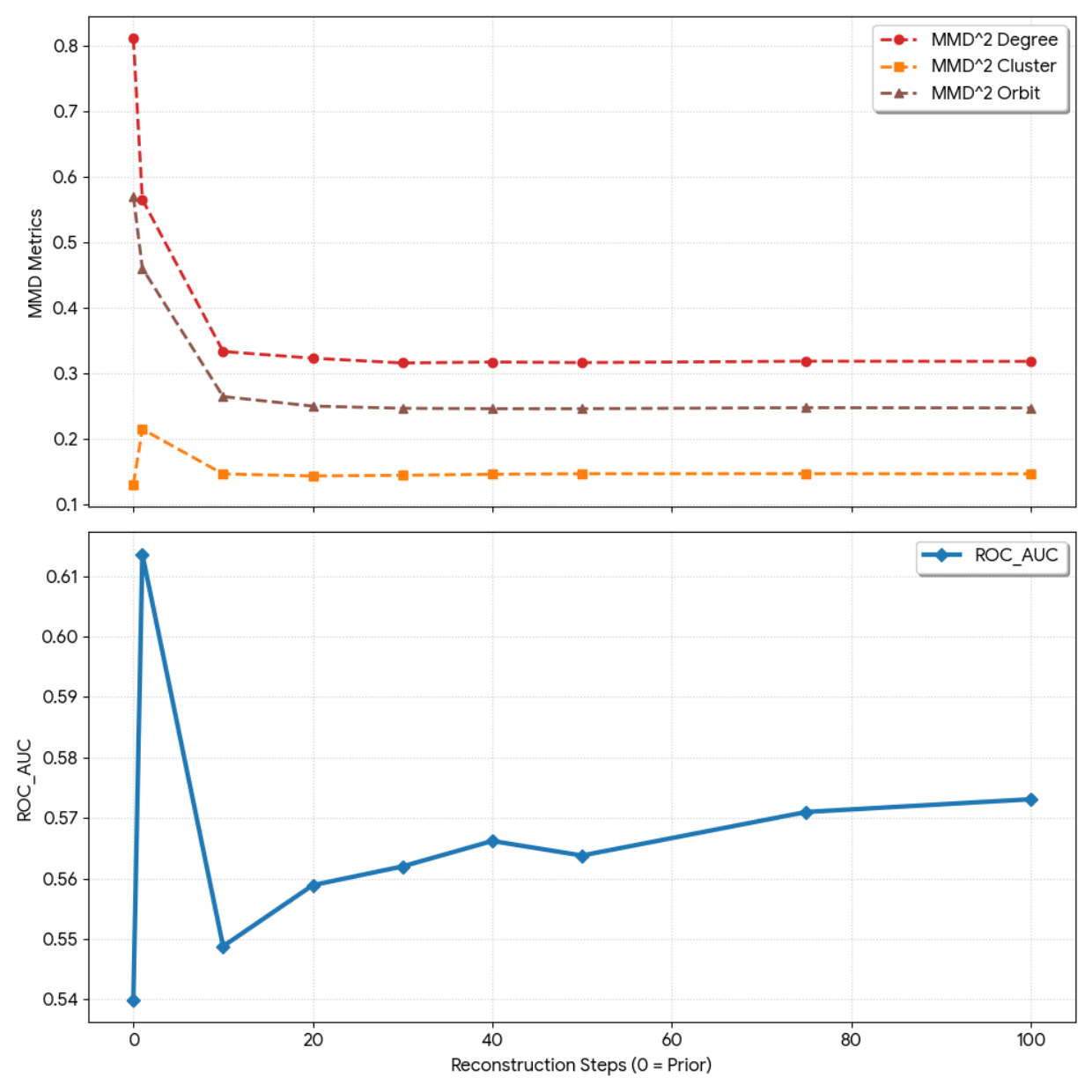}
\caption{PROTEINS dataset, expansion, 50\% drop rate}
\label{fig:proteins_05}
\end{minipage}
\end{tabular}
\end{figure}

\begin{figure}[H]
\centering
\begin{subfigure}{0.48\linewidth}
    \centering
    \includegraphics[width=\linewidth]{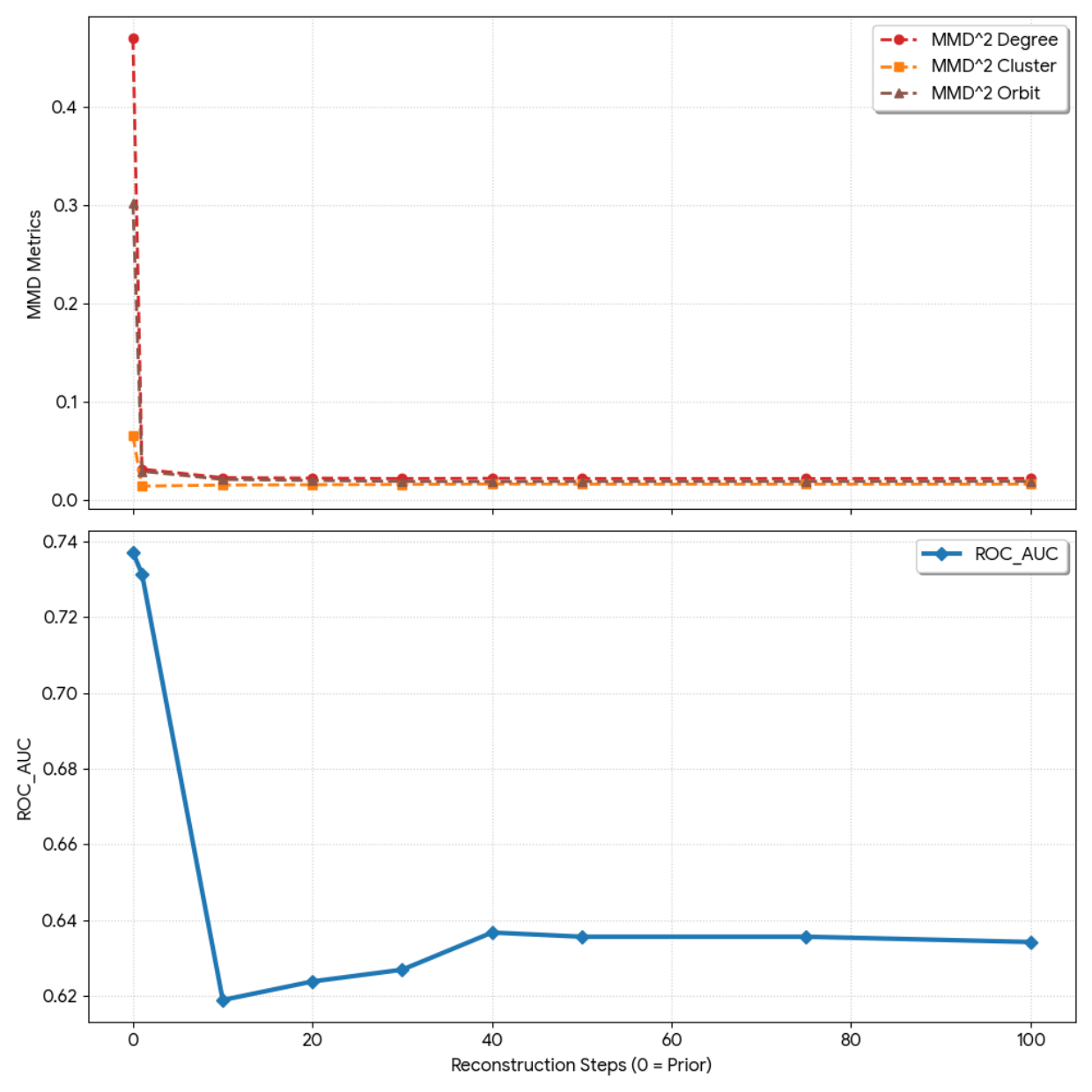}
    \caption{10\% drop rate}
    \label{fig:enzymes_01}
\end{subfigure}
\hfill
\begin{subfigure}{0.48\linewidth}
    \centering
    \includegraphics[width=\linewidth]{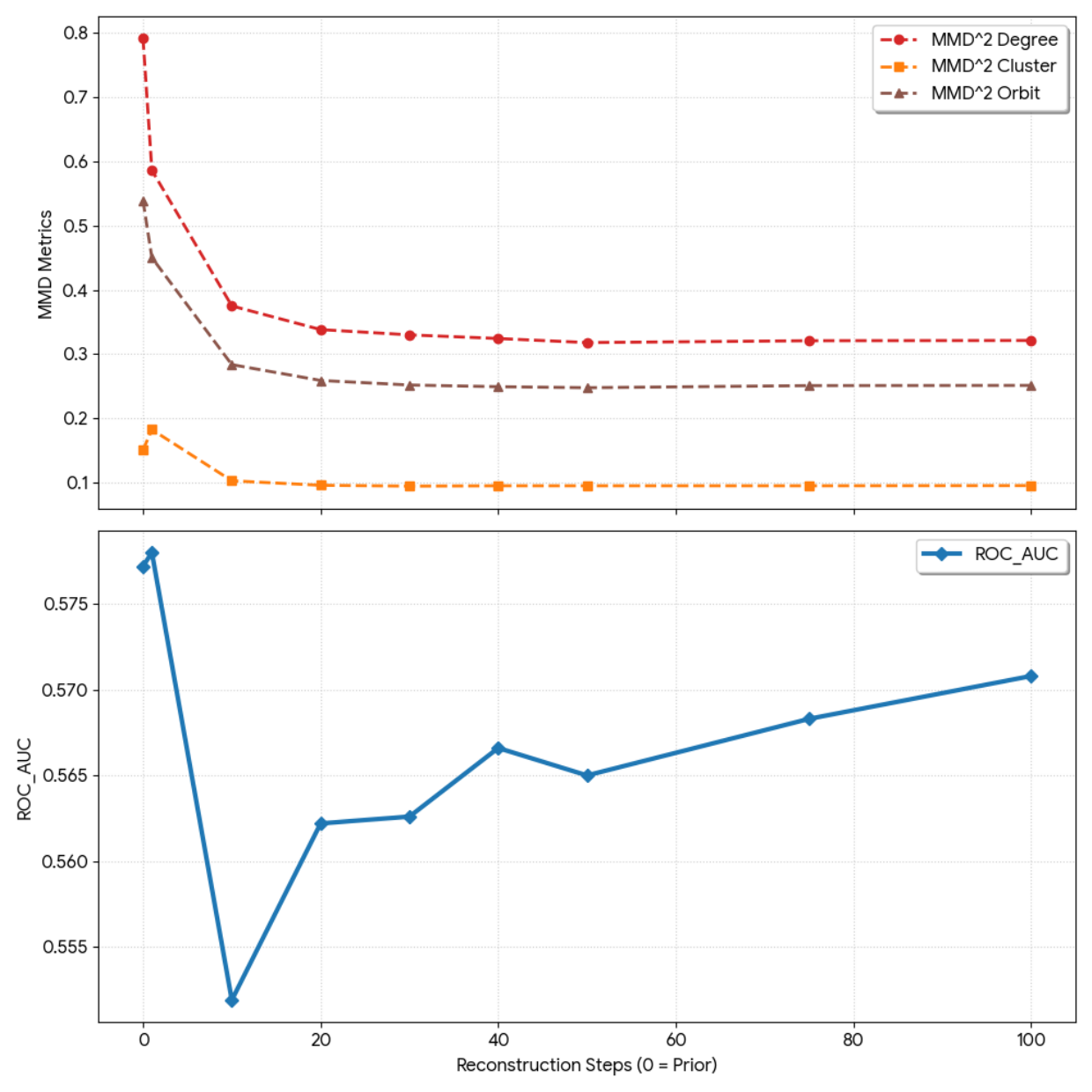}
    \caption{50\% drop rate}
    \label{fig:enzymes_05}
\end{subfigure}
\caption{ENZYMES dataset, expansion task.}
\label{fig:enzymes_pair}
\end{figure}

This section incorporates additional metrics to showcase the observed performance trade-off.
The results for IMDB are in Figs.~\ref{fig:imdb_01} and~\ref{fig:imdb_05}, for PROTEINS in Figs.~\ref{fig:proteins_01} and~\ref{fig:proteins_05}, and for ENZYMES in Figs.~\ref{fig:enzymes_01} and~\ref{fig:enzymes_05}.
While an increased number of steps yields an improvement in generating an estimated graph with statistics that more closely align with the ground-truth distribution, the reconstruction performance (measured in terms of AUC) declines relative to the initial step. Critically, the trend is found to be highly contingent on the underlying dataset's sparsity. For the dense case (IMDB), the AUC exhibits a consistent monotonic decrease after the optimal initial guess, independent of drop rates. In contrast, the sparser PROTEINS and ENZYMES datasets demonstrate an intermediate improvement as the number of steps increases, though their overall AUC still trails that achieved at the initial step ($K=1$).

% \begin{figure}[H]
%     \centering
%     \begin{minipage}{0.48\linewidth}
%         \centering
%         \includegraphics[width=\linewidth]{imdb1.png}
%         \caption{IMDB dataset, expansion task, 50\% drop rate}
%         \label{fig:imdb}
%     \end{minipage}
%     \hfill
%     \begin{minipage}{0.48\linewidth}
%         \centering
%         \includegraphics[width=\linewidth]{proteins1.png}
%         \caption{PROTEINS dataset, expansion, 50\% drop rate}
%         \label{fig:proteins}
%     \end{minipage}
% \end{figure}

\subsection{PIFM on large-scale graphs}
\label{app:cora_exp}

In this section, we train PIFM on large-scale graphs. 
In particular, we focus on CORA~\cite{yang2016revisiting}.

\paragraph{Building the dataset.} 
To enable scalable diffusion training on Cora while maintaining full-graph link prediction capabilities, we introduce a subgraph-based variant of PIFM following~\cite{Limnios2023SaGess}. We instantiate this via an edge-centered sampling scheme, where each subgraph represents a k-hop ego-network (capped at a maximum node count) centered around a seed edge from the training split. To ensure reproducibility, we sample a fixed set of seed edges (both positive and negative) and corresponding subgraphs which remain constant throughout each run.

Following the 50/15/35 (train/validation/test) edge split, we evaluate link prediction on held-out edges using the protocol established in the main paper. Furthermore, we utilize purely structural node features. By default, we concatenate: (i) Laplacian positional encodings derived from the smallest $k$ generalized eigenvectors of $Lv=\lambda Dv$ computed on the training adjacency; and (ii) a 2-dimensional local context vector comprising both raw and normalized node degrees.

\paragraph{Training.} 
To initialize PIFM, we first pre-train the NCN structural prior on the full Cora graph using only the training edge split (50\% of edges). These learned embeddings are subsequently used to initialize PIFM for inference on the test split (35\% of edges).

During PIFM training, each training edge seeds a unique edge-centered subgraph. Within each subgraph, we define the observed context as all other training edges, and a hidden region comprising all remaining node pairs (including non-edges and edges outside the context). The seed edge is explicitly masked from the context and marked as the only supervision target. Consequently, each training example tasks the model with reconstructing a single missing edge within its local neighborhood.

We construct the initial state matrix $\bbA_0$ by combining the observed context with structural prior predictions in the hidden region. The flow model is trained by flow-matching under the squared Frobenius distortion
(MSE loss), with the loss computed only on the supervised entries. While the flow ODE updates all entries within the hidden region during inference, the gradient is computed exclusively from the seed edge. This adapts the global training procedure of Algorithm 1 to a localized, subgraph-based regime. Noise is still added to the seeded edge during training, but no noise is used during inference

\paragraph{Inference.}
For inference, every held-out test edge seeds a k-hop subgraph. We define the observed context using the training edges, remove the centered test edge from the mask, and apply diffusion to the resulting hidden region. We evaluate stitched scores for held-out edges. To resolve potential overlaps, we aggregate predictions via logit averaging across all subgraphs where a specific edge is present, producing a single probability matrix over all node pairs. We then compute metrics on the held-out positive and negative edges. Since PIFM and the structural prior baseline are evaluated on the exact same set of pairs, the results isolate the specific benefits of the diffusion process.

\paragraph{Results.}

Results in Table \ref{table:cora_experiment_appendix}  show that PIFM is learning useful global structural representations even on very large graphs by exploiting the subgraph sampling approach.

\begin{table}[h!]
\centering
\caption{\small{Link prediction on \textsc{Cora} (test edges split) with an NCN structural prior. 
We report AUC, Average Precision (AP), False Positive Rate (FPR), False Negative Rate (FNR). 
We report PIFM's performance on Cora for different $k$-hop neighborhoods during subgraph sampling, and some notable hyperparameters we tuned. We include also the average node counts for each subgraph for each case. Best values for each metrics is in \textbf{bold}.} All the models are all evaluated on the same 50/15/35 edge split, and the results are using the best saved checkpoints based on AUC for each combination of the hyperparameters.}
\label{table:cora_experiment_appendix}
\resizebox{\textwidth}{!}{%{\color{blue}
\begin{tabular}{l c c c c}
\toprule
Method  & AUC$\uparrow$ & AP$\uparrow$ & FPR$\downarrow$ & FNR$\downarrow$ \\
\midrule
NCN~\cite{wangneural}                 & 93.78 & \textbf{93.89} & 23.50 & \textbf{6.71}  \\
\midrule
\text{1hop ($\sim$ 8.9 nodes)     }\\ 
{\fontsize{8}{7}\selectfont\ttfamily lr=2e-5, bs=128, train\_noise\_std=0.02, hidden\_dim=32, c\_init=1, c\_hid=8, c\_final=8}
     & \textbf{93.89} & 93.71 & \textbf{20.95} & 7.69 \\

\addlinespace[6pt] 
\text{2hop  ($\sim$ 25.8 nodes)} \\ 
{\fontsize{8}{7}\ttfamily lr=2e-5, bs=128, train\_noise\_std=0.01, hidden\_dim=32}
 & 93.82 & 93.75 & \text{22.04} & 6.98 \\
\bottomrule
\end{tabular}
}
% \vspace{-2cm}
\end{table}

%\begin{figure}[t]
 %   \centering
  %  \includegraphics[width=0.4\linewidth]{aucVSkhop.png}
   % \caption{AUC vs $k$-hop neighborhood. We observe that adding more hops enhances the performance of PIFM, as it exploits more information.}
   % \label{fig:auc_khop}
%\end{figure}

\subsection{Transferability}
\label{app:transferability}

% We have experimented the 10\% and 50\%-drop-rate IMDB-B checkpoints for PIFM and used them on PROTEINS and ENZYMES dataset with the same drop rates, and the best results and the results at the end of the 100 sampling steps of each run are as shown below.
% \coauthor{In the ``Structural priors (no flow)'' block of the transferability table below, the IMDB-B$\to$PROTEINS and ENZYMES$\to$PROTEINS rows are identical (at both 10\% and 50\%) because the structural prior is computed on the target dataset and does not depend on the source checkpoint. Please confirm this is intended, and consider merging these rows or noting explicitly that the prior baseline does not transfer, since as listed they read like a copy error.}
% \coauthor{Caption/table mismatch: the caption says only \textsc{IMDB-B} checkpoints are transferred, but the table includes \textsc{PROTEINS}- and \textsc{ENZYMES}-source rows; and it says both the best-over-$100$-steps and the final-step values are reported, yet only one value set per row is shown. Please align the caption with the table (and add the second value set if the best-vs-final discussion is to be supported).}

{
We test whether a PIFM model trained on one dataset still works on another. We take a PIFM checkpoint trained on a \emph{source} dataset and run it on a different \emph{target} dataset at the same drop rate; Table~\ref{tab:transferability} reports these transfer results.

Since the flow always starts from the prior, we split the two parts. Table~\ref{tab:transfer_prior} shows the GraphSAGE prior on its own (no flow): this is just the prior computed on the target graphs, and it does not depend on the source checkpoint. Table~\ref{tab:transferability} then shows what the transferred flow adds on top of that prior. The gap between the two tables is the effect of the flow.
}

\begin{table}[H]
\centering
\caption{\textbf{PIFM transfer.} A PIFM flow trained on the \emph{source} dataset, run on a different \emph{target} dataset at the same drop rate, starting from the target's prior (Table~\ref{tab:transfer_prior}). For each run we report the step with the highest AUC ($K$ in parentheses) and the last step ($K{=}100$). Metrics are Average Precision (AP), AUC, False Negative Rate (FNR), and False Positive Rate (FPR), all in percent.}
\label{tab:transferability}
\small
\setlength{\tabcolsep}{5pt}
\begin{tabular}{l l l c c c c}
\toprule
Source checkpoint & Target dataset & Step & AP$\uparrow$ & AUC$\uparrow$ & FNR$\downarrow$ & FPR$\downarrow$ \\
\midrule
\multirow{2}{*}{PROTEINS (10\%)} & \multirow{2}{*}{ENZYMES (10\%)}
   & best ($K{=}1$)    & 43.87 & 76.96 & 71.96 & 4.24 \\
 & & final ($K{=}100$) & 37.83 & 59.43 & 64.45 & 4.87 \\
\addlinespace
\multirow{2}{*}{IMDB-B (10\%)} & \multirow{2}{*}{PROTEINS (10\%)}
   & best ($K{=}1$)    & 46.79 & 75.53 & 47.07 & 11.62 \\
 & & final ($K{=}100$) & 43.12 & 60.24 & 43.65 & 51.66 \\
\addlinespace
\multirow{2}{*}{ENZYMES (10\%)} & \multirow{2}{*}{PROTEINS (10\%)}
   & best ($K{=}1$)    & 49.63 & 76.60 & 59.02 & 6.63 \\
 & & final ($K{=}100$) & 42.36 & 70.64 & 53.12 & 8.20 \\
\addlinespace
\multirow{2}{*}{PROTEINS (50\%)} & \multirow{2}{*}{ENZYMES (50\%)}
   & best ($K{=}1$)    & 25.68 & 61.88 & 95.37 & 1.70 \\
 & & final ($K{=}100$) & 21.05 & 58.68 & 80.82 & 8.67 \\
\addlinespace
\multirow{2}{*}{IMDB-B (50\%)} & \multirow{2}{*}{PROTEINS (50\%)}
   & best ($K{=}75$)   & 24.88 & 59.02 & 61.02 & 24.32 \\
 & & final ($K{=}100$) & 24.42 & 57.22 & 60.80 & 24.37 \\
\addlinespace
\multirow{2}{*}{ENZYMES (50\%)} & \multirow{2}{*}{PROTEINS (50\%)}
   & best ($K{=}1$)    & 26.81 & 55.80 & 86.13 & 9.97 \\
 & & final ($K{=}100$) & 24.54 & 54.26 & 76.68 & 16.23 \\
\bottomrule
\end{tabular}
\end{table}

\begin{table}[H]
\centering
\caption{\textbf{Prior only (no flow).} The GraphSAGE prior computed on the target graphs --- the starting point for every PIFM run in Table~\ref{tab:transferability}. It depends only on the target, not on the source, so it is the same for every source$\to$target pair with the same target. Metrics are AP, AUC, FNR, and FPR, all in percent.}
\label{tab:transfer_prior}
\small
\begin{tabular}{l c c c c}
\toprule
Target dataset & AP$\uparrow$ & AUC$\uparrow$ & FNR$\downarrow$ & FPR$\downarrow$ \\
\midrule
ENZYMES (10\%)  & 41.28 & 73.70 & 13.49 & 60.59 \\
PROTEINS (10\%) & 46.36 & 74.58 & 11.00 & 63.50 \\
ENZYMES (50\%)  & 23.08 & 57.72 & 40.02 & 52.16 \\
PROTEINS (50\%) & 27.71 & 53.99 & 32.16 & 66.86 \\
\bottomrule
\end{tabular}
\end{table}

From the table we can see that the model degrades the predictions a lot at lower (10\%) drop rates, making the results incomparable when comparing to results in Table 9. It can be seen that with more steps integrated, at the final step $t = 100$ the metrics are much worse than the best result over all sampling steps or the structural prior. Although at higher (50\%) drop rates the best results are more comparable to the best results in Table 2, this is due to the graphs being very corrupted under 50\% drop rate and hence the observed graphs are much more degraded than they used to be.

\section{Visual results}

% \newpage
\subsection{Examples of reconstructed graphs}
\label{app:raw_graphs}

We show here a few samples for the expansion case.
We plot the samples from ENZYMES, using a subset of the dataset used in Section~\ref{subsec:blind}. 
%\coauthor{The text elsewhere refers to samples $3$, $7$, and $29$, but the figures appear to show samples $3$, $7$, and $30$. Please reconcile the sample indices.}

\begin{figure}[H]
    \centering
    \includegraphics[width=0.6\linewidth]{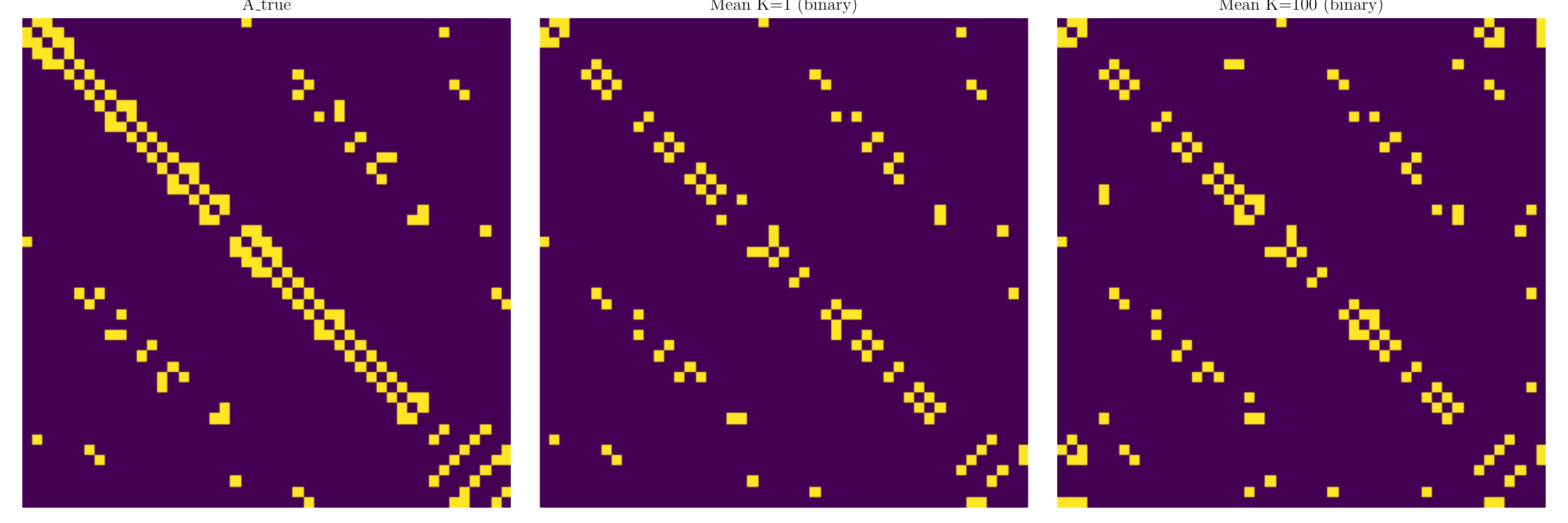}
    \caption{Graph reconstruction for sample 24, thresholded with 0.5}
    \label{fig:binary5}
\end{figure}

\begin{figure}[H]
    \centering
    \includegraphics[width=0.6\linewidth]{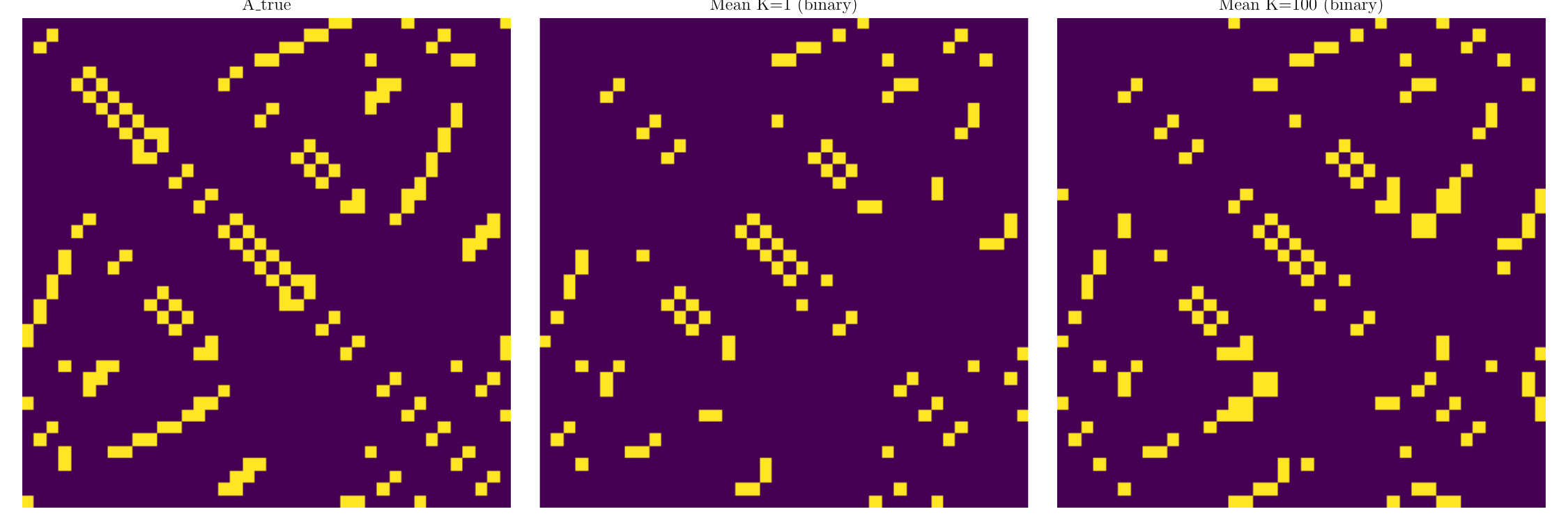}
    \caption{Graph reconstruction for sample 16, thresholded with 0.5}
    \label{fig:binary8}
\end{figure}

\begin{figure}[H]
    \centering
    \includegraphics[width=0.6\linewidth]{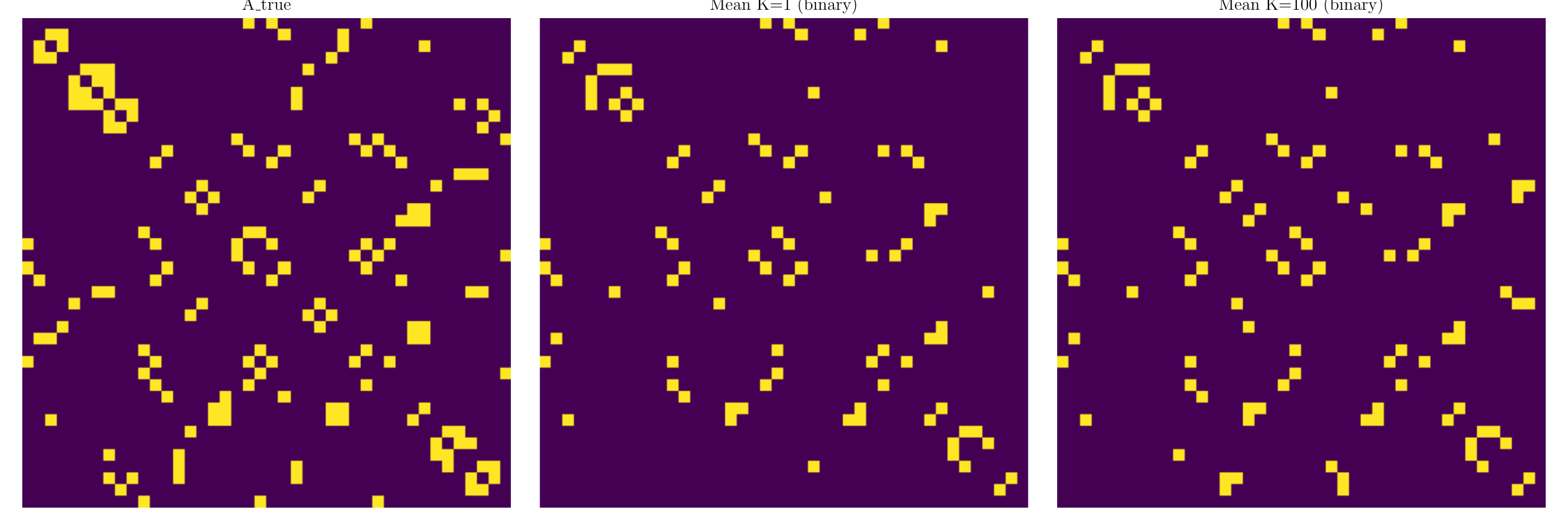}
    \caption{Graph reconstruction for sample 21, thresholded with 0.5}
    \label{fig:binary30}
\end{figure}

\paragraph{Binary comparison.}
In this case, we first compare the thresholded versions (with $0.5$) of the mean matrices.
We compute this for 3 graphs in the test set (24, 16 and 21).
The plots are in Figs.~\ref{fig:binary5},~\ref{fig:binary8} and~\ref{fig:binary30}

\begin{figure}[H]
    \centering
    \includegraphics[width=0.6\linewidth]{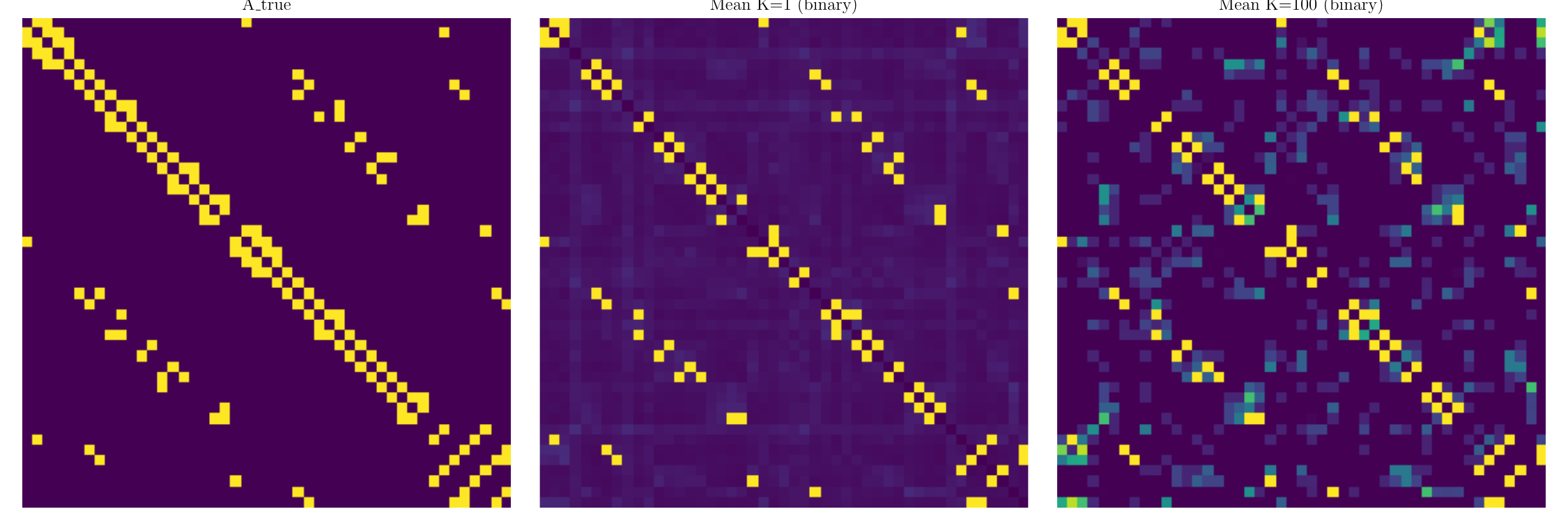}
    \caption{Graph reconstruction for sample 24, mean raw}
    \label{fig:mean5}
\end{figure}

\begin{figure}[H]
    \centering
    \includegraphics[width=0.6\linewidth]{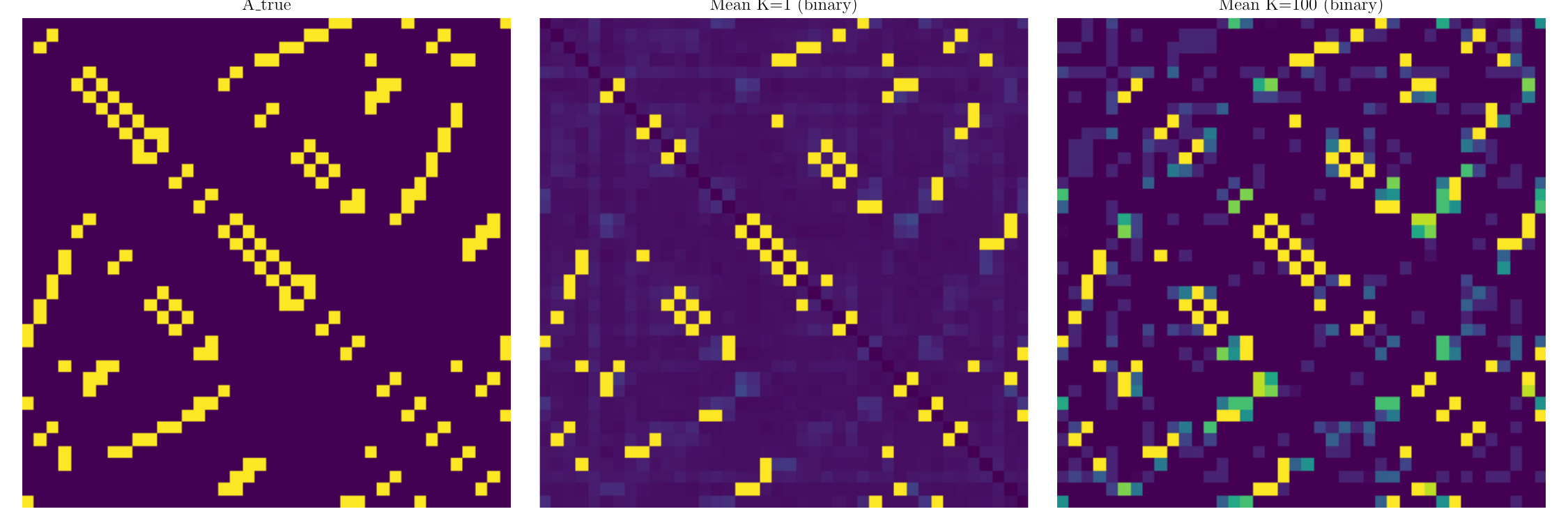}
    \caption{Graph reconstruction for sample 16, mean raw}
    \label{fig:mean8}
\end{figure}

\begin{figure}[H]
    \centering
    \includegraphics[width=0.6\linewidth]{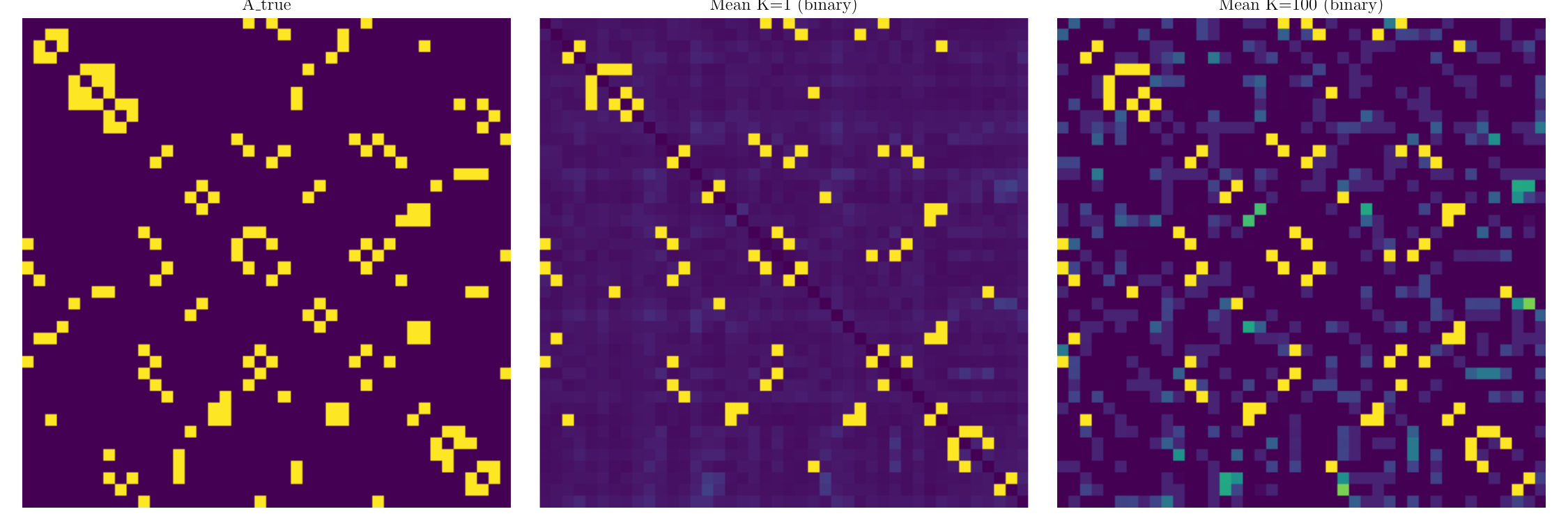}
    \caption{Graph reconstruction for sample 21, mean raw}
    \label{fig:mean30}
\end{figure}

\paragraph{Raw comparison - Mean.}
In this case, we compare the raw versions of the mean matrices.
We compute this for 3 graphs in the test set (24, 16 and 21).
The plots are in Figs.~\ref{fig:mean5},~\ref{fig:mean8} and~\ref{fig:mean30}.
Notice that the mean reconstructions for $K=100$ have values that are between 0 and 1; this can be explained by looking at individual samples (see below), which are more diverse, and therefore, they have non-overlapping set of existing edges.

\begin{figure}[H]
    \centering
    \includegraphics[width=0.6\linewidth]{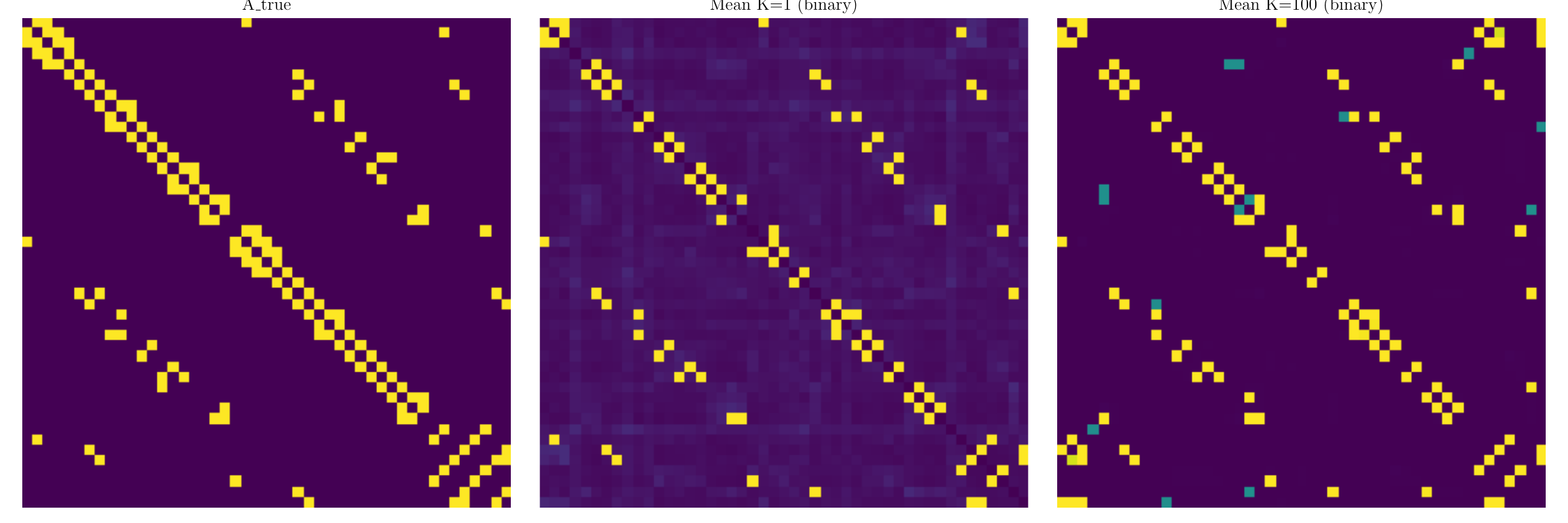}
    \caption{Graph reconstruction for sample 24, median raw}
    \label{fig:median5}
\end{figure}

\begin{figure}[H]
    \centering
    \includegraphics[width=0.6\linewidth]{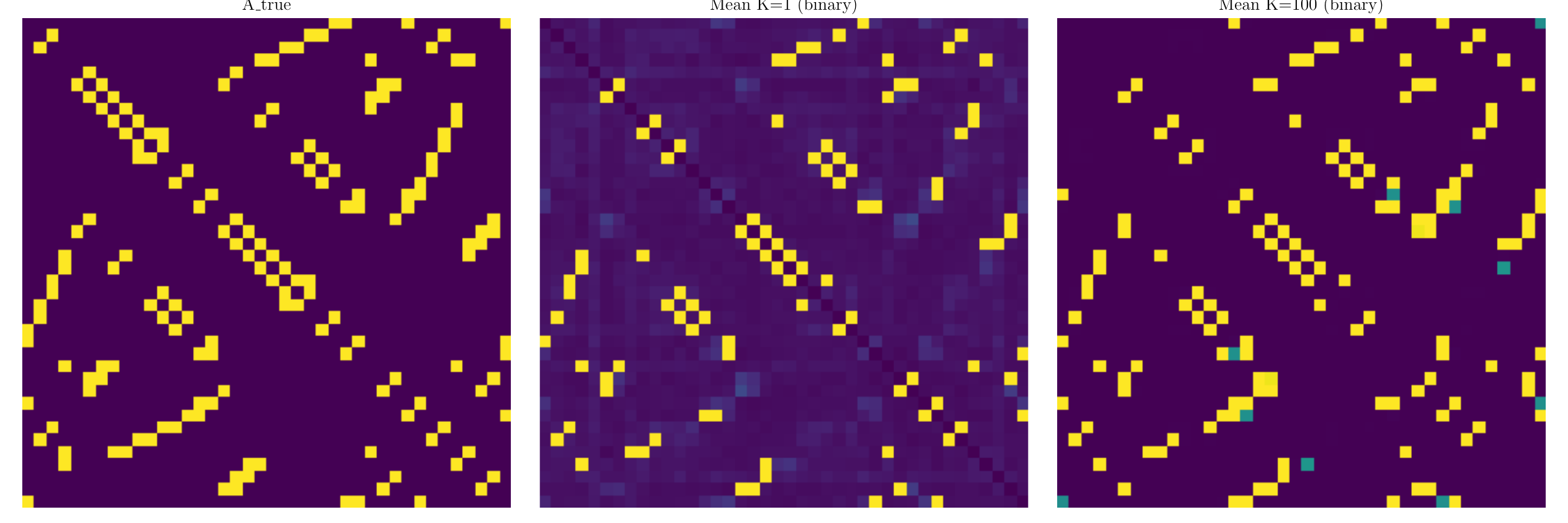}
    \caption{Graph reconstruction for sample 16, median raw}
    \label{fig:median8}
\end{figure}

\begin{figure}[H]
    \centering
    \includegraphics[width=0.6\linewidth]{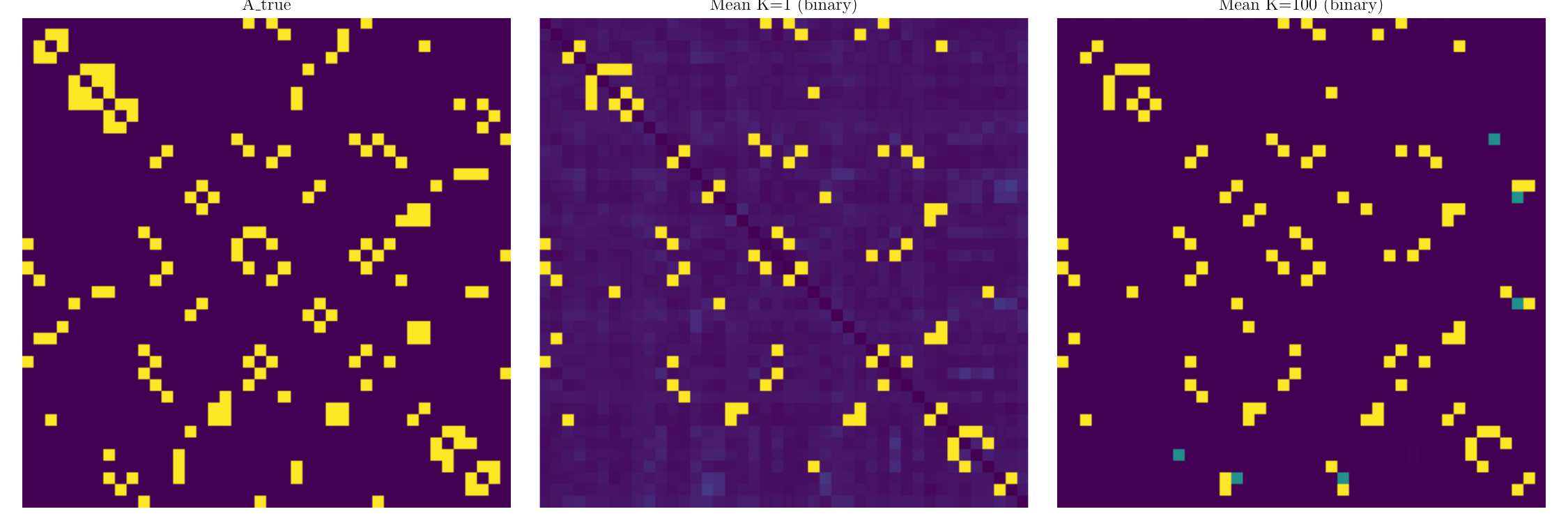}
    \caption{Graph reconstruction for sample 21, median raw}
    \label{fig:median30}
\end{figure}

\paragraph{Raw comparison - Median.}
In this case, we compare the raw versions of the median matrices.
We compute this for 3 graphs in the test set (24, 16 and 21).
The plots are in Figs.~\ref{fig:median5},~\ref{fig:median8} and~\ref{fig:median30}.

\paragraph{Individual samples for each graph.}
Lastly, we show the raw versions of different realizations (individual samples) for each graph.
We compute this for 3 graphs in the test set (24, 16 and 21).
Interestingly, the samples for $K = 100$ are more diverse (similar to the case of images in~\cite{ohayon2025posteriormean}); this diversity explains why the raw mean in Figs.~\ref{fig:mean5},~\ref{fig:mean8} and~\ref{fig:mean30} have values that are not exactly 0 or 1 (which indicates that the samples disagree on these entries rather than sharing a single structure).
%\coauthor{The text elsewhere refers to samples $3$, $7$, and $29$, but the figures appear to show samples $3$, $7$, and $30$. Please reconcile the sample indices.}

\begin{figure}[H]
    \centering
    \includegraphics[width=0.9\linewidth]{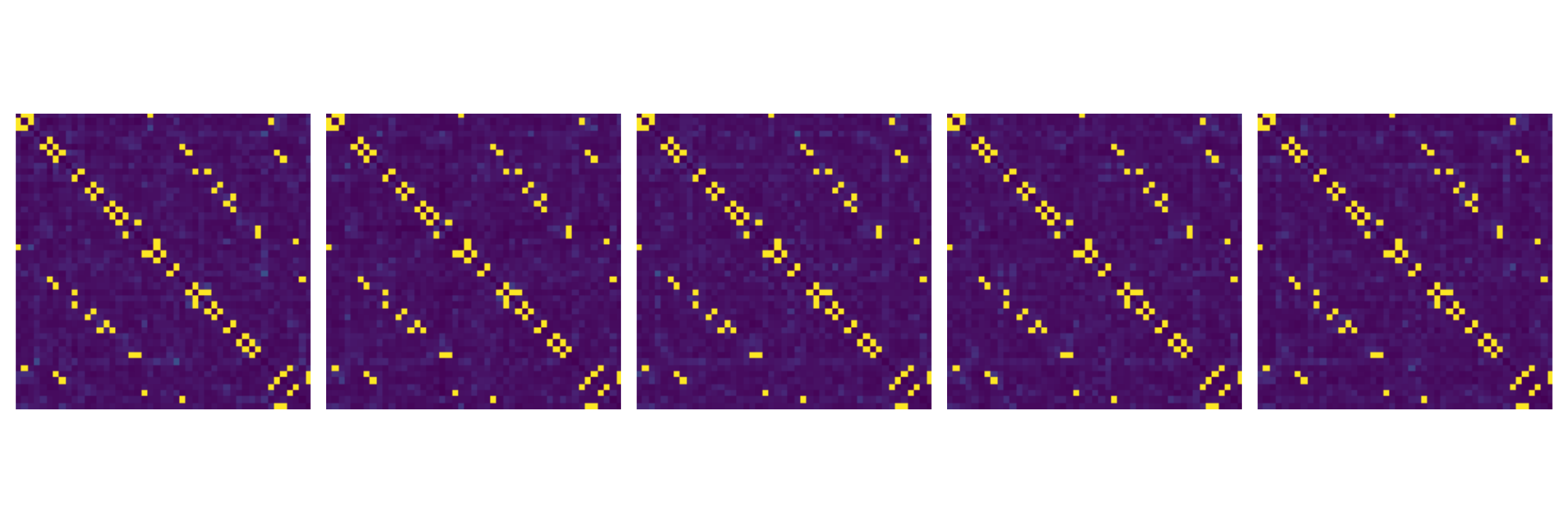}
    \caption{Individual samples for $K = 1$ and for sample 24}
    \label{fig:individual_5_k1}
\end{figure}

\begin{figure}[H]
    \centering
    \includegraphics[width=0.9\linewidth]{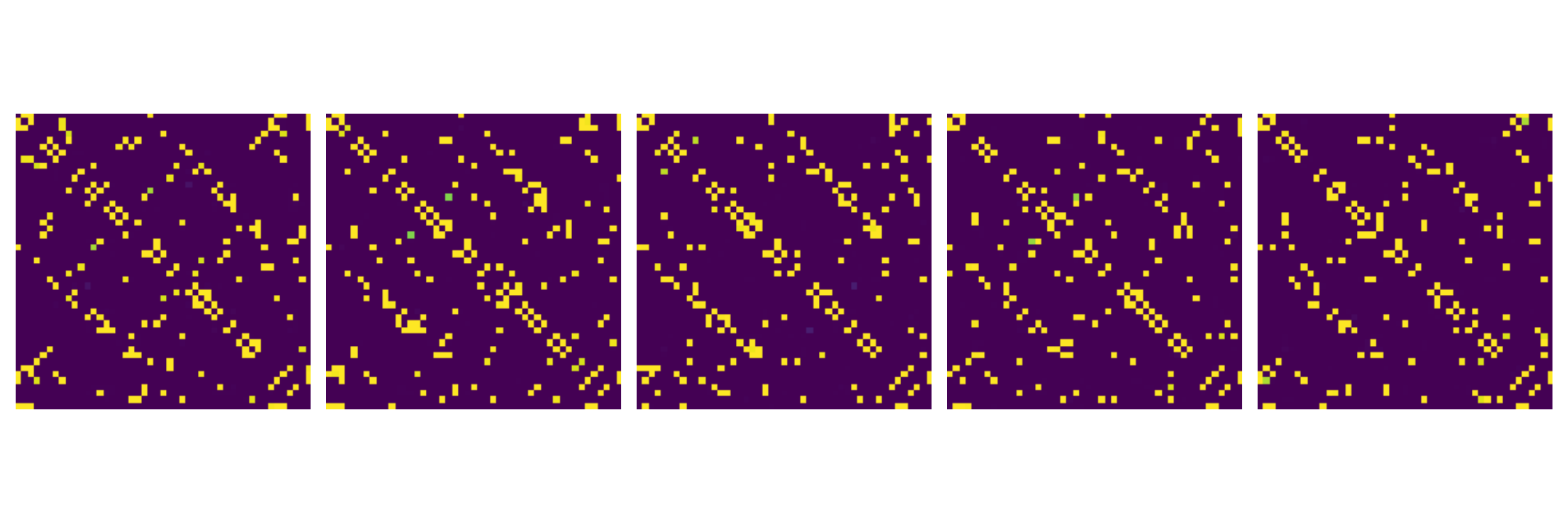}
    \caption{Individual samples for $K = 100$ and for sample 24}
    \label{fig:individual_5_k100}
\end{figure}

\begin{figure}[H]
    \centering
    \includegraphics[width=0.9\linewidth]{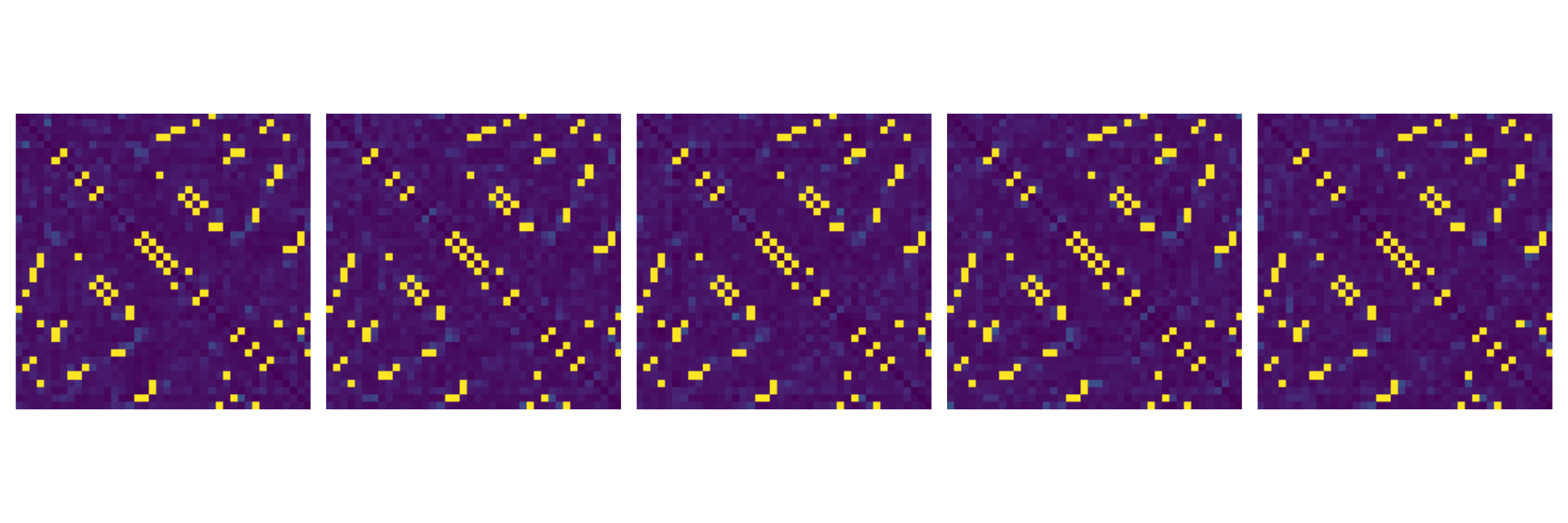}
    \caption{Individual samples for $K = 1$ and for sample 16}
    \label{fig:individual_8_k1}
\end{figure}

\begin{figure}[H]
    \centering
    \includegraphics[width=0.9\linewidth]{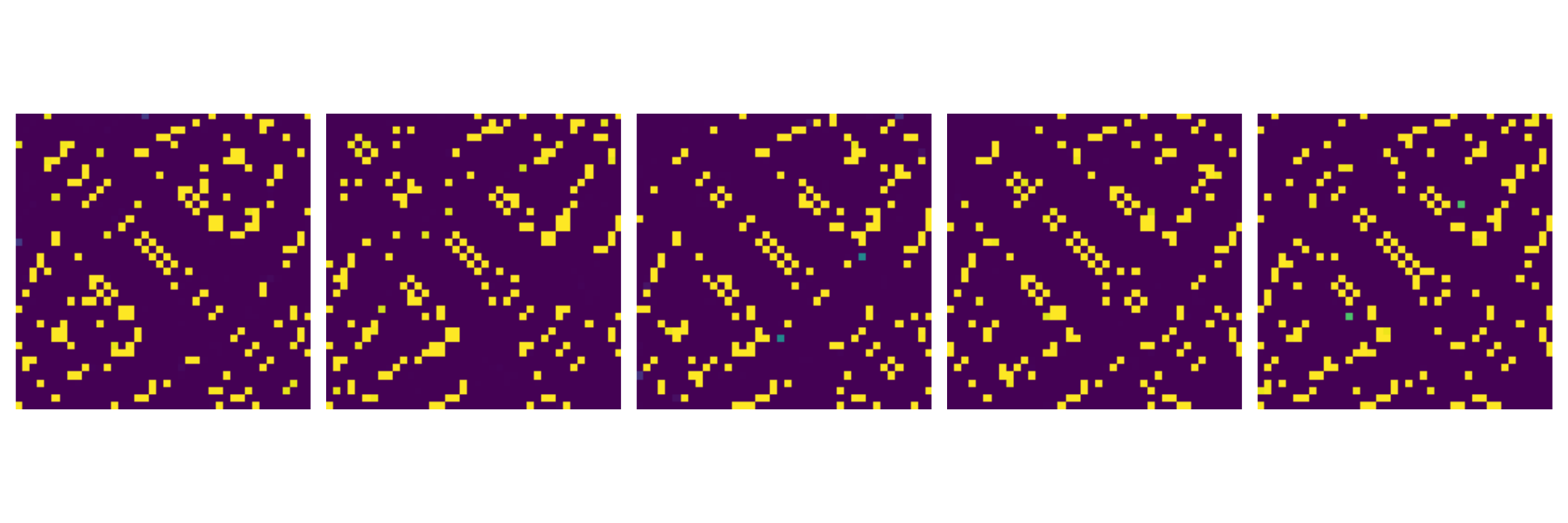}
    \caption{Individual samples for $K = 100$ and for sample 16}
    \label{fig:individual_8_k100}
\end{figure}

\begin{figure}[H]
    \centering
    \includegraphics[width=0.9\linewidth]{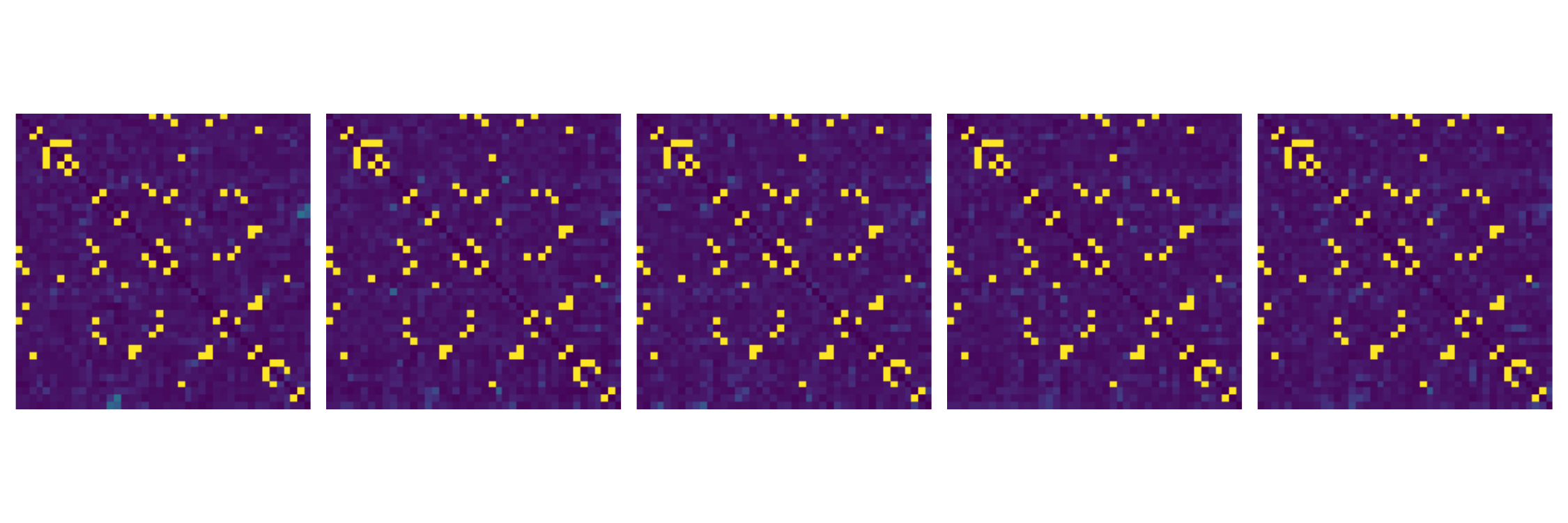}
    \caption{Individual samples for $K = 1$ and for sample 21}
    \label{fig:individual_30_k1}
\end{figure}

\begin{figure}[H]
    \centering
    \includegraphics[width=0.9\linewidth]{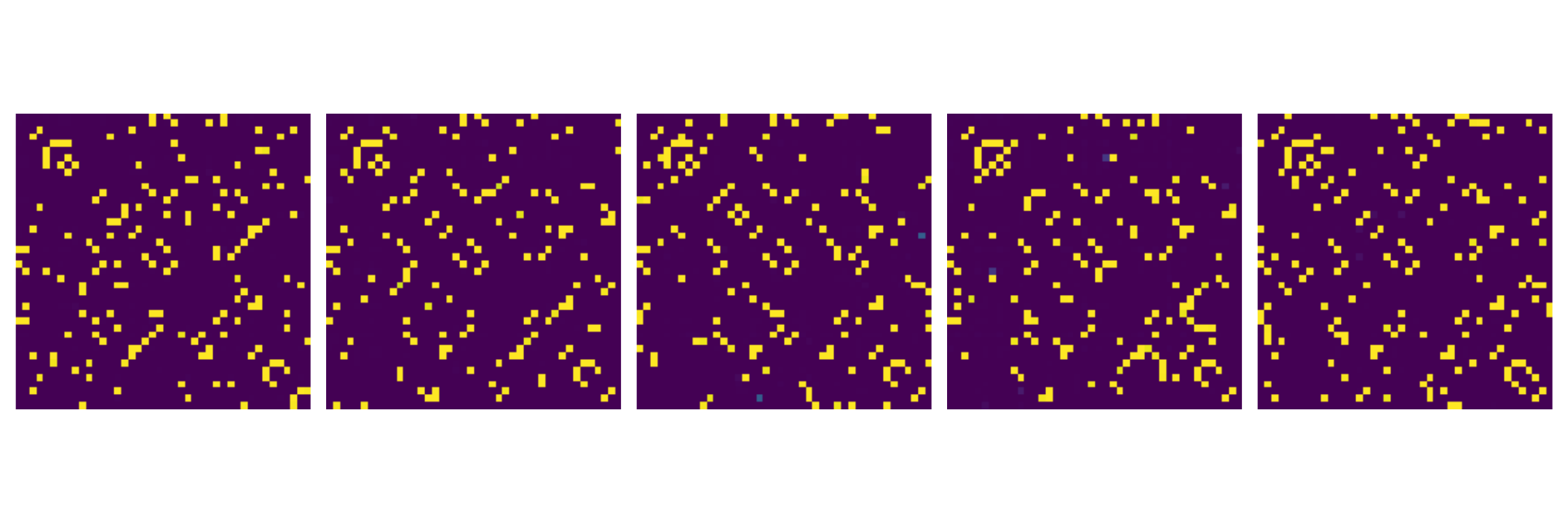}
    \caption{Individual samples for $K = 100$ and for sample 21}
    \label{fig:individual_30_k100}
\end{figure}
% \newpage
\subsection{Intermediate adjacency matrices}

\begin{figure}[H]
    \centering
    \includegraphics[width=0.7\linewidth]{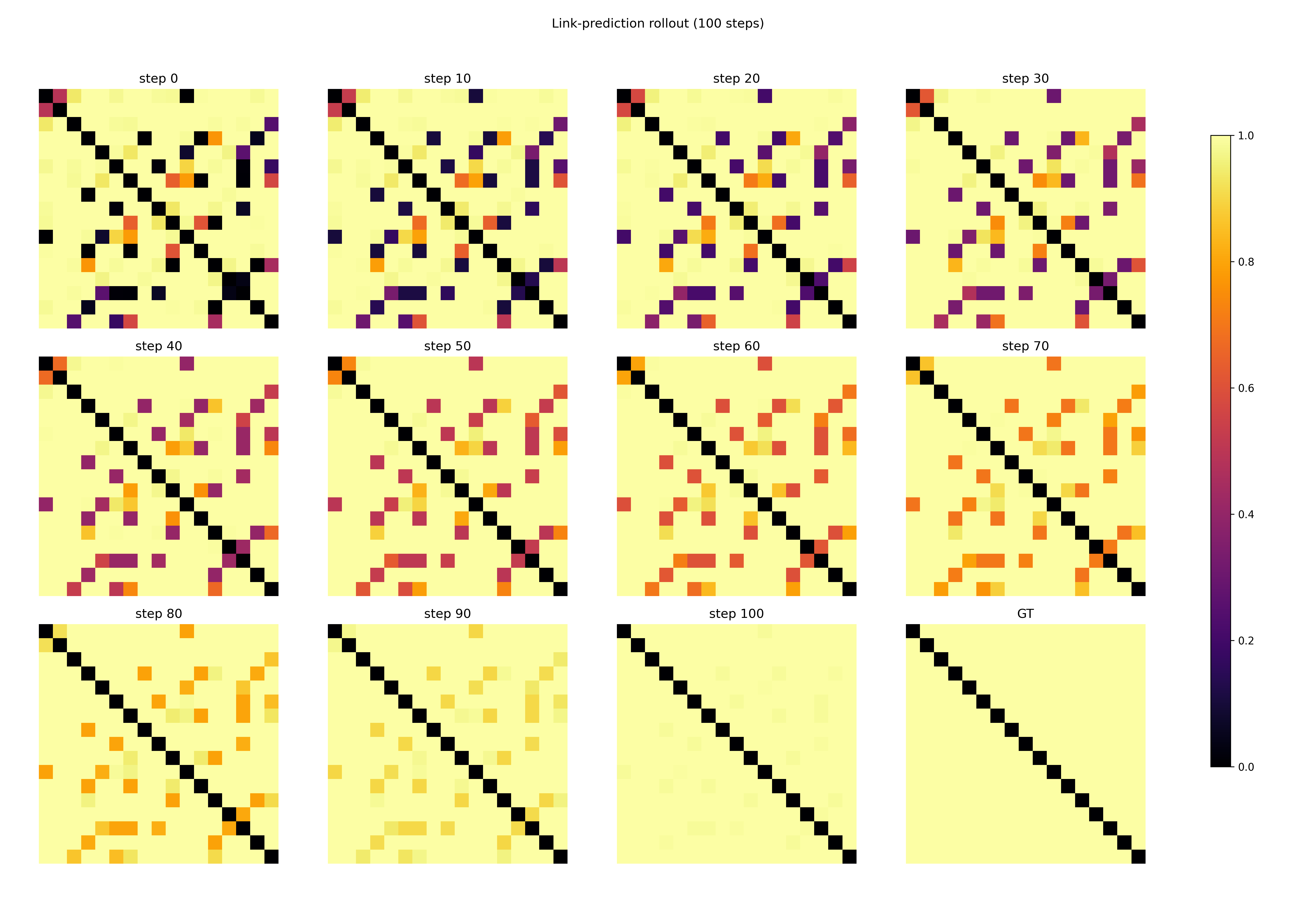}
    \caption{Visualization of IMDB 50\% drop rate reconstruction. (Graph 1)}
    \label{fig:inter1}
\end{figure}

\begin{figure}[H]
    \centering
    \includegraphics[width=0.7\linewidth]{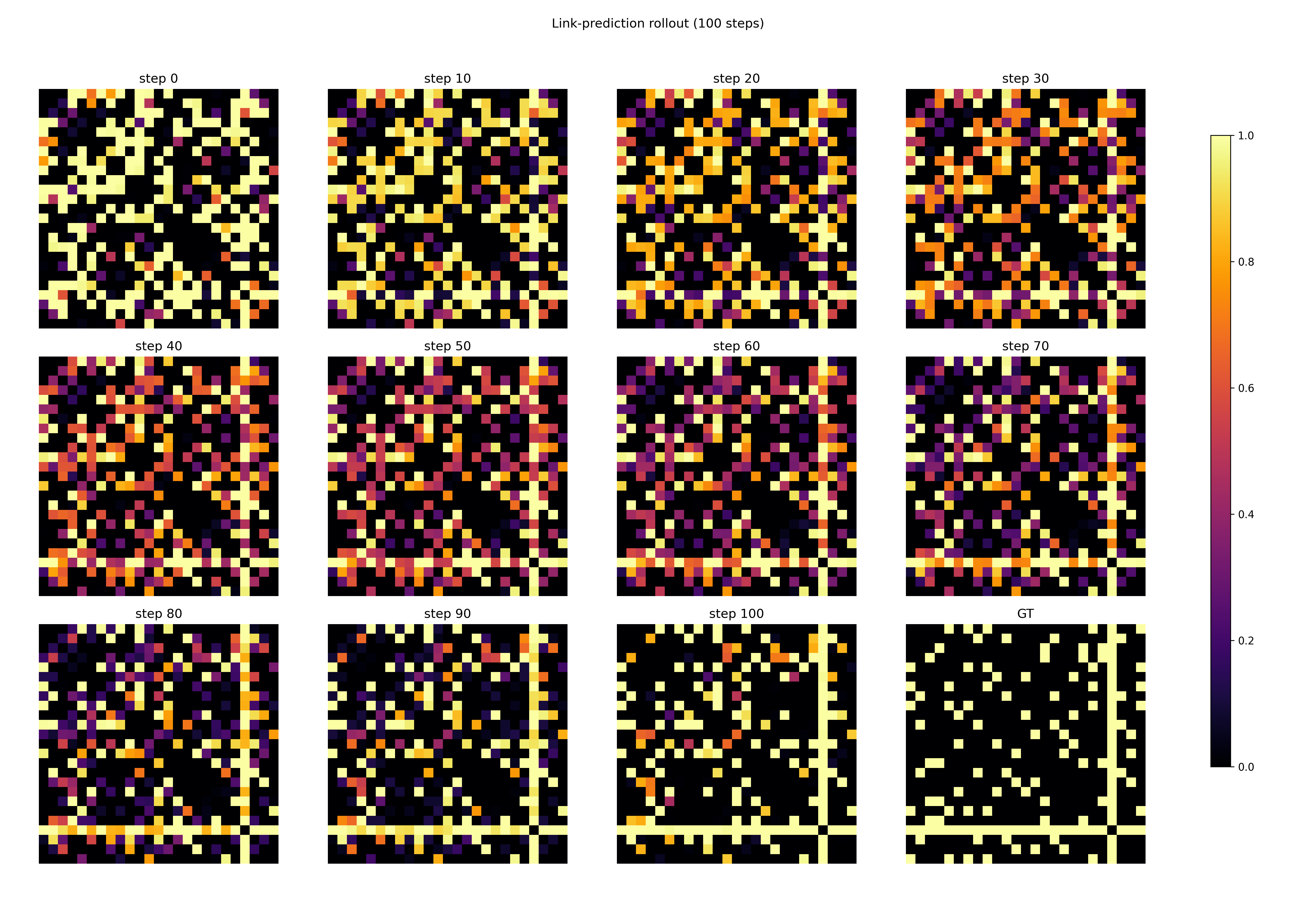}
    \caption{Visualization of IMDB 50\% drop rate reconstruction. (Graph 2)}
    \label{fig:inter2}
\end{figure}

\begin{figure}[H]
    \centering
    \includegraphics[width=0.7\linewidth]{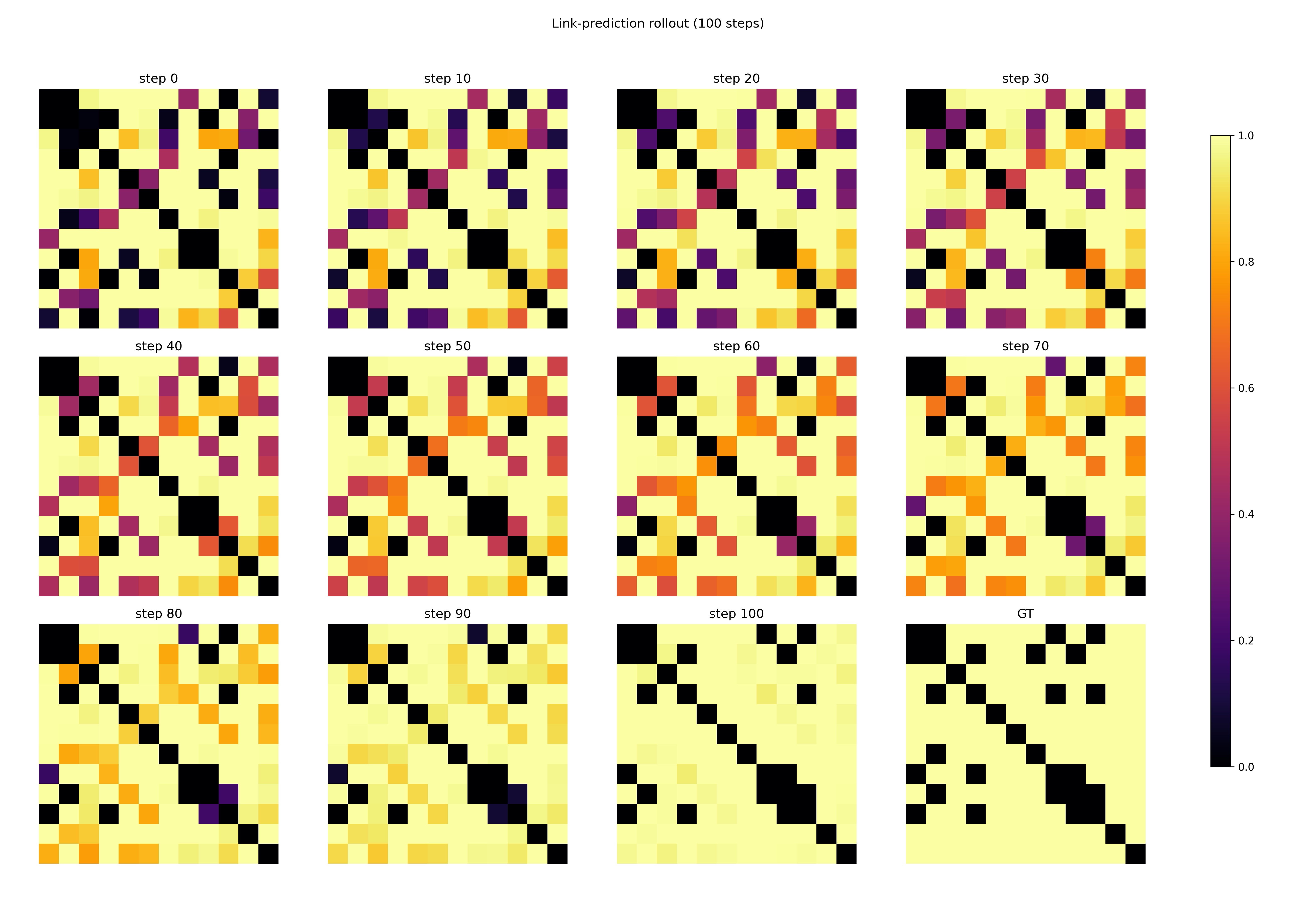}
    \caption{Visualization of IMDB 50\% drop rate reconstruction. (Graph 3)}
    \label{fig:inter3}
\end{figure}

In Figs.~\ref{fig:inter1}-~\ref{fig:inter3}, we show visualizations of the diffusion trajectory of the link-prediction sampler by "snapshotting" the predicted adjacency matrix at steps 0 to 100 in 10-step-increments of a sample path with 100 total steps. Each panel shows the raw  adjacency values, zeroed on the diagonal, rendered with the colormap with black associated to $0$ and yellow to $1$. 
The bottom-rightmost panel is the ground-truth adjacency for comparison. 

To see the sampling process, progress from left-to-right and top-to-bottom shows how the sampler denoises toward the final reconstruction (steps=100). From these images we could see the smooth transitions along the full reconstruction trajectory of PIFM.

\end{document}